\documentclass[10pt,twocolumn,letterpaper]{article}

\usepackage{iccv}
\usepackage{times}
\usepackage{epsfig}
\usepackage{graphicx}
\usepackage{amsmath}
\usepackage{amssymb}

\usepackage{comment}
\usepackage[accsupp]{axessibility}
\usepackage{subcaption}
\usepackage{multirow}
\usepackage{bm}
\usepackage{bbm}
\usepackage{color}
\usepackage{xcolor,soul}
\usepackage{colortbl}
\usepackage{mathrsfs}
\usepackage{booktabs}
\usepackage[flushleft]{threeparttable}
\usepackage[accsupp]{axessibility}

\usepackage[page,header]{appendix}
\usepackage{titletoc}

\newcommand{\sysname}{StillMix}
\newcommand{\datasetnameA}{SCUBA}
\newcommand{\datasetnameB}{SCUFO}

\usepackage[pagebackref=true,breaklinks=true,letterpaper=true,colorlinks,bookmarks=false]{hyperref}

\iccvfinalcopy 

\def\iccvPaperID{3792} 

\ificcvfinal\fi

\begin{document}



\title{Mitigating and Evaluating Static Bias of Action Representations in the Background and the Foreground}


\author{Haoxin Li\textsuperscript{\rm 1},
Yuan Liu\textsuperscript{\rm 2},
Hanwang Zhang\textsuperscript{\rm 1},
Boyang Li\textsuperscript{\rm 1}
\\
\textsuperscript{\rm 1}{Nanyang Technological University}\quad \textsuperscript{\rm 2}{Guangzhou University}\\
\texttt{\small{\{haoxin003, hanwangzhang, boyang.li\}@ntu.edu.sg, yuanliu@gzhu.edu.cn}}
}

\maketitle
\ificcvfinal\thispagestyle{empty}\fi

\begin{abstract}
In video action recognition, shortcut static features can interfere with the learning of motion features, resulting in poor out-of-distribution (OOD) generalization.
The video background is clearly a source of static bias, but the video foreground, such as the clothing of the actor, can also provide static bias. In this paper, we empirically verify the existence of foreground static bias by creating test videos with conflicting signals from the static and moving portions of the video. 
To tackle this issue, we propose a simple yet effective technique, \sysname{}, to learn robust action representations. Specifically, \sysname{} identifies bias-inducing video frames using a 2D reference network and mixes them with videos for training, serving as effective bias suppression even when we cannot explicitly extract the source of bias within each video frame or enumerate types of bias.
Finally, to precisely evaluate static bias, we synthesize two new benchmarks, \datasetnameA{} for \underline{s}tatic \underline{cu}es in the \underline{ba}ckground, and \datasetnameB{} for \underline{s}tatic \underline{cu}es in the \underline{fo}reground.
With extensive experiments, we demonstrate that \sysname{} mitigates both types of static bias and improves video representations for downstream applications. Code is available at \url{https://github.com/lihaoxin05/StillMix}.
\end{abstract}

\section{Introduction}
\label{sec:intro}

\begin{figure}[t]
\centering
\includegraphics[width=1.0\linewidth]{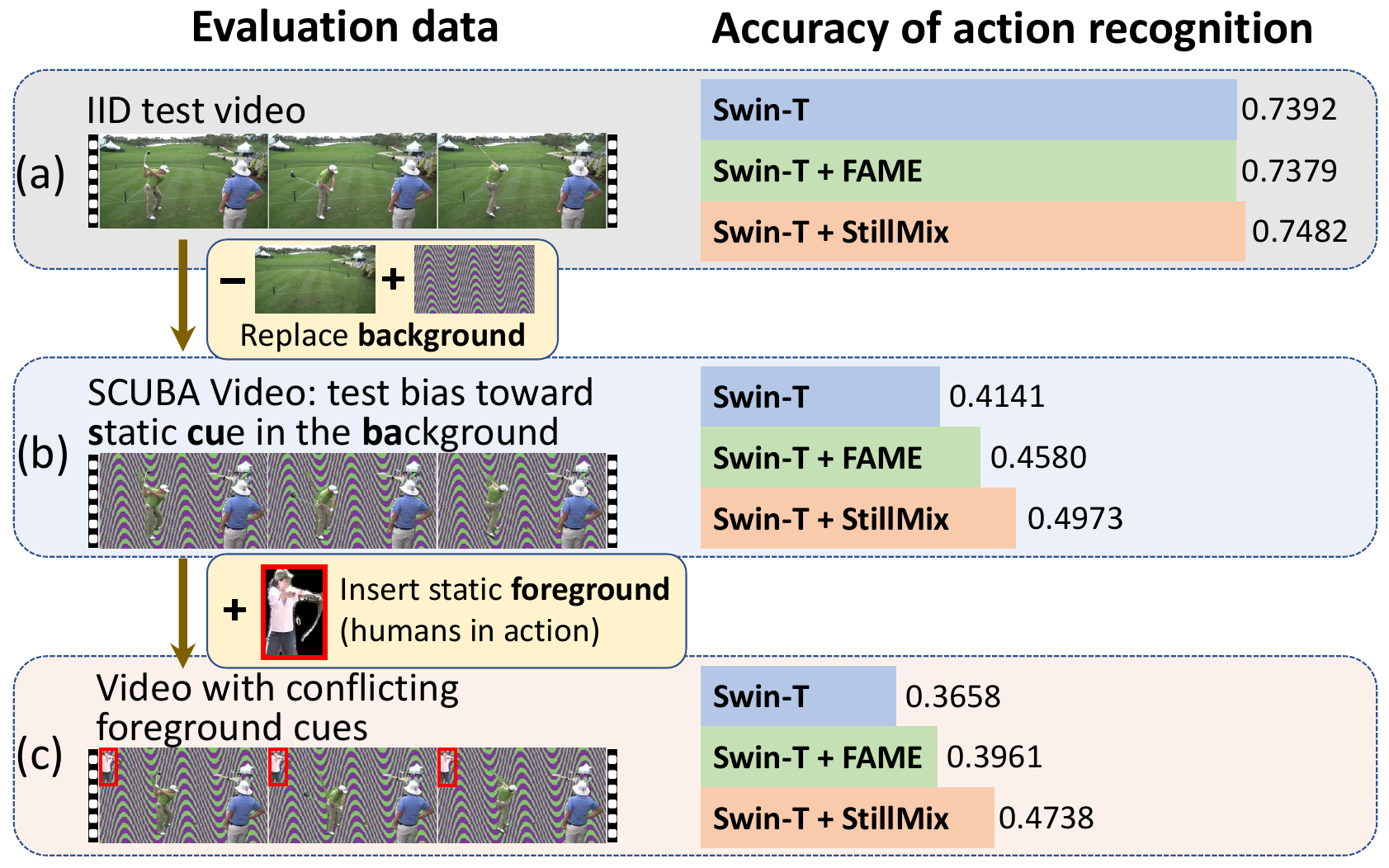}
\caption{Evaluation of background and foreground static bias. (a) Testing on IID HMDB51~\cite{kuehne2011hmdb} test videos. (b) Testing on \datasetnameA{} videos, constructed by replacing the video background with a synthetic sinusoidal stripe image. (c) Testing on videos with conflicting foreground cues, constructed by inserting a random static foreground into the \datasetnameA{} video.}
\label{fig:bg_fg_bias}
\end{figure}

Traditional computer vision techniques perform well on independent and identically distributed (IID) test data, but often lack out-of-distribution (OOD) generalization \cite{Underspecification-2020,pmlr-v139-koh21a,djolonga2021robustness}. This is intimately tied to the learning of shortcut features \cite{Ilyas2019,Geirhos_2020,geirhos2018imagenet}, which are easy to learn and correlate strongly with IID labels but cause poor OOD generalization \cite{shah2020pitfalls,wang2019learning,pezeshki2021gradient,hermann2020shapes}. In video action recognition, shortcut features often manifest as static cues. For example, a network may classify a video as \emph{golf swinging} based on its background, a golf course, even if the motion patterns indicate another action such as \emph{walking}. While static cues can provide valuable information \cite{zhang2022hierarchically,ding2022dual,zhou2022decoupling}, they often outcompete motion features \cite{huang2018makes,li2018resound,li2019repair,sevilla2021only,yun2022time} and result in low OOD performance \cite{li2019repair,wang2021removing,ilic2022appearance}. 
In contrast to the rich literature on mitigating background static bias (\eg, golf courses for \emph{golf swinging}) \cite{choi2019can,wang2021removing,zhang2021suppressing,Ding_2022_CVPR,chungenabling}, foreground static bias has been underexplored. Examples of foreground bias include swimsuits for \emph{swimming} and guitars for \emph{guitar playing} --- people can swim without swimsuits or show guitars in the video without playing them. 

\begin{figure*}[t]
\centering
\includegraphics[width=\linewidth]{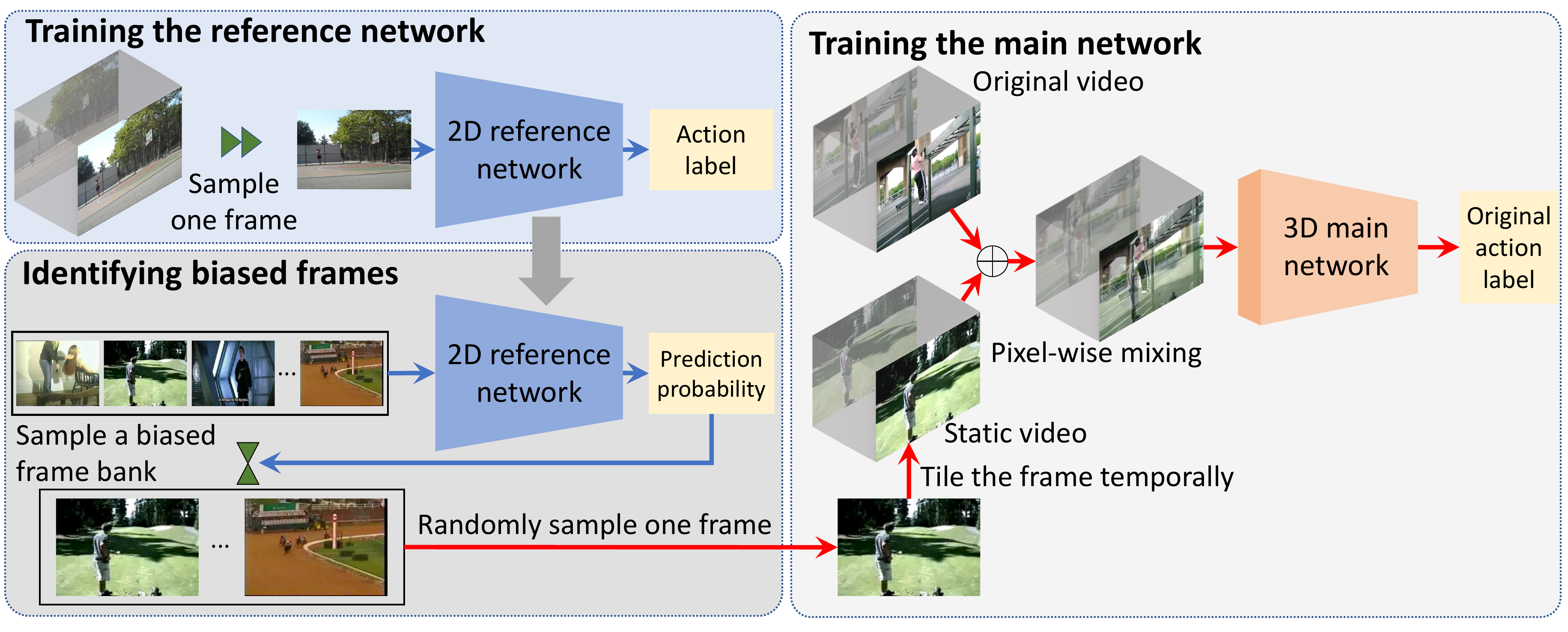}
\caption{An illustration of \sysname{}. We train a 2D reference network that classifies still frames into actions to capture static bias. With the reference network, we sample frames inducing static bias to construct a biased frame bank. We mix the frames from the bank with a given video to generate an augmented video, which is used to train a 3D main network to mitigate static bias.}
\label{fig:algo}
\end{figure*}

The first question we ask is if foreground static bias exists and if it is captured by the representations learned by neural networks. 
Our investigation technique is to create test videos with conflicting action cues from the moving part and the static part of the video. 
In the first step, shown in Figure~\ref{fig:bg_fg_bias}(b), we replace the backgrounds of IID HMDB51~\cite{kuehne2011hmdb} test videos by sinusoidal stripe images. These videos have no meaningful backgrounds, so the action information must come from the foreground. Therefore, 
models overly reliant on background static cues should perform poorly. A background debiasing technique, FAME~\cite{Ding_2022_CVPR}, coupled with a tiny Video Swin Transformer (Swin-T)~\cite{liu2021video}, works relatively well on this test. 

In the second step, shown in Figure~\ref{fig:bg_fg_bias}(c), from a single frame of a random video, we extract its foreground (mainly human actors), and insert the static foreground into all the frames of the current \datasetnameA{} video. The resultant video contains only two action features: a static foreground that indicates one action label and a moving foreground that indicates another action label. Predictions made using the static foreground would be wrong. This design allows the quantification of foreground static bias. More details can be found in Sec. S1 of the Supplementary Material. 

The results clearly show the existence of foreground static bias and its negative effects. On the second test set, both Swin-T and Swin-T+FAME suffer similar degradation and perform 5\% worse than \datasetnameA{} videos. FAME works by procedurally isolating the foreground regions from each frame and use those for training. However, it is hard to separate the foreground motion from the static foreground (\eg, clothing, equipment, or other people attributes \cite{li2018resound}) in the training videos, since both types of features are strongly tied to the human actors.





We propose \sysname{}, a technique that mitigates static bias in both the background and the foreground, without the need to explicitly isolate (or even enumerate \cite{choi2019can}) the bias-inducing content within a frame. \sysname{} identifies bias-inducing frames using a reference network and mixes them with training videos without affecting motion features. The process is illustrated in Figure~\ref{fig:algo}. Unlike FAME, \sysname{} could suppress static bias anywhere in a frame, including the background and the foreground. In Figure~\ref{fig:bg_fg_bias}, \sysname{} outperforms FAME and suffers only 2\% accuracy drop on the second benchmark, highlighting its resilience.

\begin{figure}[t]
\centering
\includegraphics[width=\linewidth]{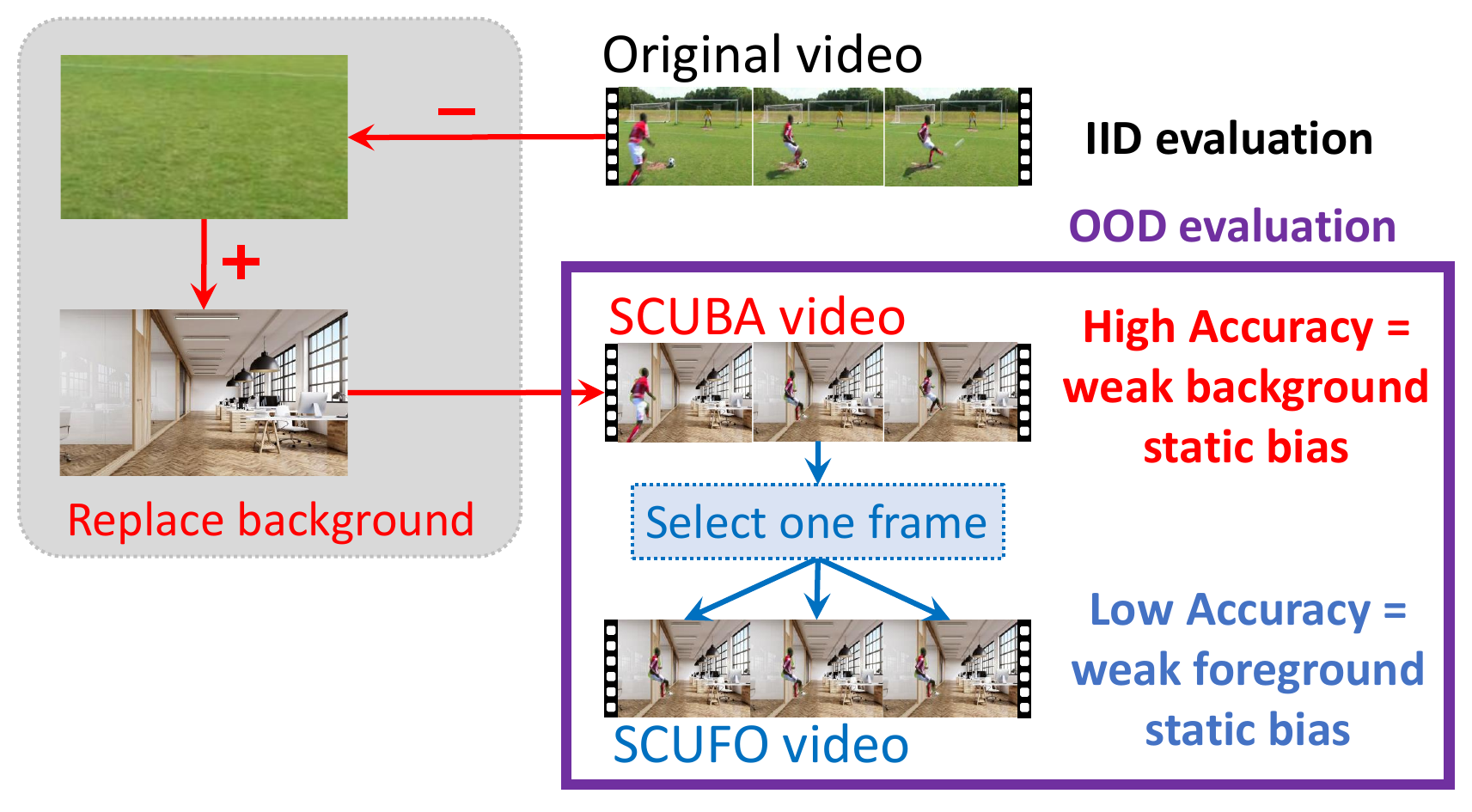}
\caption{An illustration of OOD benchmark construction. To quantify \underline{s}tatic \underline{cu}es in the \underline{ba}ckground, we reserve the foreground actions and replace the backgrounds with other images to synthesize \datasetnameA{} videos. To quantify \underline{s}tatic \underline{cu}es in the \underline{fo}reground, we randomly select one frame in the \datasetnameA{} video and stack it into a single-frame video without motion, named \datasetnameB{} videos.}
\label{fig:illu_dataset_construction}
\end{figure}

Evaluating OOD action recognition is challenging as test videos with OOD foregrounds, such as \emph{swimming} without swimsuits or \emph{cycling} while carrying a guitar, are rare. 
To pinpoint the static bias in either the background or the foreground, we create new synthetic sets of OOD benchmarks by altering the static features in IID test videos, as illustrated in Figure~\ref{fig:illu_dataset_construction}. Specifically, we retain the foregrounds of actions and replace the backgrounds with diverse natural and synthetic images. This procedure yields a test set that quantifies representation bias toward \underline{s}tatic \underline{cu}es in the \underline{ba}ckground (\datasetnameA{}). Second, we create videos that repeat a single random frame from \datasetnameA{}, producing a test set that quantifies representation bias toward \underline{s}tatic \underline{cu}es in the \underline{fo}reground (\datasetnameB{}). As these videos disassociate the backgrounds from the action and contain no motion, their actions can be recognized by only static foreground features. Thus, high accuracy on \datasetnameB{} indicates strong foreground static bias.

With the synthetic OOD benchmarks, we extensively evaluate several mainstream action recognition methods and make the following observations. First, all examined methods exhibit static bias. 
Second, existing debiasing methods like ActorCutMix \cite{zou2021learning} and FAME \cite{Ding_2022_CVPR} demonstrate resistance to background static bias, but remain vulnerable to foreground static bias. In contrast, the proposed \sysname{} consistently boosts performance of action recognition models and compares favorably with the other debiasing techniques on both background and foreground static bias. In addition, \sysname{} improves the performance of transfer learning and downstream weakly supervised action localization.

The paper makes the following contributions:
\vspace{-0.5em}
\begin{itemize}
    \setlength{\itemindent}{0em}
    \setlength{\itemsep}{-0.25em}
    \item Through quantitative experiments, we highlight the importance to address foreground static bias in learning robust action representations. 
    \item We propose \sysname{}, a video data augmentation technique to mitigate static bias in not only the background but also the foreground.
    \item We create new benchmarks to quantitatively evaluate static bias of action representations and pinpoint the source of static bias (backgrounds or foregrounds). 
    \item We compare action recognition methods on the created benchmarks to reveal their characteristics and validate the effectiveness of \sysname{}.
\end{itemize}

\section{Related Work}
\noindent\textbf{Bias Evaluation.} Biases are surface features that are easily learned by neural networks and strongly influence their predictions. Such features perform well on IID data \cite{valle2018deep,kalimeris2019sgd} but do not generalize to OOD data \cite{shah2020pitfalls,pezeshki2021gradient}. In action recognition, models easily capture static bias \cite{li2018resound,li2019repair,choi2019can,wang2021removing}. The following methods are used for bias evaluation: \emph{(1) Visualization techniques}  \cite{feichtenhofer2020deep,manttari2020interpreting} visualize the regions that models focus on to interpret the static bias qualitatively. \emph{(2) Proxy data or tasks.} Synthetic videos with altered backgrounds \cite{chungenabling}, videos with white-noise textures \cite{ilic2022appearance}, dynamic texture videos \cite{hadji2018new,broome2023recur} are used to reveal the bias toward backgrounds or texture. Proxy tasks evaluating temporal asymmetry, continuity, and causality are designed to show the static bias in video representations \cite{ghodrati2018video}. \emph{(3) Mutual information.} \cite{kowal2022deeper} quantifies the static bias using mutual information between representations of different types of videos. Although these works evaluate the static bias in the whole video, they do not specify the source of static bias. In this paper, we create new benchmarks to pinpoint the source of static bias as the background and the foreground.

\noindent\textbf{Bias Mitigation.} Prevalent techniques of mitigating bias in action representations can be broadly classified into four categories. \emph{(1) Attribute supervision.} \cite{choi2019can} uses scene pseudo-labels and human masks to discourage models from predicting scenes and recognizing actions without human, but it needs extra attribute labels. \emph{(2) Re-weighting.} \cite{li2018resound,li2019repair} identify videos containing bias and downweight them in training, but \cite{wang2019balanced} suggests merely weight adjustment is insufficient. \emph{(3) Context separation.} \cite{wang2018pulling} learns to separate action and contexts by collecting samples with similar contexts but different actions. \emph{(4) Data augmentation.} Similar to the proposed \sysname{}, a few works utilize augmented videos. BE \cite{wang2021removing} mixes a frame from a video with other frames in the same video. ActorCutMix \cite{zou2021learning}, FAME \cite{Ding_2022_CVPR}, ObjectMix \cite{kimata2022objectmix} and FreqAug \cite{kim2022spatiotemporal} carefully carve out the foreground (human actors or regions of motion), and replace the background with other images to create augmented training data. SSVC \cite{zhang2021suppressing} and MCL \cite{li2021motion} focus the models to the dynamic regions. However, these methods have not addressed static cues in the foreground. 

A particular advantage of \sysname{} is that it does not require specially designed procedures to carve out the bias-inducing pixels within the frames like ActorCutMix \cite{zou2021learning} and FAME \cite{Ding_2022_CVPR}, or even to enumerate the source of bias like \cite{choi2019can}. Rather, it automatically identifies bias-inducing frames using a reference network. Consequently, \sysname{} can suppress static bias in both the background and the foreground.

\sysname{} is also similar to two debiasing techniques designed for image recognition and text classification \cite{nam2020learning,liu2021just}, which use a reference network to identify bias-inducing data instances. However, \sysname{} exploits the special property of videos that they can be decomposed into individual frames. \sysname{} identifies bias-inducing components (frames) using 2D networks rather than whole data points as in \cite{nam2020learning,liu2021just}.

\noindent\textbf{Action Recognition.}
3D convolution or decomposed 3D convolutions \cite{6165309,7410867,carreira2017quo,7940083,Tran_2018_CVPR,lin2018temporal} are popular choices for action recognition. Two-stream architectures employ two modalities to classify actions, such as both RGB frames and optical flow \cite{NIPS2014_5353,8454294}, or videos with two different frame rates and resolutions \cite{feichtenhofer2019slowfast}. Multi-scale temporal convolutions or feature fusion are designed for fine-grained actions with strong temporal structures \cite{Zhou_2018_ECCV,hussein2018timeception,9157646,9157586}. Transformer networks are proposed to capture the long-range dependencies \cite{arnab2021vivit,bertasius2021space,liu2021video}. However, our understanding of the representations learned by these models remains limited. In this paper, we create benchmarks to evaluate what features are captured by action models and propose a simple data augmentation method that effectively improves the robustness of action models.

\section{The \sysname{} Technique}
\label{sec:approach}
In order to learn robust and generalizable action representations that are invariant to static cues, we propose a simple but effective video data augmentation technique, \sysname{}. Instead of using manually designed rules to identify and remove biased data from the training set, as in ActorCutMix \cite{zou2021learning} and FAME \cite{Ding_2022_CVPR}, \sysname{} learns to identify still frames that induce biased representation using a neural network and mitigate static bias through mixing the identified frames with videos. As a result, \sysname{} offers a flexible bias suppression technique that works for both the background and the foreground.

We begin with some notations. We denote the $i^\text{th}$ video in the training set as tensor $\boldsymbol{x}_{i}\in\mathbb{R}^{C\times T\times H\times W}$, where $C$, $T$, $H$ and $W$ are the number of channels, number of frames, height and width of the video, respectively. The associated ground-truth action label is $y_{i}$. The video $\boldsymbol{x}_{i}$ contains a sequence of frames $\langle \boldsymbol{z}_{i,j}\rangle_{j=1}^T$, $\boldsymbol{z}_{i,j}\in\mathbb{R}^{C\times H\times W}$. The training set contains $N$ training video samples and is written as $\{(\boldsymbol{x}_{i},y_{i})\}_{i=1}^N$. The goal of \sysname{} is to augment a given training sample $(\boldsymbol{x}_{i},y_{i})$ into a transformed sample $(\tilde{\boldsymbol{x}}_{i},\tilde{y}_{i})$. The procedures of \sysname{} are illustrated in Figure~\ref{fig:algo} and introduced as follows.

\vspace{0.5em}\noindent\textbf{Step 1: Training the Reference Network.} 
We identify bias-inducing frames using a 2D reference network that predicts the action label from a still frame of a video. As the still frame contains no motion, we expect the network to rely on static features to make the predictions.

Specifically, at every epoch we randomly sample a frame $\boldsymbol{z}_{i,j}\in\mathbb{R}^{C\times H\times W}$ from each video $\boldsymbol{x}_{i}$, and train the reference network $\mathcal{R}(\cdot)$ to predict the label $y_i$. The loss is
\begin{equation}
    L_{ref}=\frac{1}{N}\sum_{i=1}^{N} \ell(\mathcal{R}(\boldsymbol{z}_{i,j}),y_i),
    \label{equ:algorithm-frame-classifier}
\end{equation}
where $\ell(\cdot)$ can be any classification loss, such as the cross-entropy. After training, the reference network  $\mathcal{R}(\cdot)$ encodes the correlations between static cues within the frames and the action classes.

\vspace{0.5em}\noindent\textbf{Step 2: Identifying Biased Frames.}
The output of reference network $\mathcal{R}(\bm z_{i,j})$ is a categorical distribution over action classes. We take the probability of the predicted class 
$p_{i,j} = \mathop{\max}_{k} P(y=k|\boldsymbol{z}_{i,j})$. A high $p_{i,j}$ indicates strong correlation between $\boldsymbol{z}_{i,j}$ and the action class, which means $\boldsymbol{z}_{i,j}$ can induce static bias. Therefore, we select frames with high $p_{i,j}$ to construct the biased frame bank $S$:
\begin{equation}
    S=\left\{\boldsymbol{z}_{i,j} | p_{i,j} \geq p_{\tau} \right\},
    \label{equ:algorithm-sampling}
\end{equation}
where $p_{\tau}$ is the $\tau$-th percentile value of $p_{i,j}$. In practice, we perform another round of uniformly random selection to  control the size of the biased frame bank. 

\vspace{0.5em}\noindent\textbf{Step 3: Mixing Video and Biased Frames.} To break the strong correlation between the biased frame and the action class, we mix a video of any action class with different biased frames identified above. Specifically, in each epoch, given a video sample $(\boldsymbol{x}_{i},y_{i})$, we sample a frame $\boldsymbol{z}^{\text{biased}}$ from the biased frame bank $S$ and tile it $T$ times along the temporal dimension, yielding a static video with $T$ identical frames. We denote this operation as $\text{Tile}(\boldsymbol{z}^{\text{biased}}, T)$. The augmented video sample $\tilde{\boldsymbol{x}}_{i}$ is generated by the pixel-wise interpolation of $\boldsymbol{x}_{i}$ and the static video. The augmented video label $\tilde{y}_{i}$ is the same as the original action label $y_{i}$.
\begin{equation}
    \tilde{\boldsymbol{x}}_{i} = \lambda \boldsymbol{x}_{i} + (1-\lambda) \text{Tile}(\boldsymbol{z}^{\text{biased}}, T), ~~ \tilde{y}_{i} = y_{i},
    \label{equ:algorithm-mix}
\end{equation}
where the scalar $\lambda$ is sampled from a Beta distribution $Beta(\alpha, \beta)$.

The rationale for keeping the video label unchanged after augmentation is that the static video contains no motion and does not affects the motion patterns in the mixed video, thus it should not contribute to the action label. This setting of \sysname{} can be intuitively understood as randomly permuting the labels of the static video, so that the network is forced to ignore the correlations between the static cues in the biased frames and actions.

\vspace{0.5em}\noindent\textbf{Training with Augmented Videos.}
We apply \sysname{} to each video with a predefined probability $P_{\text{aug}}$.
\begin{equation}
    (\boldsymbol{x}_{i}^{*},y_{i}^{*})=
    \begin{cases}
    (\boldsymbol{x}_{i},y_{i}) & a_i=0 \\
    (\tilde{\boldsymbol{x}}_{i},\tilde{y}_{i}) & a_i=1
    \end{cases}, 
    a_i\sim Ber(P_{\text{aug}}),
\label{equ:dynamic-aug}
\end{equation}%
where $a$ is a scalar sampled from a Bernoulli distribution $Ber(P_{\text{aug}})$. The samples $\{(\boldsymbol{x}_{i}^{*},y_{i}^{*})\}_{i=1}^{N}$ are used to train the main network $\mathcal{F}(\cdot)$ using the following loss function:
\begin{equation}
    L=\frac{1}{N}\sum_{i=1}^{N} \ell(\mathcal{F}(\boldsymbol{x}_{i}^{*}), y_i^{*}),
    \label{equ:algorithm-loss}
\end{equation}
where $\ell(\cdot)$ could be any classification loss.

\vspace{0.5em}\noindent\textbf{Discussion.} \sysname{} aims to learn robust action representations that generalize to OOD data. One popular formulation of OOD generalization \cite{ye2021towards,koyama2020invariance,sagawa2019distributionally,krueger2021out,rojas2018invariant} considers shortcut features as features that work under a specific environment but not others. For example, a classifier that excels in well-lit environments may perform terribly in dim environments. To learn robust classifiers, we ought to discover invariant features that work equally well in all environments. More formally, the optimal predictor $\mathcal{F}^*$ can be found with the bi-level optimization
\begin{equation}
    \mathcal{F}^*=\mathop{\text{argmin}}_{\mathcal{F}} \mathop{\text{max}}_{e} \mathop{\mathbb{E}}_{\boldsymbol{x}^e, y^e} [\ell(\mathcal{F}(\boldsymbol{x}^e), y^e)],
    \label{equ:ood}
\end{equation}
where the feature-label pair, $(\boldsymbol{x}^e, y^e)$, are drawn from the data distribution $P(\boldsymbol{x},y|e)$ of environment $e$ and $\ell$ is the per-sample loss. $\boldsymbol{x}^e$ contains both class features and environment features; a good predictor $\mathcal{F}$ is sensitive to the former and ignores the latter. The optimization encourages this because if $\mathcal{F}$ utilizes features that work for environment $e_1$ but not $e_2$, the loss will increase as the $\mathop{\text{max}}_{e}$ operation will select $e_2$. 

However, directly optimizing Eq. \eqref{equ:ood}, such as in \cite{sagawa2019distributionally}, requires sampling data from all environments, which is impractical due to skewed environment distributions. 
For example, videos of people \emph{playing soccer} in tuxedos on beaches are exceedingly rare. Maximizing over all environments is also challenging. 

The mixing operation in \sysname{} may be understood in the same framework. A static frame $\boldsymbol{z}^{\text{biased}}$  can be considered as coming from an environment $e^\prime$ which biases predictions toward certain action labels. Mixing $\boldsymbol{z}^{\text{biased}}$  with $\bm{x}_i$ simulates sampling  $\boldsymbol{x}^{e^\prime}$ from the environment $e^\prime$. \sysname{} may be considered to optimize the following loss,
\begin{equation}
    \mathcal{F}^*=\mathop{\text{argmin}}_{\mathcal{F}} \mathop{\mathbb{E}}_{e} \Big[  \mathop{\mathbb{E}}_{\boldsymbol{x}^e, y^e} [\ell(\mathcal{F}(\boldsymbol{x}^e), y^e)]\Big],
    \label{equ:ood-sim}
\end{equation}
which replaces the maximization over environments in Eq.~\eqref{equ:ood} with an expectation over environments due to the random sampling of $\boldsymbol{z}^{\text{biased}}$.



\section{\datasetnameA{} and \datasetnameB{}: OOD Benchmarks}
\label{sec:dataset}

To quantitatively evaluate static bias in the background and the foreground, we create OOD benchmarks based on three commonly used video datasets, \ie, HMDB51 \cite{kuehne2011hmdb}, UCF101 \cite{soomro2012ucf101} and Kinetics-400 \cite{carreira2017quo}, as detailed below.

\subsection{Foreground Masks and Background Images}
\noindent\textbf{Foreground Masks.} To extract the foreground area of actions, we use available human-annotated masks of people for UCF101 and HMDB51. There are totally 910 videos in the UCF101 test set and 256 videos in the HMDB51 test set having foreground annotations. Since there is no human-annotated masks for Kinetics-400, we use video segmentation models \cite{sun2022coarse,su2022unified} to generate foreground masks. After filtering out the videos with small foreground masks (likely to be wrong), we obtain totally 10,190 videos in the Kinetics-400 validation set to construct the benchmark.

\begin{figure}[t]
\centering
\begin{minipage}{0.31\linewidth}
    \centering
    \includegraphics[width=\textwidth,height=1.5cm]{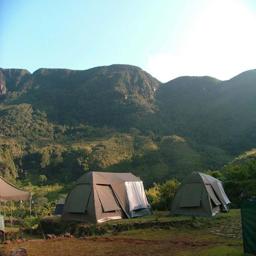}
    \subcaption[]{Place365}
    \label{fig:bg_images_place365}
\end{minipage}
\hfil
\begin{minipage}{0.31\linewidth}
    \centering
    \includegraphics[width=\textwidth,height=1.5cm]{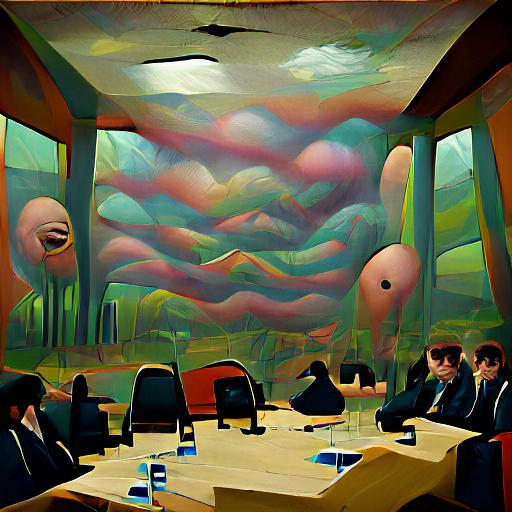}
    \subcaption[]{VQGAN-CLIP}
    \label{fig:bg_images_vqgan}
\end{minipage}
\hfil
\begin{minipage}{0.31\linewidth}
    \centering
    \includegraphics[width=\textwidth,height=1.5cm]{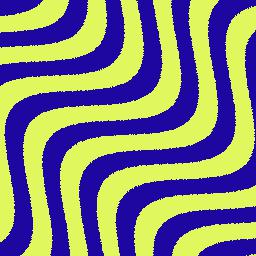}
    \subcaption[]{Sinusoid}
    \label{fig:bg_images_stripe}
\end{minipage}
\vspace{-0.5em}
\caption{Background images from different sources. (a) An image from Place365. (b) An image generated by VQGAN-CLIP from the query ``A painting of a conference room in the style of surreal art''. (c) An image of randomly generated sinusoidal stripes.}
\label{fig:bg_images}
\end{figure}

\vspace{0.5em}\noindent\textbf{Background Images.} In order to synthesize diverse test videos, we collect background images from three different image sources: 1) the test set of Place365 \cite{7968387}. 2) images generated by VQGAN-CLIP \cite{crowson2022vqgan} from a random scene category of Place365 and a random artistic style. 3) randomly generated images with S-shaped stripes defined by sinusoidal functions. For each image source, we construct a background image pool. In Figure~\ref{fig:bg_images}, we show three example background images from the three sources. More details are described in Sec. S2 of the Supplementary Material.

\subsection{Test Video Synthesis}
\noindent\textbf{Testing for Background Static Cues.} 
Given a video $\boldsymbol{x}$ with $T$ frames $\{\boldsymbol{x}_t\}_{t=1}^T$, we create a synthetic video $\hat{\boldsymbol{x}}$ by combining the foreground of $\boldsymbol{x}$ and a background image sampled from a background image pool.
\begin{equation}
    \hat{\boldsymbol{x}}_{t} = \boldsymbol{m}_{t}\odot\boldsymbol{x}_{t} + (1-\boldsymbol{m}_{t})\odot\text{Tile}(\boldsymbol{z}^{\text{bg}},T),
    \label{equ:syn-video}
\end{equation}
where $\boldsymbol{m}_{t}$ is the foreground mask, $\odot$ denotes pixel-wise multiplication, $\boldsymbol{z}^{\text{bg}}$ is a background image sampled from the image pool. $\text{Tile}(\boldsymbol{z}^{\text{bg}},T)$ repeats $\boldsymbol{z}^{\text{bg}}$ $T$ times along the temporal dimension. For each video with foreground masks, we pair it with $m$ randomly selected background images from each of the $3$ background image pools to synthesize $3m$ videos.
We set $m=10,5,1$ for HMDB51, UCF101 and Kinetics-400, respectively, since HMDB51 and UCF101 have fewer videos with foreground masks and we would like to increase the diversity of the synthetic videos.

The generated videos retain the original action foreground, including the human actors and their motion, on new random backgrounds. They are designed to test bias toward \underline{s}tatic \underline{cu}es from the \underline{ba}ckground, and are named \datasetnameA{} videos. We expect models invariant to static backgrounds to obtain high classification accuracy on  \datasetnameA{}.

\vspace{0.5em}
\noindent\textbf{Testing for Foreground Static Cues.} 
In addition, we create another set of videos to test the amount of foreground static bias in the learned representations. Foreground static cues include people and object attributes, such as bicycle helmets for \emph{cycling} and bows for \emph{archery} --- people can ride a bicycle without helmets or hold bows when not performing archery. As the \datasetnameA{} videos contain most foreground elements in the original videos, they cannot distinguish whether models rely on foreground static cues.

To this end, we create videos that contain only a single frame. Specifically, from each \datasetnameA{} video, we randomly select one frame and repeat it temporally to create a video with zero motion. As these videos quantify the representation bias toward \underline{s}tatic \underline{cu}es in the \underline{fo}reground, we name them \datasetnameB{} videos. In \datasetnameB{} videos, the foreground static features are identical to the corresponding \datasetnameA{} videos, but the motion information is totally removed. Therefore, a model invariant to foreground static features should obtain low classification accuracy on them.

We summarize the dataset statistics in Table \ref{tab:benchmark-statistics}. \datasetnameA{} and \datasetnameB{} have the same number of videos for each pair of video source and background source. We also report the domain gap between original videos to show their OOD characteristics, as explained in the next section.

\subsection{Quality Assessment}
We empirically verify that the \datasetnameA{} datasets retain the motion features of the original videos but replace background static features using the following two tests.

\vspace{0.5em}\noindent\textbf{Human Assessment.} 
To test if \datasetnameA{} preserves the motion information sufficiently for action recognition, we carry out an experiment on Amazon Mechanical Turk (AMT) to verify if human workers can recognize the actions in \datasetnameA{} videos. 

From the same original video, we randomly sampled one synthetic video and asked the AMT workers if the moving parts in the video show the labeled action. The workers are given three options: yes, no, and can't tell. We also create control questions with original videos to detect random clicking and design control groups to prevent the workers from always answering yes to synthetic videos. The final answer for each video is obtained by majority voting of three workers. Workers who do not reach at least 75\% accuracy on the control questions are rejected. More details are described in Sec. S2 of the Supplementary Material.

Collectively, the AMT workers were able to correctly recognize the actions in 96.15\% of UCF101-\datasetnameA{}, 86.33\% of HMDB51-\datasetnameA{} and 85.19\% of Kinetics400-\datasetnameA{} videos. We conclude that \datasetnameA{} videos preserve sufficient action information for humans to recognize.

\begin{table}[tb]
\setlength\tabcolsep{1.5pt}
\setlength\abovecaptionskip{-0.5em}
\caption{Statistics of the created benchmarks.}
\label{tab:benchmark-statistics}
\begin{center}
\scriptsize
\begin{tabular}{@{}lcccccc@{}}
\toprule
\begin{tabular}[x]{@{}c@{}} Video\\Source\end{tabular} & \begin{tabular}[x]{@{}c@{}}\# Original\\Videos\end{tabular} & \begin{tabular}[x]{@{}c@{}}Background\\Source\end{tabular} & \begin{tabular}[x]{@{}c@{}}\# Synthetic\\Videos\end{tabular} & \begin{tabular}[x]{@{}c@{}}\# Domain Gap\\ of \datasetnameA \end{tabular}  & \begin{tabular}[x]{@{}c@{}}\# Domain Gap\\ of \datasetnameB \end{tabular}   \\
\midrule
\multirow{3}{*}{HMDB51} & \multirow{3}{*}{256} & Place365 & 2,560 & 5.646$\pm$0.276 & 5.745$\pm$0.290\\
    ~ & ~ & VQGAN-CLIP & 2,560 & 8.178$\pm$0.685 & 8.307$\pm$0.533 \\
    ~ & ~ & Sinusoid & 2,560 & 11.739$\pm$0.444 & 11.998$\pm$0.932 \\
\midrule
\multirow{3}{*}{UCF101} & \multirow{3}{*}{910} & Place365 & 4,550 & 20.493$\pm$2.199 & 20.829$\pm$2.093 \\
    ~ & ~ & VQGAN-CLIP & 4,550 & 51.320$\pm$5.790 & 55.202$\pm$9.477 \\
    ~ & ~ & Sinusoid & 4,550 & 52.249$\pm$3.522 & 52.534$\pm$5.930 \\
\midrule
\multirow{3}{*}{Kinetics-400} & \multirow{3}{*}{10,190} & Place365 & 10,190 & 6.094$\pm$0.208 & 6.455$\pm$0.224 \\
    ~ & ~ & VQGAN-CLIP & 10,190 & 7.504$\pm$0.296 & 8.052$\pm$0.273 \\
    ~ & ~ & Sinusoid & 10,190 & 7.211$\pm$0.311 & 7.766$\pm$0.148 \\
\bottomrule
\end{tabular}
\end{center}
\end{table}

\vspace{0.5em}\noindent\textbf{Domain Gaps of the Static Features.} To verify if \datasetnameA{} and \datasetnameB{} have successfully replaced the background static features and qualify as OOD test sets, we test if a classifier based on purely static features trained on IID videos can generalize to \datasetnameA{} and \datasetnameB{}.

Using a variation of scene representation bias \cite{choi2019can}, we define the domain gap $G_{scene}$ as
\begin{small}
\vspace{-0.2em}
\begin{equation}
G_{scene}=Acc(D_{\text{ori}},\Phi_{scene})/Acc(D_{\text{syn}},\Phi_{scene}).
\end{equation}
\end{small}%
Here $\Phi_{scene}$ is the average frame feature extracted from a ResNet-50 pretrained on Place365 \cite{7968387}. Thus, the extracted feature captures static scene information, mostly from the background. We train a linear classifier on the original video training set and apply it to the original test set $D_{\text{ori}}$, obtaining the accuracy $Acc(D_{\text{ori}},\Phi_{scene})$. After that, we apply the same classifier to the synthetic dataset $D_{\text{syn}}$, obtaining the accuracy $Acc(D_{\text{syn}},\Phi_{scene})$. A higher ratio indicates greater domain gap with respect to static features. 

In Table \ref{tab:benchmark-statistics}, we show the means and standard deviations computed from three random repeats of video synthesis. We observe large domain gaps, ranging from 5.6-fold to 52-fold decrease in accuracy on the synthetic test set. This demonstrates the static features of synthetic videos differ substantially from the original videos and the synthetic videos can serve as OOD tests. Moreover, the low standard deviations show that the effects of random sampling are marginal. In later experiments, we use the dataset from one random seed.

\section{Experiments}
\label{sec:exp}
In this section, we compare the performance of several mainstream action recognition methods on IID and OOD test data and validate the effectiveness of \sysname{}.

\subsection{Comparing Methods}
\noindent\textbf{Action Recognition Models.} (1) TSM \cite{lin2019tsm}, a temporal shift module learning spatiotemporal features with 2D CNN. (2) SlowFast \cite{feichtenhofer2019slowfast}, a two-branch 3D CNN learning spatiotemporal signals under two frame rates. (3) Video Swin Transformer \cite{liu2021video}, an adapted Swin Transformer \cite{liu2021swin} for videos. We use the tiny version, denoted as Swin-T.

\vspace{0.2em}\noindent\textbf{Video Data Augmentation and Debiasing Methods.} We compare the debiasing performance of several video data augmentation and debiasing methods by adapting them to supervised action recognition. (1) Mixup \cite{zhang2017mixup} and VideoMix \cite{yun2020videomix}. (2) SDN \cite{choi2019can}. (3) BE \cite{wang2021removing}, ActorCutMix\cite{zou2021learning} and FAME \cite{Ding_2022_CVPR}. We adapt these three self-supervised debiasing methods as data augmentations, which carve out the foreground and replace the background as in the original papers. All the data augmentation techniques are applied stochastically as in \cite{gontijo2020tradeoffs}. More implementation details are described in Sec. S3 of the Supplementary Material.

\subsection{Evaluation Metrics}
We conduct the following experiments on Kinetics-400, UCF101 and HMDB51.  First, we perform IID tests on the original test sets and use the top-1 accuracy as metrics. After that, we perform OOD tests on \datasetnameA{} and \datasetnameB{} and report the average top-1 accuracy across background image sources. Note that higher accuracy on \datasetnameA{} is better (low background static bias), while lower accuracy on \datasetnameB{} is better (low foreground static bias). 

To show the performance of utilizing pure foreground motion information, we propose another performance metric called contrasted accuracy (Contra. Acc.). As one \datasetnameB{} video is derived from a \datasetnameA{} video, we count one correct prediction if the model is correct on the \datasetnameA{} but incorrect on the associated \datasetnameB{} video.

We further evaluate on the synthetic videos with conflicting foreground cues (Figure \ref{fig:bg_fg_bias}). Finally, we also evaluate on ARAS~\cite{duan2022mitigating}, a real-world OOD dataset with rare scenes, to show the performance of scene bias reduction.

\begin{table}[t]
\begin{center}
\caption{IID and OOD test accuracy (\%) of augmentation and debiasing methods on Kinetics-400. $\dagger$ indicates adaptation from self-supervised debiasing methods. Confl-FG denotes synthetic videos with conflicting foreground cues. All models are pretrained on ImageNet.}
\label{tab:res_aug_k400}
\footnotesize
\setlength\tabcolsep{0.5pt}
\begin{tabular}{@{}clp{0.01cm}cccccc@{}}
\toprule
\multirow{3}{*}{Model} & \multirow{3}{*}{\begin{tabular}[x]{@{}l@{}} Augmentation\\or Debiasing\end{tabular}} && \multirow{3}{*}{IID} & \multicolumn{5}{c}{OOD} \\
\cmidrule{5-9}
~ & ~ && ~ & \multirow{2}{*}{\begin{tabular}[x]{@{}c@{}} Avg\\\datasetnameA{}\end{tabular}$\big\uparrow$} & \multirow{2}{*}{\begin{tabular}[x]{@{}c@{}} Avg\\\datasetnameB{}\end{tabular}$\big\downarrow$} & \multirow{2}{*}{\begin{tabular}[x]{@{}c@{}} Contra.\\Acc.\end{tabular}$\big\uparrow$} & \multirow{2}{*}{\begin{tabular}[x]{@{}c@{}} Confl-\\FG\end{tabular}$\big\uparrow$} & \multirow{2}{*}{ARAS$\big\uparrow$} \\ \\
\midrule
\multirow{8}{*}{TSM} & No && 71.13 & 37.39 & 17.22 & 22.80 & 20.15 & 57.86 \\
~ & Mixup && 71.33 & 40.81 & 17.53 & 25.98 & 23.48 & 58.05 \\
~ & VideoMix && \textbf{71.35} & 38.87 & 17.25 & 24.57 & 23.43 & 56.61 \\
~ & SDN && 69.99 & 36.95 & 16.55 & 22.38 & 20.29 & 55.06 \\
~ & BE$\dagger$ && 71.30 & 37.89 & 16.08 & 24.35 & 20.11 & 57.47 \\
~ & ActorCutMix$\dagger$ && 71.07 & 40.42 & 16.29 & 26.52 & 21.41 & 57.09 \\
~ & FAME$\dagger$ && 71.13 & \textbf{40.91} & 18.34 & 25.63 & 24.41 & 57.47 \\
~ & \cellcolor{lightgray}\sysname{} (Ours) &\cellcolor{lightgray}& \cellcolor{lightgray}71.28 & \cellcolor{lightgray}40.48 & \cellcolor{lightgray}\textbf{5.23} & \cellcolor{lightgray}\textbf{36.07} & \cellcolor{lightgray}\textbf{25.73} & \cellcolor{lightgray}\textbf{59.69} \\
\midrule
\multirow{8}{*}{Swin-T} & No && 73.95 & 41.74 & 18.17 & 25.93 & 25.25 & 60.17 \\
~ & Mixup && 73.91 & 43.95 & 17.92 & 28.24 & 27.64 & 59.59 \\
~ & VideoMix && 73.80 & 43.17 & 19.26 & 26.40 & 29.37 & 60.95 \\
~ & SDN && 72.23 & 42.34 & 21.46 & 24.46 & 27.14 & 60.26 \\
~ & BE$\dagger$ && 73.93 & 43.40 & 19.56 & 26.28 & 26.67 & 59.79 \\
~ & ActorCutMix$\dagger$ && \textbf{73.97} & 45.70 & 19.39 & 28.64 & 29.02 & 61.23 \\
~ & FAME$\dagger$ && 73.81 & \textbf{48.79} & 21.27 & 30.03 & 29.50 & 60.37 \\
~ & \cellcolor{lightgray}\sysname{} (Ours) &\cellcolor{lightgray}& \cellcolor{lightgray}73.86 & \cellcolor{lightgray}44.10 & \cellcolor{lightgray}\textbf{5.51} & \cellcolor{lightgray}\textbf{39.41} & \cellcolor{lightgray}\textbf{30.77} & \cellcolor{lightgray}\textbf{62.49} \\
\bottomrule
\end{tabular}
\end{center}
\vspace{-1em}
\end{table}

\begin{table}[t]
\begin{center}
\caption{IID and OOD test accuracy (\%) of augmentation and debiasing methods on HMDB51. All models are pretrained on Kinetics-400.}
\label{tab:res_aug_hmdb51}
\scriptsize
\setlength\tabcolsep{0.8pt}
\begin{tabular}{@{}clp{0.005cm}ccccc@{}}
\toprule
\multirow{3}{*}{Model} & \multirow{3}{*}{\begin{tabular}[x]{@{}l@{}} Augmentation\\or Debiasing\end{tabular}} && \multirow{3}{*}{IID} & \multicolumn{4}{c}{OOD} \\
\cmidrule{5-8}
~ & ~ && ~ & \multirow{2}{*}{\begin{tabular}[x]{@{}c@{}} Avg\\\datasetnameA{}\end{tabular}$\big\uparrow$} & \multirow{2}{*}{\begin{tabular}[x]{@{}c@{}} Avg\\\datasetnameB{}\end{tabular}$\big\downarrow$} & \multirow{2}{*}{\begin{tabular}[x]{@{}c@{}} Contra.\\Acc.\end{tabular}$\big\uparrow$} & \multirow{2}{*}{\begin{tabular}[x]{@{}c@{}} Confl-\\FG\end{tabular}$\big\uparrow$} \\ \\
\midrule
\multirow{8}{*}{TSM} & No && 70.39$\pm$0.51 & 38.03$\pm$1.39 & 19.23$\pm$1.30 & 22.02$\pm$0.64 & 25.44$\pm$1.31 \\
~ & Mixup && \textbf{72.00$\pm$0.47} & 39.76$\pm$1.72 & 19.08$\pm$1.37 & 23.76$\pm$0.84 & 26.94$\pm$1.23 \\
~ & VideoMix && 70.72$\pm$0.12 & 35.71$\pm$1.57 & 17.48$\pm$1.11 & 21.03$\pm$0.55 & 22.19$\pm$1.47 \\
~ & SDN && 69.51$\pm$0.30 & 37.05$\pm$0.73 & 17.60$\pm$0.37 & 23.74$\pm$0.95 & 28.38$\pm$0.87 \\
~ & BE && 71.22$\pm$0.24 & 38.48$\pm$1.42 & 19.45$\pm$1.06 & 22.39$\pm$0.67 & 25.21$\pm$1.35 \\
~ & ActorCutMix && 70.52$\pm$0.82 & 38.40$\pm$0.53 & 19.61$\pm$0.56 & 21.94$\pm$0.40 & 26.16$\pm$0.36 \\
~ & FAME && 70.39$\pm$0.88 & 47.19$\pm$1.52 & 22.33$\pm$0.91 & 28.21$\pm$0.89 & 33.98$\pm$2.09 \\
~ & \cellcolor{lightgray}\sysname{} &\cellcolor{lightgray}& \cellcolor{lightgray}71.52$\pm$0.38 & \cellcolor{lightgray}\textbf{48.23$\pm$0.96} & \cellcolor{lightgray}\textbf{8.43$\pm$0.88} & \cellcolor{lightgray}\textbf{42.05$\pm$0.99} & \cellcolor{lightgray}\textbf{36.89$\pm$1.09} \\
\midrule
\multirow{8}{*}{Swin-T} & No && 73.92$\pm$0.74 & 43.93$\pm$0.78 & 20.46$\pm$0.71 & 27.84$\pm$1.28 & 36.58$\pm$1.65 \\
~ & Mixup && 74.58$\pm$0.43 & 43.10$\pm$1.13 & 21.17$\pm$0.66 & 26.09$\pm$1.05 & 36.62$\pm$2.98 \\
~ & VideoMix && 73.31$\pm$0.53 & 39.39$\pm$0.71 & 20.44$\pm$0.73 & 23.13$\pm$0.54 & 32.68$\pm$1.04 \\
~ & SDN && 74.66$\pm$0.82 & 40.02$\pm$1.48 & 20.22$\pm$1.24 & 22.88$\pm$1.05 & 34.87$\pm$2.43 \\
~ & BE && 74.31$\pm$0.41 & 43.56$\pm$1.38 & 19.96$\pm$0.71 & 27.84$\pm$1.32 & 35.99$\pm$0.67 \\
~ & ActorCutMix && 74.05$\pm$0.60 & 46.79$\pm$1.38 & 22.07$\pm$0.36 & 28.12$\pm$1.27 & 36.97$\pm$1.63 \\
~ & FAME && 73.79$\pm$0.29 & 51.40$\pm$1.54 & 26.92$\pm$0.71 & 29.66$\pm$2.11 & 39.61$\pm$1.87 \\
~ & \cellcolor{lightgray}\sysname{} &\cellcolor{lightgray}& \cellcolor{lightgray}\textbf{74.82$\pm$0.43} & \cellcolor{lightgray}\textbf{51.81$\pm$1.78} & \cellcolor{lightgray}\textbf{13.39$\pm$0.71} & \cellcolor{lightgray}\textbf{40.28$\pm$1.61} & \cellcolor{lightgray}\textbf{47.38$\pm$1.50} \\
\bottomrule
\end{tabular}
\end{center}
\end{table}

\begin{table}[t]
\begin{center}
\caption{IID and OOD test accuracy (\%) of augmentation and debiasing methods on UCF101. All models are pretrained on Kinetics-400.}
\label{tab:res_aug_ucf101}
\scriptsize
\setlength\tabcolsep{0.8pt}
\begin{tabular}{@{}clp{0.005cm}ccccc@{}}
\toprule
\multirow{3}{*}{Model} & \multirow{3}{*}{\begin{tabular}[x]{@{}l@{}} Augmentation\\or Debiasing\end{tabular}} && \multirow{3}{*}{IID} & \multicolumn{4}{c}{OOD} \\
\cmidrule{5-8}
~ & ~ && ~ & \multirow{2}{*}{\begin{tabular}[x]{@{}c@{}} Avg\\\datasetnameA{}\end{tabular}$\big\uparrow$} & \multirow{2}{*}{\begin{tabular}[x]{@{}c@{}} Avg\\\datasetnameB{}\end{tabular}$\big\downarrow$} & \multirow{2}{*}{\begin{tabular}[x]{@{}c@{}} Contra.\\Acc.\end{tabular}$\big\uparrow$} & \multirow{2}{*}{\begin{tabular}[x]{@{}c@{}} Confl-\\FG\end{tabular}$\big\uparrow$} \\ \\
\midrule
\multirow{8}{*}{TSM} & No && 94.62$\pm$0.08 & 25.60$\pm$1.36 & 4.21$\pm$0.84 & 21.83$\pm$1.48 & 27.68$\pm$1.35 \\
~ & Mixup && \textbf{94.71$\pm$0.14} & 27.80$\pm$0.95 & 4.04$\pm$0.81 & 24.17$\pm$1.00 & 30.31$\pm$1.10 \\
~ & VideoMix && 94.50$\pm$0.19 & 31.55$\pm$1.68 & 5.77$\pm$0.74 & 26.69$\pm$1.38 & 30.69$\pm$1.79 \\
~ & SDN && 93.84$\pm$0.27 & 19.91$\pm$0.61 & 3.10$\pm$0.19 & 17.19$\pm$0.51 & 20.89$\pm$0.36 \\
~ & BE && 94.49$\pm$0.14 & 25.91$\pm$1.37 & 4.62$\pm$0.84 & 21.82$\pm$1.38 & 28.06$\pm$1.32 \\
~ & ActorCutMix && 94.47$\pm$0.15 & \textbf{38.11$\pm$1.48} & 4.56$\pm$0.16 & 33.90$\pm$1.51 & 38.12$\pm$2.12 \\
~ & FAME && 93.72$\pm$0.09 & 35.72$\pm$1.15 & 3.67$\pm$0.52 & 32.28$\pm$1.28 & 34.58$\pm$0.93 \\
~ & \cellcolor{lightgray}\sysname{} &\cellcolor{lightgray}& \cellcolor{lightgray}94.30$\pm$0.14 & \cellcolor{lightgray}37.18$\pm$1.29 & \cellcolor{lightgray}\textbf{0.79$\pm$0.12} & \cellcolor{lightgray}\textbf{36.47$\pm$1.24} & \cellcolor{lightgray}\textbf{40.59$\pm$0.80} \\
\midrule
\multirow{8}{*}{Swin-T} & No && \textbf{96.21$\pm$0.19} & 42.31$\pm$2.24 & 5.78$\pm$0.68 & 36.82$\pm$2.12 & 44.65$\pm$2.10 \\
~ & Mixup && 96.17$\pm$0.14 & 46.16$\pm$1.74 & 5.93$\pm$0.43 & 40.46$\pm$1.96 & 47.16$\pm$2.82 \\
~ & VideoMix && 96.00$\pm$0.02 & 41.40$\pm$1.11 & 13.27$\pm$0.85 & 29.37$\pm$0.91 & 42.59$\pm$1.51 \\
~ & SDN && 95.76$\pm$0.11 & 39.25$\pm$2.32 & 2.98$\pm$0.88 & 36.42$\pm$1.74 & 48.47$\pm$2.06 \\
~ & BE && 96.06$\pm$0.11 & 43.98$\pm$0.80 & 5.54$\pm$0.94 & 38.62$\pm$1.13 & 46.62$\pm$0.96 \\
~ & ActorCutMix && 95.87$\pm$0.19 & \textbf{58.61$\pm$0.48} & 11.92$\pm$0.25 & 46.87$\pm$0.45 & 56.88$\pm$0.39 \\
~ & FAME && 95.81$\pm$0.15 & 40.90$\pm$1.57 & 6.36$\pm$0.71 & 35.14$\pm$1.66 & 28.21$\pm$1.83 \\
~ & \cellcolor{lightgray}\sysname{} &\cellcolor{lightgray}& \cellcolor{lightgray}96.02$\pm$0.08 & \cellcolor{lightgray}58.22$\pm$0.41 & \cellcolor{lightgray}\textbf{3.44$\pm$0.51} & \cellcolor{lightgray}\textbf{54.90$\pm$0.77} & \cellcolor{lightgray}\textbf{57.30$\pm$0.60} \\
\bottomrule
\end{tabular}
\end{center}
\vspace{-1em}
\end{table}

\subsection{Results on IID and OOD Benchmarks}
Table \ref{tab:res_aug_k400}, \ref{tab:res_aug_hmdb51} and \ref{tab:res_aug_ucf101} compare the IID and OOD performance of different video data augmentation and debiasing methods on Kinetics-400, HMDB51 and UCF101. Given limited computational resources, we ran experiments on Kinetics-400 using a single seed. However, on the smaller HMDB51 and UCF101, we repeated experiments with three seeds. In Sec. S1 of the Supplementary Material, we provide more detailed results (\eg, tests on videos with conflicting
foreground cues and ARAS~\cite{duan2022mitigating}).

\vspace{0.2em}\noindent\textbf{OOD data cause performance degradation.} Comparing the performance of TSM and Swin-T on IID and OOD tests, we observe that they perform much worse (more than 20\%) on \datasetnameA{} than IID videos. Given that human workers can recognize the action in more than 85\% of \datasetnameA{} videos, the results indicate that the models are not robust to the domain shifts, probably due to the reliance of static background features; when the backgrounds are replaced, performance deterioration ensues.

\vspace{0.2em}\noindent\textbf{IID tests do not fully reveal representation quality.} Comparing the performance of different augmentation and debiasing methods, we observe that all methods obtain similar accuracies on IID tests, which fall within a 2\% band. However, they show vastly different performance on \datasetnameA{} and \datasetnameB{} --- the maximum difference is larger than 15\%. Therefore, we argue that IID tests alone may not be good indicators of the robustness of action representations.

In particular, VideoMix, SDN and BE provide little debiasing effects. Mixup leads to inconsistent performance gains. ActorCutMix and FAME consistently improve performance on \datasetnameA{}. Nevertheless, they decrease performance (increase accuracy) on \datasetnameB{}, which suggests that they improve performance on \datasetnameA{} partially by increasing reliance on foreground static features. The action features learned with ActorCutMix and FAME are likely still vulnerable to foreground static bias.

\vspace{0.5em}\noindent\textbf{\sysname{} alleviates foreground and background static bias.} \sysname{} boosts the performance on both \datasetnameA{} and \datasetnameB{} videos and obtains the best contrasted accuracy (Contra. Acc.). The significant improvements on \datasetnameB{} indicate that \sysname{} can suppress static bias from the entire video and not only the background. In addition, \sysname{} outperforms other methods on videos with conflicting foreground cues as well as ARAS. Overall, these results demonstrate the ability of \sysname{} to reduce static bias that is difficult to exhaustively name or pixel-wise cut out.




\subsection{\sysname{} Improves Representation Learning}
We further investigate the effects of \sysname{} on improving representation learning through the following tests.

\vspace{0.5em}\noindent\textbf{Transferring action features across datasets.} We evaluate the representations learned with different augmentation and debiasing methods by their capability to transfer to different datasets. We adopt the linear probing protocol, which trains a linear classifier on the target dataset on top of the backbone network trained on the source dataset. Table~\ref{tab:transfer} shows the results of TSM, where \sysname{} obtains the best performance, especially in transferring across small datasets.

\vspace{0.5em}\noindent\textbf{Downstream weakly supervised action localization.} We evaluate the representations learned with \sysname{} by their ability to improve downstream weakly supervised action localization. We pretrain TSM on Kinetics-400 with \sysname{}. After that, we extract RGB features for each video segments on THUMOS14 \cite{idrees2017thumos} and use the extracted features to train weakly supervised action localization models BaSNet \cite{lee2020background} and CoLA \cite{zhang2021cola}. \sysname{} improves the performance by more than 1.0\% of average mAP for BaSNet and more than 0.5\% of average mAP for CoLA.

\begin{table}[tb]
\begin{center}
\caption{Action recognition accuracy (\%) of transferring features across Kinetics-400, UCF101, and HMDB51.}
\label{tab:transfer}
\footnotesize
\setlength\tabcolsep{1pt}
\begin{tabular}{@{}lp{0.01cm}cccc@{}}
\toprule
\multirow{2}{*}{\begin{tabular}[x]{@{}l@{}} Augmentation\\or Debiasing\end{tabular}} && \multicolumn{4}{c}{Source$\rightarrow$Target} \\
\cmidrule{3-6}
~ && K400$\rightarrow$UCF & K400$\rightarrow$HMDB & HMDB$\rightarrow$UCF & UCF$\rightarrow$HMDB \\
\midrule
No && 92.52 & 66.67 & 61.64 & 44.95 \\
Mixup && 93.07 & 68.69 & 63.58 & 46.60 \\
VideoMix && 93.55 & 69.22 & 61.49 & 40.33 \\
SDN && 92.81 & 63.79 & 61.12 & 41.90 \\
BE && 93.10 & 67.45 & 62.71 & 45.88 \\
ActorCutMix && 92.73 & 67.39 & 61.67 & 42.92 \\
FAME && 93.87 & 67.84 & 58.87 & 44.99 \\
\sysname{} && \textbf{93.89} & \textbf{70.07} & \textbf{65.69} & \textbf{47.99} \\
\bottomrule
\end{tabular}
\end{center}
\end{table}

\begin{table}[tb]
\begin{center}
\caption{Weakly supervised action localization performance of features learned by \sysname{}.}
\label{tab:loc}
\footnotesize
\setlength\tabcolsep{4pt}
\begin{tabular}{@{}cccc@{}}
\toprule
Method & Feature & Debiasing & Avg mAP@IoU=[0.1:0.9] \\
\midrule
\multirow{2}{*}{BaSNet} & TSM (RGB) & No & 0.1810 \\
~  & TSM (RGB) & \sysname{} & \textbf{0.1935} \\
\midrule
\multirow{2}{*}{CoLA} & TSM (RGB) & No & 0.2380 \\
~  & TSM (RGB) & \sysname{} & \textbf{0.2436} \\
\bottomrule
\end{tabular}
\end{center}
\vspace{-1em}
\end{table}

\subsection{Ablation Study}
\label{subsec:ablation}
We conduct ablation study on UCF101 and HMDB51 to examine design choices of \sysname{}. 

\vspace{0.5em}\noindent\textbf{Debiasing works the best when the reference network and the main network share the same architecture.} We compare the results of \sysname{} with different network structures in Table~\ref{tab:reference_structure}. When the structures of the reference network and the main network are identical, the OOD performance is the best and the IID performance is very close to the best, indicating good bias mitigation. We hypothesize that networks with same architecture tend to learn the same bias. As a result, using a reference network with the same architecture as the main network could be the most effective at identifying bias-inducing frames. 


\vspace{0.5em}\noindent\textbf{Sampling biased frames improves debiasing.} We compare three frame sampling strategies when constructing the biased frame bank: (1) \emph{No RefNet}: the frame bank is uniformly sampled from the whole dataset; (2) \emph{RefNet}: as in \sysname{}, we sample frames with high prediction probabilities from the reference network according to Eq.~\eqref{equ:algorithm-sampling}; (3)~\emph{RefNet Inversed}: contrary to \sysname{}, we sample frames with low prediction probabilities from the reference network, $S=\left\{\boldsymbol{z}_{i,j} | p_{i,j} < p_{\tau} \right\}$. Table~\ref{tab:frame_sampling} shows results of ImageNet pretrained TSM and Swin-T. The reference network (RefNet)  approach  achieves the best OOD performance, whereas RefNet Inversed performs the worst. 

We observe the difference between RefNet and No RefNet is small on UCF101 but is large on HMDB51. We attribute this to the prevalence of bias-inducing frames in UCF101. MMAction2~\cite{2020mmaction2} trained TSN~\cite{8454294} using only three frames per video on UCF101 and achieved 83.03\% classification accuracy but achieved only 48.95\% with 8 frames on HMDB51\footnote{\url{https://github.com/open-mmlab/mmaction2/blob/02a06bb3180e951b00ccceb48dab055f95acd1a7/configs/recognition/tsn/README.md}}. This shows many frames in UCF101 contain static cues correlated with the class labels. Random sampling can yield many bias-inducing frames on UCF101 but cannot do so on HMDB51, where the strength of RefNet becomes apparent. 

In Sec. S1 of the Supplementary Material, we provide more ablation studies showing that mixing action labels in \sysname{} decreases performance and sufficient mixing strength (\ie, small values of $\lambda$ in Eq. \eqref{equ:algorithm-mix}) is necessary for debiasing.

\begin{table}[tb]
\begin{center}
\caption{Action recognition accuracy (\%) of \sysname{} with different reference network structures. All networks are pretrained on ImageNet.}
\label{tab:reference_structure}
\footnotesize
\setlength\tabcolsep{4pt}
\begin{tabular}{@{}ccp{0.01cm}ccp{0.01cm}cc@{}}
\toprule
\multirow{3}{*}{\begin{tabular}[x]{@{}c@{}}Main\\Network\end{tabular}} & \multirow{3}{*}{\begin{tabular}[x]{@{}c@{}}Reference\\Network\end{tabular}} && \multicolumn{2}{c}{UCF101} && \multicolumn{2}{c}{HMDB51} \\
\cmidrule{4-5}\cmidrule{7-8}
~ & ~ && \multirow{2}{*}{IID} & \multirow{2}{*}{\begin{tabular}[x]{@{}c@{}} Contra.\\Acc.\end{tabular}} && \multirow{2}{*}{IID} & \multirow{2}{*}{\begin{tabular}[x]{@{}c@{}} Contra.\\Acc.\end{tabular}} \\ \\
\midrule
\multirow{3}{*}{TSM} & ResNet50-2D && 87.29 & \textbf{24.60} && 54.66 & \textbf{33.14} \\
~ & SlowFast-2D && 87.44 & 22.20 && 55.03 & 30.51 \\
~ & Swin-T-2D && 86.72 & 23.08 && 55.05 & 31.62 \\
\midrule
\multirow{3}{*}{SlowFast} & ResNet50-2D && 84.85 & 18.86 && 50.74 & 20.89 \\
~ & SlowFast-2D && 84.96 & \textbf{19.76} && 51.53 & \textbf{21.21} \\
~ & Swin-T-2D && 85.16 & 19.18 && 51.85 & 20.28 \\
\midrule
\multirow{3}{*}{Swin-T} & ResNet50-2D && 88.59 & 31.09 && 56.10 & 18.44 \\
~ & SlowFast-2D && 88.60 & 29.34 && 54.43 & 19.25 \\
~ & Swin-T-2D && 88.92 & \textbf{32.14} && 55.36 & \textbf{21.40} \\
\bottomrule
\end{tabular}
\end{center}
\end{table}

\begin{table}[tb]
\begin{center}
\caption{Action recognition accuracy (\%) of \sysname{} with different frame sampling strategies.}
\label{tab:frame_sampling}
\footnotesize
\setlength\tabcolsep{4pt}
\begin{tabular}{@{}ccp{0.01cm}ccp{0.01cm}cc@{}}
\toprule
\multirow{3}{*}{\begin{tabular}[x]{@{}c@{}}Main\\Network\end{tabular}} & \multirow{3}{*}{\begin{tabular}[x]{@{}c@{}}Sampling\\Strategy\end{tabular}} && \multicolumn{2}{c}{UCF101} && \multicolumn{2}{c}{HMDB51} \\
\cmidrule{4-5}\cmidrule{7-8}
~ & ~ && \multirow{2}{*}{IID} & \multirow{2}{*}{\begin{tabular}[x]{@{}c@{}}Contra.\\Acc.\end{tabular}} && \multirow{2}{*}{IID} & \multirow{2}{*}{\begin{tabular}[x]{@{}c@{}} Contra.\\Acc.\end{tabular}} \\ \\
\midrule
\multirow{3}{*}{TSM} & No RefNet && 87.39 & 24.49 && 54.07 & 31.21 \\
~ & RefNet && 87.29 & \textbf{24.60} && 54.66 & \textbf{33.14} \\
~ & RefNet Inversed && 87.38 & 23.53 && 54.79 & 29.17 \\
\midrule
\multirow{3}{*}{SlowFast} & No RefNet && 85.03 & 18.98 && 51.79 & 20.94 \\
~ & RefNet && 84.96 & \textbf{19.76} && 51.53 & \textbf{21.21} \\
~ & RefNet Inversed && 84.33 & 18.77 && 50.94 & 18.61 \\
\midrule
\multirow{3}{*}{Swin-T} & No RefNet && 88.37 & 31.24 && 55.62 & 18.89 \\
~ & RefNet && 88.92 & \textbf{32.14} && 55.36 & \textbf{21.40} \\
~ & RefNet Inversed && 88.59 & 30.51 && 56.34 & 18.18 \\
\bottomrule
\end{tabular}
\end{center}
\vspace{-1em}
\end{table}

\subsection{Performance on Something-Something-V2}
To validate the effectiveness of different debiasing methods on recognizing fine-grained actions with strong temporal structures, we perform tests on Something-Something-V2 \cite{goyal2017something}. In Table \ref{tab:ssv2}, we show the performance of different debiasing methods with TSM as the base model. Since SDN and ActorCutMix require bounding boxes of human, which are time-consuming to extract, we did not include the results of these two methods. The results show that \sysname{} outperforms other data augmentation methods, illustrating its effectiveness on fine-grained action videos.

\begin{table}[tb]
\begin{center}
\caption{Action recognition accuracy (\%) of different debiasing methods on Something-Something-V2.}
\label{tab:ssv2}
\setlength\tabcolsep{6pt}
\begin{tabular}{@{}lc@{}}
\toprule
~~Debiasing~~ & ~~Accuracy~~ \\
\midrule
~~No & 57.49 \\
~~Mixup & 57.86 \\
~~VideoMix & 58.23 \\
~~BE & 57.68 \\
~~FAME & 58.10 \\
~~\sysname{} (Ours) & \textbf{58.68} \\
\bottomrule
\end{tabular}
\end{center}
\vspace{-1em}
\end{table}

\section{Conclusion and Discussion}
To learn robust and generalizable action representations, we explore techniques that mitigate static bias in both the background and the foreground. We propose a simple yet effective video data augmentation method, \sysname{}, and create two new sets of OOD benchmarks, \datasetnameA{} and \datasetnameB{}, to quantify static bias in the background and the foreground. Through extensive evaluation, we conclude that \sysname{} mitigates static bias in the background and the foreground and improves the performance of transferring learning and downstream tasks. In contrast, existing debiasing methods remain vulnerable to foreground static bias despite their robustness to background static bias. 

Despite the strengths of \sysname{} on mitigating static bias in the background and the foreground, it has the following limitations: (1) additional computational overhead in training the reference network (about 8\% of the training time of the main network); and (2) little improvement (and little degradation) on IID tests. 

For future work, we believe that evaluating static bias in large pretrained models with the created benchmarks and adapting \sysname{} to mitigate static bias in such models would be promising directions.

\section*{Acknowledgments}
This work has been supported by the Nanyang Associate Professorship and the National Research Foundation Fellowship (NRF-NRFF13-2021-0006), Singapore. The computational work for this article was partially performed on resources of the National Supercomputing Centre, Singapore (\url{https://www.nscc.sg}). Any opinions, findings, conclusions, or recommendations expressed in this material are those of the authors and do not reflect the views of the funding agencies.

\clearpage
\appendix
\titlecontents{section}[0em]{\smallskip}{\bfseries\thecontentslabel\hspace{1em}}{\hspace*{3}}{\,\,\titlerule*[0.6pc]{.}\contentspage}
\titlecontents{subsection}[1em]{\smallskip}{\thecontentslabel\hspace{1em}}{\hspace*{3}}{\,\,\titlerule*[0.6pc]{.}\contentspage}

\renewcommand{\thesection}{S\arabic{section}}
\renewcommand{\thetable}{S\arabic{table}}
\renewcommand{\thefigure}{S\arabic{figure}}
\setcounter{section}{0}
\setcounter{table}{0}
\setcounter{figure}{0}

\ificcvfinal\thispagestyle{empty}\fi


\begin{center}
\textbf{\Large Supplementary Material}
\end{center}
\vspace{1em}

{
\hypersetup{linkcolor=black}
\startcontents[sections]
\printcontents[sections]{l}{1}{\setcounter{tocdepth}{2}}
}

\begin{figure}[!h]
\centering
\includegraphics[width=0.23\linewidth,height=0.38\linewidth]{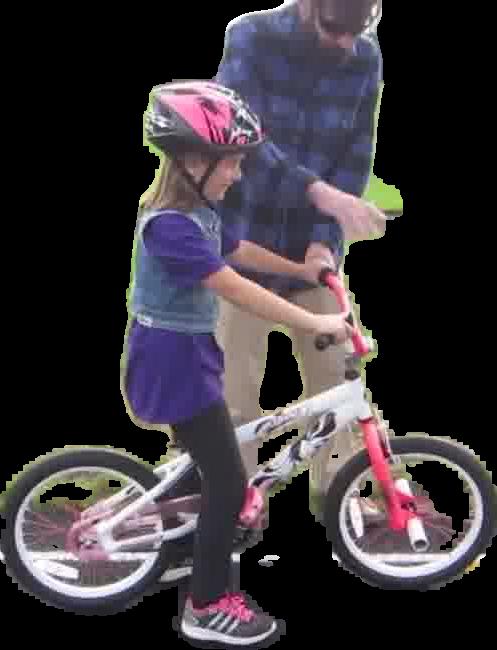}
\includegraphics[width=0.23\linewidth,height=0.38\linewidth]{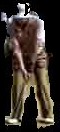}
\includegraphics[width=0.23\linewidth,height=0.38\linewidth]{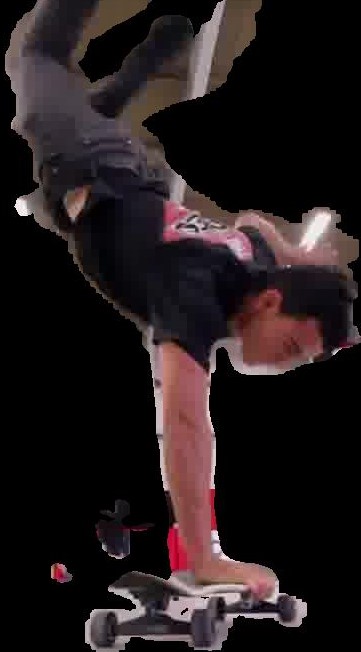}
\includegraphics[width=0.23\linewidth,height=0.38\linewidth]{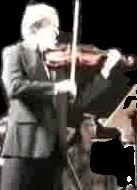}
\includegraphics[width=0.23\linewidth,height=0.38\linewidth]{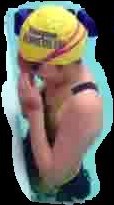}
\includegraphics[width=0.23\linewidth,height=0.38\linewidth]{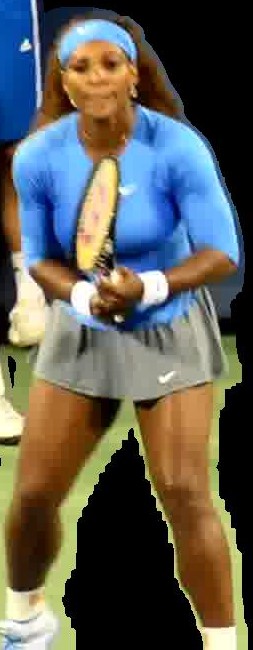}
\includegraphics[width=0.23\linewidth,height=0.38\linewidth]{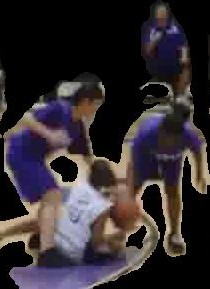}
\includegraphics[width=0.23\linewidth,height=0.38\linewidth]{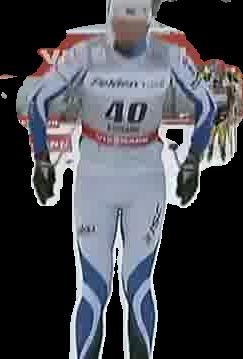}
\caption{Examples of extracted foreground images.}
\label{ss_fig:vis-fg}
\end{figure}

\begin{figure}[!h]
\centering
\begin{minipage}{1.0\linewidth}
    \includegraphics[width=0.24\linewidth,height=0.15\linewidth]{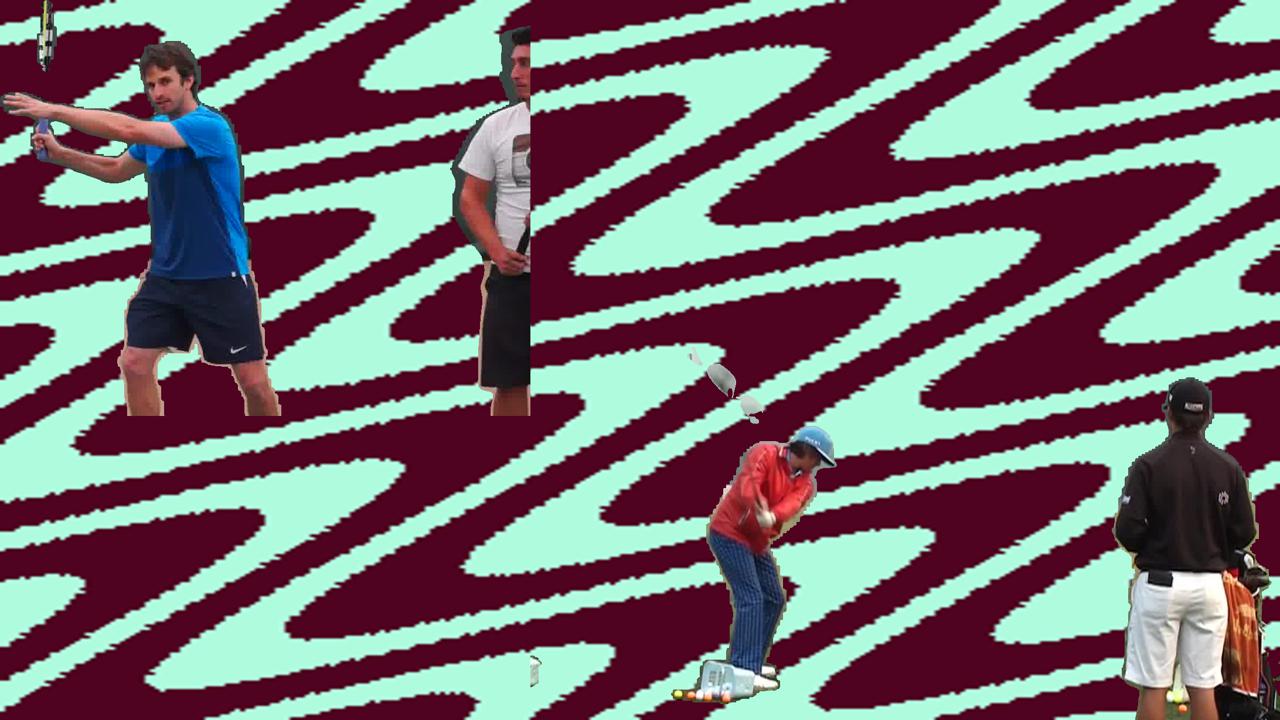}
    \includegraphics[width=0.24\linewidth,height=0.15\linewidth]{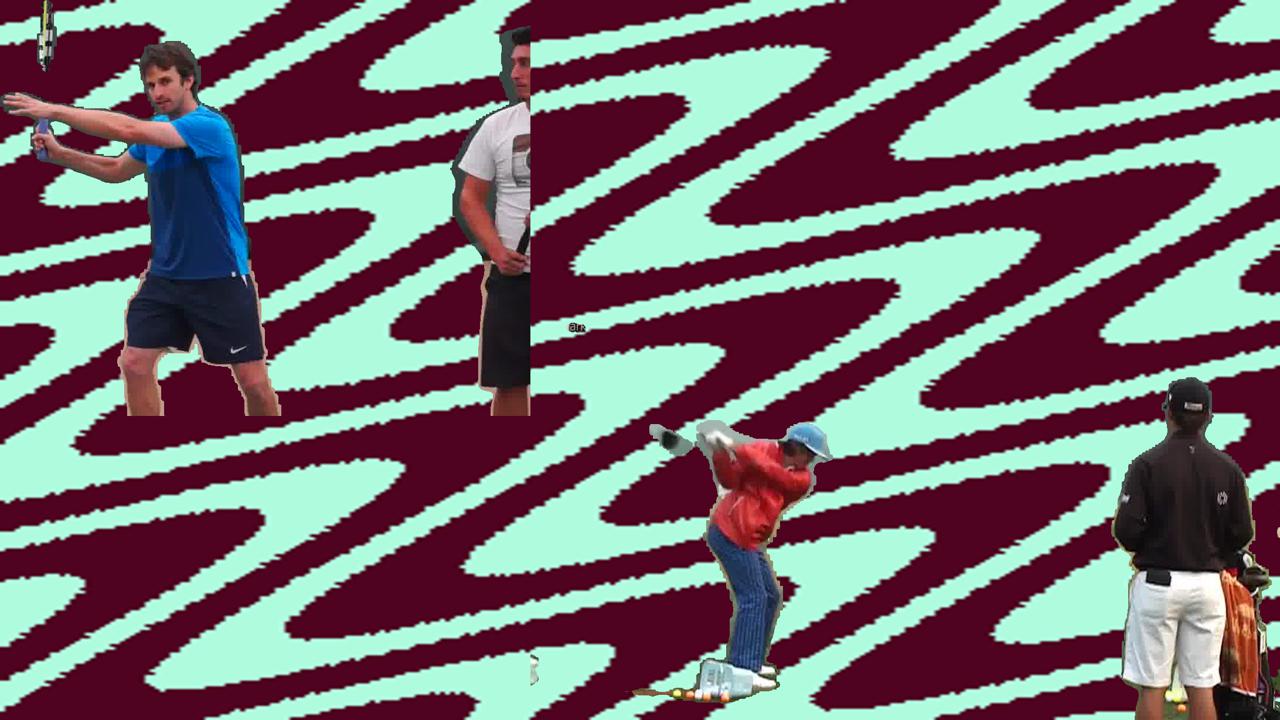}
    \includegraphics[width=0.24\linewidth,height=0.15\linewidth]{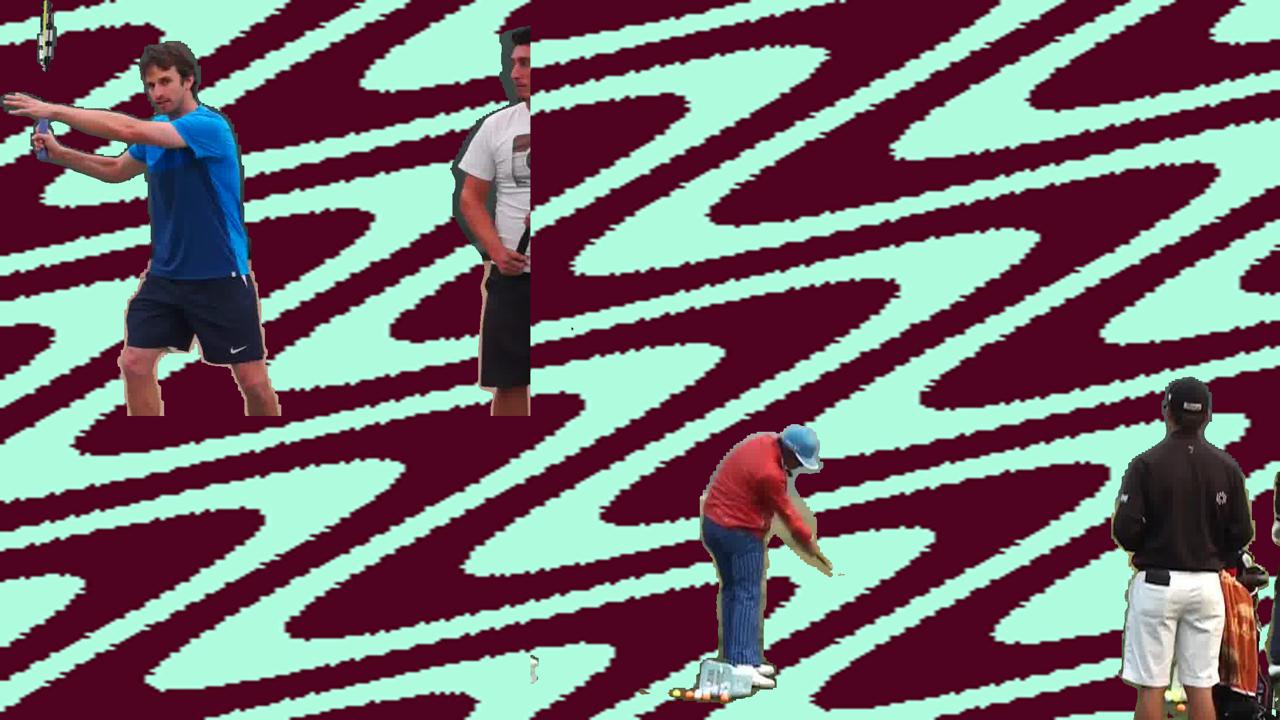}
    \includegraphics[width=0.24\linewidth,height=0.15\linewidth]{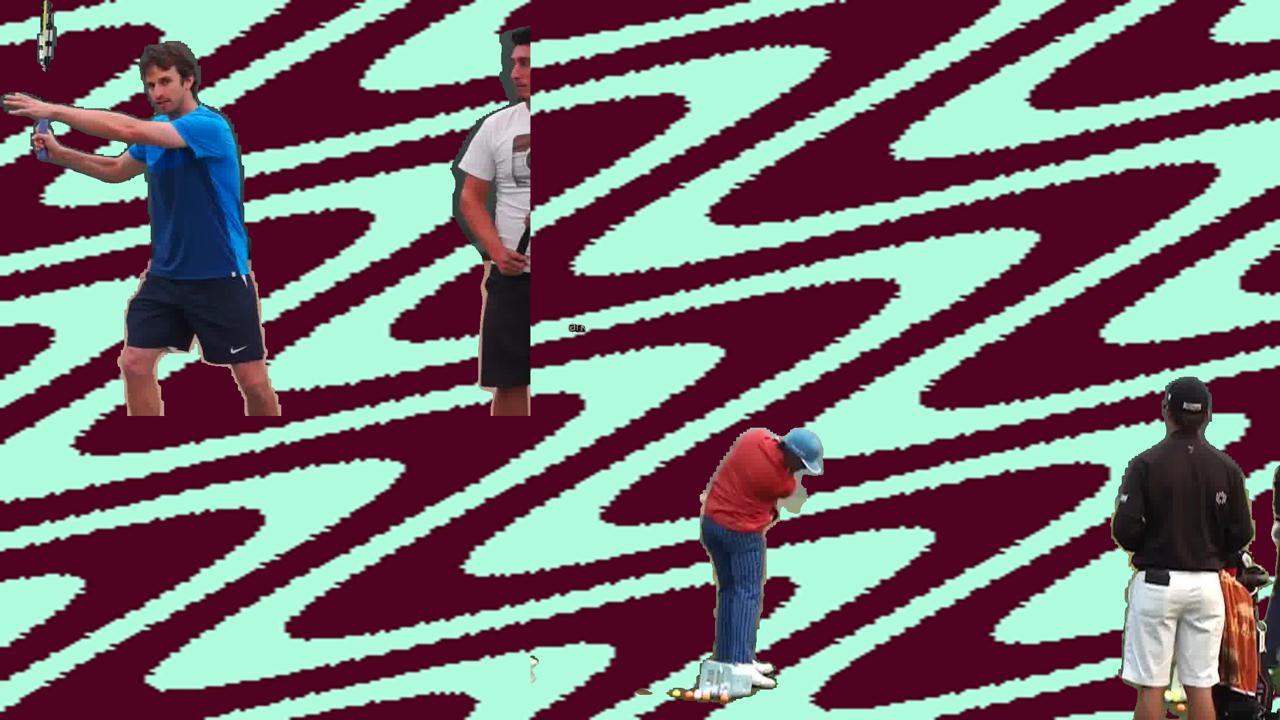}
    \centerline{\textit{Golf driving} with static \textit{playing tennis}}
    \vspace{0.01pt}
\end{minipage}
\begin{minipage}{1.0\linewidth}
    \includegraphics[width=0.24\linewidth,height=0.15\linewidth]{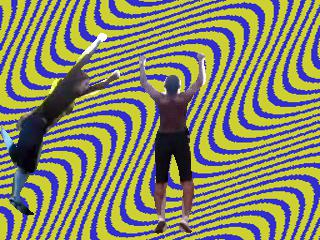}
    \includegraphics[width=0.24\linewidth,height=0.15\linewidth]{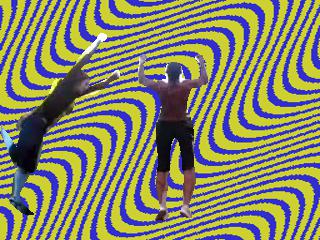}
    \includegraphics[width=0.24\linewidth,height=0.15\linewidth]{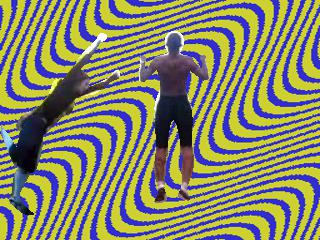}
    \includegraphics[width=0.24\linewidth,height=0.15\linewidth]{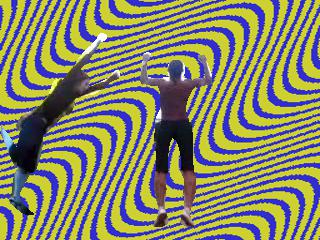}
    \centerline{\textit{Pullup} with static \textit{somersault}}
    \vspace{0.01pt}
\end{minipage}
\begin{minipage}{1.0\linewidth}
    \includegraphics[width=0.24\linewidth,height=0.15\linewidth]{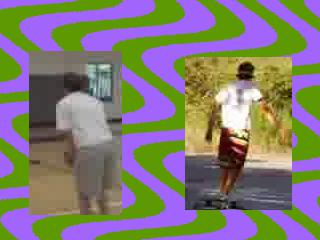}
    \includegraphics[width=0.24\linewidth,height=0.15\linewidth]{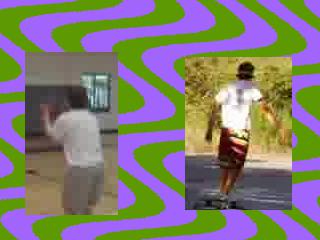}
    \includegraphics[width=0.24\linewidth,height=0.15\linewidth]{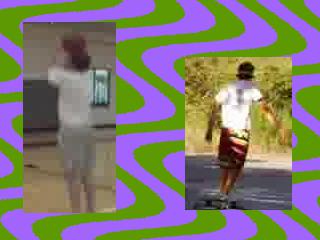}
    \includegraphics[width=0.24\linewidth,height=0.15\linewidth]{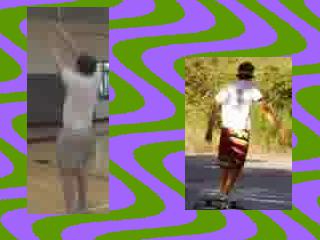}
    \centerline{\textit{Basketball} with static \textit{SkateBoarding}}
\end{minipage}
\caption{Examples of the videos with conflicting foreground cues.}
\label{ss_fig:sample_video_conflicting}
\end{figure}

\section{Experimental Results and Analysis}
\label{ss_sec:exp}

\subsection{Extracted Foreground Images}
In Figure~\ref{ss_fig:vis-fg}, we show several extracted foreground images using the foreground masks in Kinetics-400. From these images, we observe that the foreground motion and the foreground static cues share the same pixels. Therefore, unlike directly separating the pixels of foregrounds and backgrounds in ActorCutMix~\cite{zou2021learning} and FAME~\cite{Ding_2022_CVPR}, it is difficult to separate the pixels of foreground motion and foreground static cues for debiasing foreground static bias. As a result, in this paper, we propose \sysname{} to debias without the need to explicitly extract foreground static cues within a frame. In addition, due to this difficulty, it is hard to create test videos by simply replacing the foreground static cues and preserving the foreground motion. Thus, we alternatively create videos with conflicting foreground cues (Figure 1 of the main paper) and \datasetnameB{} videos (Sec. 4 of the main paper) to evaluate foreground static bias.

\begin{table}[tb]
\begin{center}
\caption{Action recognition accuracy (\%) of different augmentation and debiasing methods on videos with conflicting foreground cues.}
\label{ss_tab:conflicting}
\footnotesize
\setlength\tabcolsep{1pt}
\begin{tabular}{@{}lp{0.01cm}ccp{0.01cm}ccp{0.01cm}cc@{}}
\toprule
\multirow{2}{*}{\begin{tabular}[x]{@{}l@{}} Augmentation\\or Debiasing\end{tabular}} && \multicolumn{2}{c}{Kinetics-400} && \multicolumn{2}{c}{UCF101} && \multicolumn{2}{c}{HMDB51} \\
\cmidrule{3-4}\cmidrule{6-7}\cmidrule{9-10}
~ && Multi-class & Binary && Multi-class & Binary && Multi-class & Binary \\
\midrule
No && 25.25 & 72.83 && 44.65 & 81.94 && 36.58 & 85.03 \\
Mixup && 27.64 & 74.48 && 47.16 & 81.85 && 36.62 & 82.06 \\
VideoMix && 29.37 & 72.50 && 42.59 & 72.21 && 32.68 & 76.07 \\
SDN && 27.14 & 71.14 && 48.47 & 83.83 && 34.87 & 81.88 \\
BE && 26.67 & 72.99 && 46.62 & 81.73 && 35.99 & 85.30 \\
ActorCutMix && 29.02 & 74.02 && 56.88 & 79.60 && 36.97 & 81.07 \\
FAME && 29.50 & 73.83 && 28.21 & 71.70 && 39.61 & 81.56 \\
\sysname{} && \textbf{30.77} & \textbf{85.51} && \textbf{57.30} & \textbf{88.80} && \textbf{47.38} & \textbf{92.46} \\
\bottomrule
\end{tabular}
\end{center}
\end{table}

\subsection{Testing on Videos with Conflicting Foreground Cues}
A video with conflicting foreground cues is synthesized from a \datasetnameA{}-Sinusoid video by the following steps:
\begin{enumerate}
    \setlength{\itemindent}{0em}
    \setlength{\itemsep}{-0.25em}
    \item Randomly sample a video with foreground masks but different action label from the \datasetnameA{}-Sinusoid video.
    \item Randomly sample a frame in the sampled video and use the foreground mask to extract the foreground (mainly containing human actors) as a static foreground.
    \item Randomly select a spatial position in the \datasetnameA{}-Sinusoid video to insert the static foreground such that the inserted static foreground does not overlap with the moving foreground.
    \item Insert the static foreground into all the frames of the \datasetnameA{}-Sinusoid video at the selected spatial position.
    \item Resize the resultant video to the size of the \datasetnameA{}-Sinusoid video.
    \item Keep the label of the resultant video as the same as the \datasetnameA{}-Sinusoid video.
\end{enumerate}

The resultant video contains two action features, one on the static foreground and the other on the moving foreground. We show some example videos in Figure \ref{ss_fig:sample_video_conflicting}. A robust action recognition model should not be affected by the inserted static foregrounds and obtain high accuracy.

We use two metrics to evaluate the performance on the videos with conflicting foreground cues: (1) Multi-class classification accuracy: each video is classified into $N$ action classes, where $N$ is the number of classes defined in the datasets. (2) Binary classification accuracy: each video is classified into two action classes, one indicating the action in the moving foreground and the other indicating the ``action'' in the static foreground.

Table~\ref{ss_tab:conflicting} shows the accuracies of different data augmentation and debiasing methods with Swin-T as the base model. From the results, we observe that \sysname{} obtains the best performance, especially in binary classification (\ie, outperforming other methods by more than 4\%). Although ActorCutMix and FAME outperform \sysname{} on \datasetnameA{} videos (refer to Table 2 of the main paper and Table~\ref{ss_tab:res_aug_ucf101} in the Supplementary Material), they perform worse than \sysname{} on the videos with conflicting foreground cues. The results indicate that FAME and ActorCutMix capture foreground static features as shortcuts instead of learning robust motion features; when the static foregrounds exist, they are interfered to predict the static ``action''. In contrast, \sysname{} shows better robustness to the foreground static features.

\begin{table}[tb]
\begin{center}
\caption{Action recognition accuracy (\%) of different augmentation and debiasing methods on ARAS.}
\label{ss_tab:aras}
\setlength\tabcolsep{6pt}
\footnotesize
\begin{tabular}{@{}lccc@{}}
\toprule
\multirow{2}{*}{\begin{tabular}[x]{@{}l@{}} Augmentation\\or Debiasing\end{tabular}} & \multicolumn{3}{c}{Main Network} \\
\cmidrule{2-4}
~ & TSM & SlowFast & Swin-T \\
\midrule
No & 57.86 & 50.14 & 60.17 \\
Mixup & 58.05 & 50.63 & 59.59 \\
VideoMix & 56.61 & 47.44 & 60.95 \\
SDN & 55.06 & 48.80 & 60.26 \\
BE & 57.47 & 50.92 & 59.79 \\
ActorCutMix & 57.09 & \textbf{51.40} & 61.23 \\
FAME & 57.47 & 48.51 & 60.37 \\
\sysname{} (Ours) & \textbf{59.69} & \textbf{51.40} & \textbf{62.49} \\
\bottomrule
\end{tabular}
\end{center}
\end{table}

\subsection{Testing on ARAS}
To assess the effectiveness of different methods on mitigating scene bias, we conduct tests on a real-world OOD video dataset, ARAS \cite{duan2022mitigating}, which contains actions defined in Kinetics-400 with rare scenes.

After trained on Kinetics-400, the models are directly tested on a balanced test set of ARAS as in \cite{duan2022mitigating}. Table~\ref{ss_tab:aras} shows the accuracies of different data augmentation and debiasing methods. From the results, we observe that \sysname{} obtains the best performance, illustrating its effectiveness on mitigating scene bias in real-world videos.


\begin{figure}[t]
    \centering
    \includegraphics[width=\linewidth]{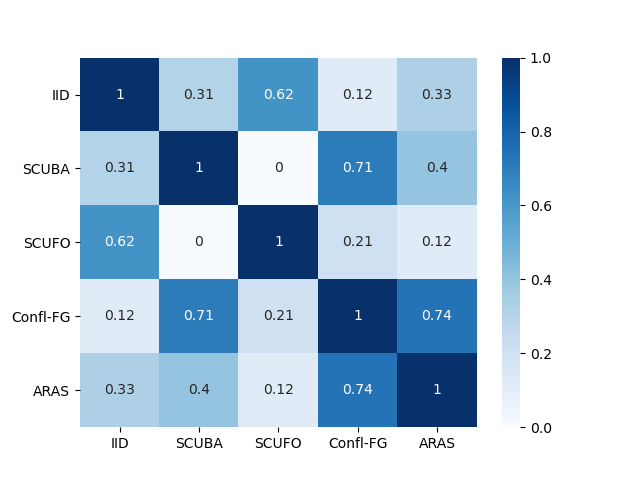}
    \caption{Correlations between different evaluations are not strong. Confl-FG denotes videos with conflicting foreground cues.}
    \label{ss_fig:rank_corr}
\end{figure}

\subsection{Correlations between Different Evaluations}
To determine the level of similarity between different evaluations, we calculate the Spearman's rank correlation coefficients between the performances of different methods on any pair of evaluation data. We use the performance of Swin-T on Kinetics-400 for calculation. The results are shown in Figure \ref{ss_fig:rank_corr}, from which we observe that the correlation coefficients are lower than 0.75, indicating that the correlations between any pair of the evaluation data are not strong. Therefore, these evaluation data may assess the action representations from different perspectives and could not directly replace each other.

\begin{table}[t]
\begin{center}
\caption{IID and OOD test accuracy (\%) of \sysname{} based on UniformerV2.}
\label{ss_tab:res_uniformerv2}
\footnotesize
\setlength\tabcolsep{0.5pt}
\begin{tabular}{@{}llp{0.01cm}cccccc@{}}
\toprule
\multirow{3}{*}{Dataset} & \multirow{3}{*}{Debiasing} && \multirow{3}{*}{IID} & \multicolumn{5}{c}{OOD} \\
\cmidrule{5-9}
~ & ~ && ~ & \multirow{2}{*}{\begin{tabular}[x]{@{}c@{}} Avg\\\datasetnameA{}\end{tabular}$\big\uparrow$} & \multirow{2}{*}{\begin{tabular}[x]{@{}c@{}} Avg\\\datasetnameB{}\end{tabular}$\big\downarrow$} & \multirow{2}{*}{\begin{tabular}[x]{@{}c@{}} Contra.\\Acc.\end{tabular}$\big\uparrow$} & \multirow{2}{*}{\begin{tabular}[x]{@{}c@{}} Confl-\\FG\end{tabular}$\big\uparrow$} & \multirow{2}{*}{ARAS$\big\uparrow$} \\ \\
\midrule
\multirow{2}{*}{Kinetics-400} & No && 87.84 & 68.45 & 43.98 & 26.97 & 42.09 & 81.97 \\
~ & \sysname{} && \textbf{88.28} & \textbf{70.23} & \textbf{43.24} & \textbf{29.28} & \textbf{42.53} & \textbf{82.35} \\
\midrule
\multirow{2}{*}{HMDB51} & No && \textbf{82.94} & 61.73 & 50.75 & 15.43 & 34.10 & -- \\
~ & \sysname{} && 82.35 & \textbf{62.21} & \textbf{49.75} & \textbf{16.59} & \textbf{34.49} & -- \\
\midrule
\multirow{2}{*}{UCF101} & No && 98.18 & 63.65 & 38.73 & 26.45 & 47.58 & -- \\
~ & \sysname{} && \textbf{98.36} & \textbf{64.56} & \textbf{38.61} & \textbf{27.53} & \textbf{48.07} & -- \\
\bottomrule
\end{tabular}
\end{center}
\end{table}

\subsection{Performance of UniformerV2}
With \sysname{}, we finetune UniformerV2~\cite{li2022uniformerv2}, the SOTA opensource action recognition model. We finetune UniFormerV2-L/14 from CLIP and K710 pretrained weights, using 32 frames as inputs. The results are shown in Table~\ref{ss_tab:res_uniformerv2}. As UniformerV2 adopts pretrained weights from CLIP, an image-only network, it has strong static bias and performs poorly on \datasetnameB{} and Confl-FG. For example, Swin-T achieves 36.58\% on HMDB51-Confl-FG (Table \ref{ss_tab:conflicting}) but UniformerV2, having gone through much more pretraining, achieves only 34.10\%. Hence, finetuning with \sysname{} on small datasets like HMDB51 and UCF101 could not substantially correct this bias, but \sysname{} still shows improvements over the original model.

\begin{table*}[tb]
\begin{center}
\caption{Action recognition accuracy (\%) of transferring the learned representations across datasets.}
\label{ss_tab:transfer}
\footnotesize
\setlength\tabcolsep{4pt}
\vspace{-0.5em}
\begin{tabular}{@{}cp{0.01cm}lp{0.01cm}cccc@{}}
\toprule
\multirow{2}{*}{Network} && \multirow{2}{*}{\begin{tabular}[x]{@{}l@{}} Augmentation\\or Debiasing\end{tabular}} && \multicolumn{4}{c}{Source$\rightarrow$Target} \\
\cmidrule{5-8}
~ && ~ && Kinetics400$\rightarrow$UCF101 & Kinetics400$\rightarrow$HMDB51 & HMDB51$\rightarrow$UCF101 & UCF101$\rightarrow$HMDB51 \\
\midrule
\multirow{8}{*}{TSM} && No && 92.52 & 66.67 & 61.64 & 44.95 \\
~ && Mixup && 93.07 & 68.69 & 63.58 & 46.60 \\
~ && VideoMix && 93.55 & 69.22 & 61.49 & 40.33 \\
~ && SDN && 92.81 & 63.79 & 61.12 & 41.90 \\
~ && BE && 93.10 & 67.45 & 62.71 & 45.88 \\
~ && ActorCutMix && 92.73 & 67.39 & 61.67 & 42.92 \\
~ && FAME && 93.87 & 67.84 & 58.87 & 44.99 \\
~ && \sysname{} && \textbf{93.89} & \textbf{70.07} & \textbf{65.69} & \textbf{47.99} \\
\midrule
\multirow{8}{*}{SlowFast} && No && 91.86 & 67.32 & 42.38 & 40.59 \\
~ && Mixup && 90.14 & 65.49 & 42.63 & 43.86 \\
~ && VideoMix && 89.80 & 64.25 & 42.90 & 39.30 \\
~ && SDN && 89.29 & 61.30 & 43.46 & 38.91 \\
~ && BE && 91.91 & 67.12 & 40.89 & 41.29 \\
~ && ActorCutMix && 91.73 & 67.19 & 43.66 & 39.17 \\
~ && FAME && 91.01 & 65.10 & 39.49 & 39.37 \\
~ && \sysname{} && \textbf{92.49} & \textbf{67.84} & \textbf{46.23} & \textbf{44.77} \\
\midrule
\multirow{8}{*}{Swin-T} && No && 95.74 & 72.03 & 75.67 & 52.83 \\
~ && Mixup && 95.40 & 72.42 & 76.61 & 51.59 \\
~ && VideoMix && 95.32 & 71.24 & 74.95 & 50.59 \\
~ && SDN && 94.90 & 70.13 & 74.42 & 49.87 \\
~ && BE && 95.43 & 71.63 & 76.78 & 53.31 \\
~ && ActorCutMix && 95.72 & 72.55 & 75.57 & 52.81 \\
~ && FAME && 95.40 & 70.78 & 75.67 & 50.55 \\
~ && \sysname{} && \textbf{95.77} & \textbf{72.75} & \textbf{78.50} & \textbf{53.71} \\
\bottomrule
\end{tabular}
\end{center}
\vspace{-2em}
\end{table*}

\begin{table}[!ht]
\begin{center}
\caption{Action recognition accuracy (\%) of \sysname{} with different frame sampling strategies.}
\label{ss_tab:sampling}
\setlength\tabcolsep{4pt}
\footnotesize
\vspace{-0.5em}
\begin{tabular}{@{}cccp{0.01cm}cc@{}}
\toprule
\multirow{3}{*}{Network} & \multirow{3}{*}{Pretrain} & \multirow{3}{*}{Sampling strategy} && \multicolumn{2}{c}{HMDB51} \\
\cmidrule{5-6}
~ & ~ & ~ && \multirow{2}{*}{IID} & \multirow{2}{*}{\begin{tabular}[x]{@{}c@{}} Contra.\\Acc.\end{tabular}} \\ \\
\midrule
\multirow{6}{*}{TSM} & \multirow{3}{*}{ImageNet} & No RefNet && 54.07 & 31.21 \\
~ & ~ & RefNet && 54.66 & \textbf{33.14} \\
~ & ~ & RefNet Inversed && 54.79 & 29.17 \\
\cmidrule{2-6}
~ & \multirow{3}{*}{Kinetics400} & No RefNet && 71.87 & 41.86 \\
~ & ~ & RefNet && 71.52 & \textbf{42.05} \\
~ & ~ & RefNet Inversed && 72.27 & 38.98 \\
\midrule
\multirow{6}{*}{SlowFast} & \multirow{3}{*}{ImageNet} & No RefNet && 51.79 & 20.94 \\
~ & ~ & RefNet && 51.53 & \textbf{21.21} \\
~ & ~ & RefNet Inversed && 50.94 & 18.61 \\
\cmidrule{2-6}
~ & \multirow{3}{*}{Kinetics400} & No RefNet && 76.27 & 34.34 \\
~ & ~ & RefNet && 76.52 & \textbf{35.20} \\
~ & ~ & RefNet Inversed && 76.12 & 33.83 \\
\midrule
\multirow{6}{*}{Swin-T} & \multirow{3}{*}{ImageNet} & No RefNet && 55.62 & 18.89 \\
~ & ~ & RefNet && 55.36 & \textbf{21.40} \\
~ & ~ & RefNet Inversed && 56.34 & 18.18 \\
\cmidrule{2-6}
~ & \multirow{3}{*}{Kinetics400} & No RefNet && 75.16 & 39.66 \\
~ & ~ & RefNet && 74.82 & \textbf{40.28} \\
~ & ~ & RefNet Inversed && 75.62 & 37.82 \\
\bottomrule
\end{tabular}
\end{center}
\vspace{3em}
\end{table}

\begin{table}[!ht]
\begin{center}
\caption{Action recognition accuracy (\%) of \sysname{} with label mixing. $\lambda^{\prime}=1$ means \sysname{} without label mixing (the default setting).}
\label{ss_tab:mix_label}
\setlength\tabcolsep{5pt}
\footnotesize
\vspace{-0.5em}
\begin{tabular}{@{}ccp{0.01cm}ccp{0.01cm}cc@{}}
\toprule
\multirow{3}{*}{Pretrain} & \multirow{3}{*}{$\lambda^{\prime}$} && \multicolumn{2}{c}{UCF101} && \multicolumn{2}{c}{HMDB51} \\
\cmidrule{4-5}\cmidrule{7-8}
~ & ~ && \multirow{2}{*}{IID} & \multirow{2}{*}{\begin{tabular}[x]{@{}c@{}} Contra.\\Acc.\end{tabular}} && \multirow{2}{*}{IID} & \multirow{2}{*}{\begin{tabular}[x]{@{}c@{}} Contra.\\Acc.\end{tabular}} \\ \\
\midrule
\multirow{3}{*}{ImageNet} & 1 && 87.29 & \textbf{24.60} && 54.66 & \textbf{33.14} \\
~ & 0.8 && 86.40 & 14.56 && 54.44 & 22.69 \\
~ & $\lambda$ && 84.35 & 10.70 && 49.76 & 5.06 \\
\midrule
\multirow{3}{*}{Kinetics400} & 1 && 94.30 & \textbf{36.47} && 71.52 & \textbf{42.05} \\
~ & 0.8 && 94.70 & 31.54 && 72.07 & 35.19 \\
~ & $\lambda$ && 93.85 & 18.82 && 70.92 & 19.44 \\
\bottomrule
\end{tabular}
\end{center}
\vspace{-0.5em}
\end{table}

\begin{table}[!ht]
\begin{center}
\caption{Action recognition accuracy (\%) of \sysname{} with different Beta distributions parameters. IN and K400 denote ImageNet and Kinetics-400 respectively.}
\label{ss_tab:beta}
\setlength\tabcolsep{2.5pt}
\scriptsize
\vspace{-0.5em}
\begin{tabular}{@{}ccllp{0.01cm}ccp{0.01cm}cc@{}}
\toprule
\multirow{3}{*}{Pretrain} & \multirow{3}{*}{$Beta(\alpha, \beta)$} & \multirow{3}{*}{\begin{tabular}[x]{@{}l@{}}Mean\\of $\lambda$\end{tabular}} & \multirow{3}{*}{\begin{tabular}[x]{@{}l@{}}Variance\\of $\lambda$\end{tabular}} && \multicolumn{2}{c}{UCF101} && \multicolumn{2}{c}{HMDB51} \\
\cmidrule{6-7}\cmidrule{9-10}
~ & ~ & ~ & ~ && \multirow{2}{*}{IID} & \multirow{2}{*}{\begin{tabular}[x]{@{}c@{}} Contra.\\Acc.\end{tabular}} && \multirow{2}{*}{IID} & \multirow{2}{*}{\begin{tabular}[x]{@{}c@{}} Contra.\\Acc.\end{tabular}} \\ \\
\midrule
\multirow{5}{*}{IN} & $(300, 100)$ & $0.75$ & $0.00047$ && 85.98 & 12.43 && 50.70 & 7.95 \\
~ & $(100, 300)$ & $0.25$ & $0.00047$ && 87.24 & 22.05 && 54.92 & \textbf{38.06} \\
~ & $(200, 200)$ & $0.5$ & $0.00062$ && 87.29 & \textbf{24.60} && 54.66 & 33.14 \\
~ & $(100, 100)$ & $0.5$ & $0.0012$ && 87.41 & 22.20 && 55.08 & 32.18 \\
~ & $(20, 20)$ & $0.5$ & $0.0060$ && 87.42 & 23.01 && 54.95 & 33.06 \\
\midrule
\multirow{5}{*}{K400} & $(300, 100)$ & $0.75$ & $0.00047$ && 94.49 & 27.58 && 71.65 & 26.22 \\
~ & $(100, 300)$ & $0.25$ & $0.00047$ && 94.36 & \textbf{37.38} && 72.29 & \textbf{49.41} \\
~ & $(200, 200)$ & $0.5$ & $0.00062$ && 94.30 & 36.47 && 71.52 & 42.05 \\
~ & $(100, 100)$ & $0.5$ & $0.0012$ && 94.41 & 36.64 && 72.18 & 41.94 \\
~ & $(20, 20)$ & $0.5$ & $0.0060$ && 94.60 & 36.12 && 71.85 & 42.69 \\
\bottomrule
\end{tabular}
\end{center}
\vspace{-2.5em}
\end{table}

\begin{table}[!ht]
\begin{center}
\caption{Action recognition accuracy (\%) of \sysname{} with different $P_{aug}$.}
\label{ss_tab:Paug}
\setlength\tabcolsep{2pt}
\footnotesize
\vspace{-0.5em}
\begin{tabular}{@{}ccp{0.01cm}ccp{0.01cm}ccp{0.01cm}cc@{}}
\toprule
\multirow{3}{*}{Pretrain} & \multirow{3}{*}{$P_{aug}$} && \multicolumn{2}{c}{UCF101} && \multicolumn{2}{c}{HMDB51} && \multicolumn{2}{c}{Kinetics400} \\
\cmidrule{4-5}\cmidrule{7-8}\cmidrule{10-11}
~ & ~ && \multirow{2}{*}{IID} & \multirow{2}{*}{\begin{tabular}[x]{@{}c@{}} Contra.\\Acc.\end{tabular}} && \multirow{2}{*}{IID} & \multirow{2}{*}{\begin{tabular}[x]{@{}c@{}} Contra.\\Acc.\end{tabular}} && \multirow{2}{*}{IID} & \multirow{2}{*}{\begin{tabular}[x]{@{}c@{}} Contra.\\Acc.\end{tabular}} \\ \\
\midrule
\multirow{3}{*}{ImageNet} & $0.25$ && 87.29 & \textbf{24.60} && 53.92 & 26.20 && 71.28 & 36.07 \\
~ & $0.5$ && 87.19 & 22.78 && 54.66 & 33.14 && 70.71 & \textbf{39.00} \\
~ & $0.75$ && 86.90 & 20.81 && 54.34 & \textbf{34.56} && 70.15 & 38.97 \\
\midrule
\multirow{3}{*}{Kinetics400} & $0.25$ && 94.30 & \textbf{36.47} && 71.52 & 42.05 && - & - \\
~ & $0.5$ && 94.31 & 35.10 && 72.40 & 45.88 && - & - \\
~ & $0.75$ && 94.28 & 34.19 && 71.83 & \textbf{47.83} && - & - \\
\bottomrule
\end{tabular}
\end{center}
\vspace{-1em}
\end{table}

\begin{table*}[tb]
\begin{center}
\caption{Action recognition accuracy (\%) of different augmentation and debiasing methods on the Kinetics-400, Kinetics400-\datasetnameA{} and Kinetics400-\datasetnameB{} datasets. $\dagger$ indicates adapting from self-supervised debiasing methods.}
\label{ss_tab:res_aug_k400}
\footnotesize
\setlength\tabcolsep{3pt}
\begin{tabular}{@{}cclp{0.01cm}cp{0.01cm}cccp{0.01cm}cccp{0.01cm}c@{}}
\toprule
\multirow{2}{*}{Model} & \multirow{2}{*}{Pretrain} & \multirow{2}{*}{\begin{tabular}[x]{@{}l@{}} Augmentation\\or Debiasing\end{tabular}} && \multirow{2}{*}{Kinetics-400} && \multicolumn{3}{c}{Kinetics400-\datasetnameA{} ($\uparrow$)} && \multicolumn{3}{c}{Kinetics400-\datasetnameB{} ($\downarrow$)} && \multirow{2}{*}{\begin{tabular}[x]{@{}c@{}}Contra.\\Acc.\end{tabular}($\big\uparrow$)} \\
\cmidrule{7-9} \cmidrule{11-13}
~ & ~ & ~ && ~ && Place365 & VQGAN-CLIP & Sinusoid && Place365 & VQGAN-CLIP & Sinusoid \\
\midrule\midrule
\multirow{8}{*}{TSM} & \multirow{8}{*}{ImageNet} & No && 71.13 && 40.44 & 37.39 & 34.34 && 18.55 & 16.61 & 16.49 && 22.80 \\
~ & ~ & Mixup && 71.33 && 42.82 & 40.65 & 38.95 && 18.66 & 17.20 & 16.72 && 25.98 \\
~ & ~ & VideoMix && \textbf{71.35} && 40.93 & 38.96 & 36.71 && 17.89 & 17.13 & 16.73 && 24.57 \\
~ & ~ & SDN && 69.99 && 38.42 & 35.71 & 36.72 && 16.98 & 15.31 & 17.35 && 22.38 \\
~ & ~ & BE$\dagger$ && 71.30 && 41.07 & 38.19 & 34.42 && 17.05 & 16.14 & 15.04 && 24.35 \\
~ & ~ & ActorCutMix$\dagger$ && 71.07 && 42.89 & 40.89 & 37.47 && 17.60 & 15.63 & 15.65 && 26.52 \\
~ & ~ & FAME$\dagger$ && 71.13 && 43.30 & 40.41 & \textbf{39.01} && 18.96 & 17.73 & 18.32 && 25.63 \\
~ & ~ & \cellcolor{lightgray}\sysname{} (Ours) &\cellcolor{lightgray}& \cellcolor{lightgray}71.28 &\cellcolor{lightgray}& \cellcolor{lightgray}\textbf{43.31} & \cellcolor{lightgray}\textbf{40.97} & \cellcolor{lightgray}37.15 &\cellcolor{lightgray}& \cellcolor{lightgray}\textbf{6.27} & \cellcolor{lightgray}\textbf{5.17} & \cellcolor{lightgray}\textbf{4.24} &\cellcolor{lightgray}& \cellcolor{lightgray}\textbf{36.07} \\
\midrule
\multirow{8}{*}{SlowFast} & \multirow{8}{*}{ImageNet} & No && 65.63 && 36.96 & 35.54 & 34.88 && 21.74 & 20.19 & 20.51 && 18.98 \\
~ & ~ & Mixup && 65.16 && 37.65 & 35.63 & 34.98 && 20.85 & 18.81 & 18.81 && 20.17 \\
~ & ~ & VideoMix && 64.26 && 35.23 & 33.81 & 34.29 && 19.84 & 18.05 & 18.57 && 19.41 \\
~ & ~ & SDN && 63.49 && 33.71 & 31.60 & 31.86 && 19.81 & 18.21 & 19.69 && 17.13 \\
~ & ~ & BE$\dagger$ && 65.65 && 36.15 & 34.66 & 33.64 && 19.13 & 17.25 & 18.30 && 20.15 \\
~ & ~ & ActorCutMix$\dagger$ && \textbf{65.79} && 39.61 & \textbf{38.28} & 36.05 && 20.08 & 18.58 & 18.95 && 22.01 \\
~ & ~ & FAME$\dagger$ && 65.13 && \textbf{40.11} & 37.48 & \textbf{38.63} && 20.89 & 19.21 & 20.79 && 22.07 \\
~ & ~ & \cellcolor{lightgray}\sysname{} (Ours) &\cellcolor{lightgray}& \cellcolor{lightgray}65.65 &\cellcolor{lightgray}& \cellcolor{lightgray}37.63 & \cellcolor{lightgray}35.71 & \cellcolor{lightgray}35.46 &\cellcolor{lightgray}& \cellcolor{lightgray}\textbf{14.78} & \cellcolor{lightgray}\textbf{13.35} & \cellcolor{lightgray}\textbf{12.70} &\cellcolor{lightgray}& \cellcolor{lightgray}\textbf{25.01} \\
\midrule
\multirow{8}{*}{Swin-T} & \multirow{8}{*}{ImageNet} & No && 73.95 && 40.81 & 39.67 & 44.75 && 17.89 & 15.89 & 20.73 && 25.93 \\
~ & ~ & Mixup && 73.91 && 42.59 & 42.05 & 47.22 && 17.33 & 15.74 & 20.68 && 28.24 \\
~ & ~ & VideoMix && 73.80 && 41.89 & 41.00 & 46.63 && 18.72 & 16.36 & 22.71 && 26.40 \\
~ & ~ & SDN && 72.23 && 39.59 & 39.90 & 47.52 && 20.04 & 19.21 & 25.13 && 24.46 \\
~ & ~ & BE$\dagger$ && 73.93 && 41.88 & 41.86 & 46.47 && 18.73 & 17.12 & 22.84 && 26.28 \\
~ & ~ & ActorCutMix$\dagger$ && \textbf{73.97} && 44.16 & 45.06 & 47.87 && 19.58 & 17.06 & 21.54 && 28.64 \\
~ & ~ & FAME$\dagger$ && 73.81 && \textbf{49.01} & \textbf{47.71} & \textbf{49.66} && 21.38 & 19.10 & 23.33 && 30.03 \\
~ & ~ & \cellcolor{lightgray}\sysname{} (Ours) &\cellcolor{lightgray}& \cellcolor{lightgray}73.86 &\cellcolor{lightgray}& \cellcolor{lightgray}43.44 & \cellcolor{lightgray}42.81 & \cellcolor{lightgray}46.05 &\cellcolor{lightgray}& \cellcolor{lightgray}\textbf{4.76} & \cellcolor{lightgray}\textbf{4.37} & \cellcolor{lightgray}\textbf{7.41} &\cellcolor{lightgray}& \cellcolor{lightgray}\textbf{39.41} \\
\bottomrule
\end{tabular}
\end{center}
\end{table*}

\begin{table*}[tb]
\begin{center}
\caption{Action recognition accuracy (\%) of different augmentation and debiasing methods on the HMDB51, HMDB51-\datasetnameA{} and HMDB51-\datasetnameB{} datasets. K400 denotes Kinetics-400. $\dagger$ indicates adapting from self-supervised debiasing methods.}
\label{ss_tab:res_aug_hmdb51}
\footnotesize
\setlength\tabcolsep{3pt}
\begin{tabular}{@{}cclp{0.01cm}cp{0.01cm}cccp{0.01cm}cccp{0.01cm}c@{}}
\toprule
\multirow{2}{*}{Model} & \multirow{2}{*}{Pretrain} & \multirow{2}{*}{\begin{tabular}[x]{@{}l@{}} Augmentation\\or Debiasing\end{tabular}} && \multirow{2}{*}{HMDB51} && \multicolumn{3}{c}{HMDB51-\datasetnameA{} ($\uparrow$)} && \multicolumn{3}{c}{HMDB51-\datasetnameB{} ($\downarrow$)} && \multirow{2}{*}{\begin{tabular}[x]{@{}c@{}}Contra.\\Acc.\end{tabular}($\big\uparrow$)} \\
\cmidrule{7-9} \cmidrule{11-13}
~ & ~ & ~ && ~ && Place365 & VQGAN-CLIP & Sinusoid && Place365 & VQGAN-CLIP & Sinusoid \\
\midrule\midrule
\multirow{8}{*}{TSM} & \multirow{8}{*}{K400} & No && 70.39 && 45.09 & 42.16 & 26.84 && 23.26 & 20.03 & 14.40 && 22.02 \\
~ & ~ & Mixup && \textbf{72.00} && 46.25 & 44.07 & 28.96 && 22.60 & 19.92 & 14.71 && 23.76 \\
~ & ~ & VideoMix && 70.72 && 42.68 & 41.46 & 23.00 && 20.98 & 18.99 & 12.46 && 21.03 \\
~ & ~ & SDN && 69.51 && 40.79 & 38.92 & 31.44 && 19.18 & 14.91 & 18.70 && 23.74 \\
~ & ~ & BE$\dagger$ && 71.22 && 45.39 & 42.81 & 27.25 && 23.42 & 20.52 & 14.40 && 22.39 \\
~ & ~ & ActorCutMix$\dagger$ && 70.52 && 45.81 & 42.32 & 27.08 && 23.29 & 20.43 & 15.12 && 21.94 \\
~ & ~ & FAME$\dagger$ && 70.39 && 52.03 & \textbf{53.21} & 36.33 && 26.04 & 23.34 & 17.60 && 28.21 \\
~ & ~ & \cellcolor{lightgray}\sysname{} (Ours) &\cellcolor{lightgray}& \cellcolor{lightgray}71.52 &\cellcolor{lightgray}& \cellcolor{lightgray}\textbf{53.91} & \cellcolor{lightgray}52.66 & \cellcolor{lightgray}\textbf{38.13} &\cellcolor{lightgray}& \cellcolor{lightgray}\textbf{11.63} & \cellcolor{lightgray}\textbf{8.29} & \cellcolor{lightgray}\textbf{5.38} &\cellcolor{lightgray}& \cellcolor{lightgray}\textbf{42.05} \\
\midrule
\multirow{8}{*}{SlowFast} & \multirow{8}{*}{K400} & No && 76.25 && 47.36 & 48.23 & 35.00 && 23.26 & 23.14 & 18.78 && 28.37 \\
~ & ~ & Mixup && 75.69 && 48.72 & 50.20 & 35.38 && 24.42 & 23.99 & 19.60 && 28.22 \\
~ & ~ & VideoMix && 75.62 && 48.19 & 48.87 & 34.81 && 23.98 & 23.80 & 19.17 && 27.25 \\
~ & ~ & SDN && 76.17 && 34.27 & 37.80 & 30.41 && \textbf{9.39} & \textbf{12.04} & \textbf{12.30} && 24.99 \\
~ & ~ & BE$\dagger$ && 75.82 && 46.46 & 47.29 & 34.51 && 23.50 & 23.46 & 19.20 && 27.74 \\
~ & ~ & ActorCutMix$\dagger$ && 75.49 && 53.28 & 52.94 & 38.96 && 26.57 & 26.42 & 21.28 && 28.96 \\
~ & ~ & FAME$\dagger$ && 74.66 && \textbf{57.61} & \textbf{58.27} & \textbf{47.08} && 25.68 & 24.02 & 21.52 && 34.32 \\
~ & ~ & \cellcolor{lightgray}\sysname{} (Ours) &\cellcolor{lightgray}& \cellcolor{lightgray}\textbf{76.52} &\cellcolor{lightgray}& \cellcolor{lightgray}48.31 & \cellcolor{lightgray}47.75 & \cellcolor{lightgray}40.31 &\cellcolor{lightgray}& \cellcolor{lightgray}17.76 & \cellcolor{lightgray}17.51 & \cellcolor{lightgray}15.65 &\cellcolor{lightgray}& \cellcolor{lightgray}\textbf{35.20} \\
\midrule
\multirow{8}{*}{Swin-T} & \multirow{8}{*}{K400} & No && 73.92 && 47.61 & 42.77 & 41.41 && 20.68 & 17.90 & 22.80 && 27.84 \\
~ & ~ & Mixup && 74.58 && 46.70 & 42.49 & 40.12 && 21.25 & 18.47 & 23.78 && 26.09 \\
~ & ~ & VideoMix && 73.31 && 41.33 & 38.18 & 38.67 && 19.64 & 18.82 & 22.85 && 23.13 \\
~ & ~ & SDN && 74.66 && 41.96 & 40.82 & 37.29 && 19.99 & 19.62 & 21.06 && 22.88 \\
~ & ~ & BE$\dagger$ && 74.31 && 47.36 & 42.94 & 40.39 && 20.91 & 17.55 & 21.41 && 27.84 \\
~ & ~ & ActorCutMix$\dagger$ && 74.05 && 50.13 & 46.51 & 43.73 && 22.16 & 20.26 & 23.80 && 28.12 \\
~ & ~ & FAME$\dagger$ && 73.79 && \textbf{54.71} & \textbf{53.67} & 45.81 && 27.10 & 27.26 & 26.40 && 29.66 \\
~ & ~ & \cellcolor{lightgray}\sysname{} (Ours) &\cellcolor{lightgray}& \cellcolor{lightgray}\textbf{74.82} &\cellcolor{lightgray}& \cellcolor{lightgray}53.27 & \cellcolor{lightgray}52.43 & \cellcolor{lightgray}\textbf{49.73} &\cellcolor{lightgray}& \cellcolor{lightgray}\textbf{13.39} & \cellcolor{lightgray}\textbf{12.66} & \cellcolor{lightgray}\textbf{14.13} &\cellcolor{lightgray}& \cellcolor{lightgray}\textbf{40.28} \\
\midrule\midrule
\multirow{8}{*}{TSM} & \multirow{8}{*}{ImageNet} & No && 47.56 && 19.39 & 16.99 & 8.49 && 11.78 & 11.12 & 6.41 && 6.50 \\
~ & ~ & Mixup && 51.68 && 23.96 & 18.76 & 14.06 && 18.49 & 15.04 & 12.10 && 5.66 \\
~ & ~ & VideoMix && 48.32 && 21.41 & 18.44 & 10.27 && 13.23 & 12.42 & 7.64 && 7.17 \\
~ & ~ & SDN && 45.40 && 22.57 & 16.63 & 13.11 && 11.37 & 8.08 & 8.15 && 10.72 \\
~ & ~ & BE$\dagger$ && 48.87 && 22.13 & 16.86 & 13.11 && 17.39 & 14.33 & 11.39 && 4.66 \\
~ & ~ & ActorCutMix$\dagger$ && 48.39 && 25.52 & 21.38 & 11.57 && 15.27 & 13.16 & 8.39 && 9.03 \\
~ & ~ & FAME$\dagger$ && 45.73 && 26.37 & 24.34 & 15.46 && 14.69 & 15.69 & 10.44 && 10.71 \\
~ & ~ & \cellcolor{lightgray}\sysname{} (Ours) &\cellcolor{lightgray}& \cellcolor{lightgray}\textbf{54.66} &\cellcolor{lightgray}& \cellcolor{lightgray}\textbf{39.52} & \cellcolor{lightgray}\textbf{38.41} & \cellcolor{lightgray}\textbf{33.00} &\cellcolor{lightgray}& \cellcolor{lightgray}\textbf{6.98} & \cellcolor{lightgray}\textbf{6.09} & \cellcolor{lightgray}\textbf{3.58} &\cellcolor{lightgray}& \cellcolor{lightgray}\textbf{33.14} \\
\midrule
\multirow{8}{*}{SlowFast} & \multirow{8}{*}{ImageNet} & No && 47.65 && 21.24 & 16.85 & 10.43 && 17.84 & 15.63 & 11.76 && 5.62 \\
~ & ~ & Mixup && 48.67 && 23.22 & 18.75 & 12.60 && 19.57 & 16.19 & 12.41 && 5.70 \\
~ & ~ & VideoMix && 47.38 && 21.11 & 17.68 & 9.53 && 16.78 & 15.57 & 10.92 && 5.49 \\
~ & ~ & SDN && 44.29 && 18.05 & 13.07 & 11.49 && 16.31 & \textbf{12.07} & 11.21 && 3.06 \\
~ & ~ & BE$\dagger$ && 45.10 && 19.40 & 17.21 & 8.54 && \textbf{13.77} & 14.35 & \textbf{8.52} && 7.73 \\
~ & ~ & ActorCutMix$\dagger$ && 49.21 && 28.40 & 25.16 & 15.67 && 21.64 & 20.30 & 15.53 && 8.34 \\
~ & ~ & FAME$\dagger$ && 45.97 && 27.83 & 27.29 & 17.12 && 20.26 & 20.86 & 14.87 && 10.18 \\
~ & ~ & \cellcolor{lightgray}\sysname{} (Ours) &\cellcolor{lightgray}& \cellcolor{lightgray}\textbf{51.53} &\cellcolor{lightgray}& \cellcolor{lightgray}\textbf{33.55} & \cellcolor{lightgray}\textbf{32.14} & \cellcolor{lightgray}\textbf{25.81} &\cellcolor{lightgray}& \cellcolor{lightgray}14.34 & \cellcolor{lightgray}13.68 & \cellcolor{lightgray}12.59 &\cellcolor{lightgray}& \cellcolor{lightgray}\textbf{21.21} \\
\midrule
\multirow{8}{*}{Swin-T} & \multirow{8}{*}{ImageNet} & No && 53.62 && 23.45 & 19.39 & 18.24 && 18.25 & 14.80 & 15.94 && 6.56 \\
~ & ~ & Mixup && 55.86 && 25.42 & 18.96 & 17.12 && 20.53 & 14.50 & 15.22 && 6.66 \\
~ & ~ & VideoMix && \textbf{56.17} && 26.31 & 21.98 & 18.02 && 19.60 & 17.24 & 16.14 && 7.89 \\
~ & ~ & SDN && 53.16 && 22.29 & 18.28 & 17.59 && 19.71 & 15.98 & 15.69 && 4.96 \\
~ & ~ & BE$\dagger$ && 53.90 && 23.32 & 17.64 & 14.49 && 18.74 & 13.88 & 13.46 && 5.51 \\
~ & ~ & ActorCutMix$\dagger$ && 54.07 && 29.23 & 25.16 & 22.59 && 22.36 & 19.08 & 17.75 && 8.84 \\
~ & ~ & FAME$\dagger$ && 53.18 && 22.29 & 26.46 & 23.88 && 23.15 & 19.91 & 19.39 && 9.42 \\
~ & ~ & \cellcolor{lightgray}\sysname{} (Ours) &\cellcolor{lightgray}& \cellcolor{lightgray}55.36 &\cellcolor{lightgray}& \cellcolor{lightgray}\textbf{34.73} & \cellcolor{lightgray}\textbf{30.84} & \cellcolor{lightgray}\textbf{30.83} &\cellcolor{lightgray}& \cellcolor{lightgray}\textbf{16.13} & \cellcolor{lightgray}\textbf{11.60} & \cellcolor{lightgray}\textbf{13.20} &\cellcolor{lightgray}& \cellcolor{lightgray}\textbf{21.40} \\
\bottomrule
\end{tabular}
\vspace{4em}
\end{center}
\end{table*}

\begin{table*}[tb]
\begin{center}
\caption{Action recognition accuracy (\%) of different augmentation and debiasing methods on the UCF101, UCF101-\datasetnameA{} and UCF101-\datasetnameB{} datasets. K400 denotes Kinetics-400. $\dagger$ indicates adapting from self-supervised debiasing methods.}
\label{ss_tab:res_aug_ucf101}
\footnotesize
\setlength\tabcolsep{3pt}
\begin{tabular}{@{}cclp{0.01cm}cp{0.01cm}cccp{0.01cm}cccp{0.01cm}c@{}}
\toprule
\multirow{2}{*}{Model} & \multirow{2}{*}{Pretrain} & \multirow{2}{*}{\begin{tabular}[x]{@{}l@{}} Augmentation\\or Debiasing\end{tabular}} && \multirow{2}{*}{UCF101} && \multicolumn{3}{c}{UCF101-\datasetnameA{} ($\uparrow$)} && \multicolumn{3}{c}{UCF101-\datasetnameB{} ($\downarrow$)} && \multirow{2}{*}{\begin{tabular}[x]{@{}c@{}}Contra.\\Acc.\end{tabular}($\big\uparrow$)} \\
\cmidrule{7-9} \cmidrule{11-13}
~ & ~ & ~ && ~ && Place365 & VQGAN-CLIP & Sinusoid && Place365 & VQGAN-CLIP & Sinusoid \\
\midrule\midrule
\multirow{8}{*}{TSM} & \multirow{8}{*}{K400} & No && 94.62 && 26.79 & 22.66 & 27.36 && 5.63 & 4.12 & 2.89 && 21.83 \\
~ & ~ & Mixup && \textbf{94.71} && 29.03 & 24.85 & 29.51 && 5.20 & 3.94 & 2.99 && 24.17 \\
~ & ~ & VideoMix && 94.50 && 32.76 & 29.99 & 31.89 && 6.73 & 5.63 & 4.94 && 26.69 \\
~ & ~ & SDN && 93.83 && 22.15 & 18.37 & 19.22 && 3.46 & 2.53 & 3.32 && 17.19 \\
~ & ~ & BE$\dagger$ && 94.49 && 27.25 & 22.99 & 27.52 && 6.07 & 4.42 & 3.36 && 21.82 \\
~ & ~ & ActorCutMix$\dagger$ && 94.47 && \textbf{38.95} & 37.63 & 37.74 && 4.84 & 4.85 & 3.99 && 33.90 \\
~ & ~ & FAME$\dagger$ && 93.73 && 36.80 & \textbf{37.76} & 32.61 && 4.63 & 4.10 & 2.28 && 32.28 \\
~ & ~ & \cellcolor{lightgray}\sysname{} (Ours) &\cellcolor{lightgray}& \cellcolor{lightgray}94.30 &\cellcolor{lightgray}& \cellcolor{lightgray}37.40 & \cellcolor{lightgray}33.85 & \cellcolor{lightgray}\textbf{40.30} &\cellcolor{lightgray}& \cellcolor{lightgray}\textbf{0.97} & \cellcolor{lightgray}\textbf{0.81} & \cellcolor{lightgray}\textbf{0.60} &\cellcolor{lightgray}& \cellcolor{lightgray}\textbf{36.47} \\
\midrule
\multirow{8}{*}{SlowFast} & \multirow{8}{*}{K400} & No && 95.96 && 34.34 & 31.00 & 30.19 && 2.44 & 1.51 & 1.14 && 30.25 \\
~ & ~ & Mixup && \textbf{96.14} && 36.60 & 33.20 & 32.58 && 4.50 & 2.89 & 2.81 && 30.94 \\
~ & ~ & VideoMix && 95.98 && 38.71 & 39.90 & 31.03 && 4.85 & 3.86 & 3.80 && 31.57 \\
~ & ~ & SDN && 95.02 && 32.24 & 29.25 & 24.32 && 4.64 & 3.02 & 1.95 && 25.72 \\
~ & ~ & BE$\dagger$ && 95.98 && 35.24 & 31.31 & 30.66 && 3.02 & 2.09 & 1.72 && 30.24 \\
~ & ~ & ActorCutMix$\dagger$ && 95.76 && \textbf{47.69} & \textbf{51.69} & \textbf{45.12} && 7.43 & 5.96 & 6.68 && \textbf{42.04} \\
~ & ~ & FAME$\dagger$ && 95.69 && 39.22 & 40.82 & 30.63 && 4.42 & 3.68 & 3.03 && 33.31 \\
~ & ~ & \cellcolor{lightgray}\sysname{} (Ours) &\cellcolor{lightgray}& \cellcolor{lightgray}95.85 &\cellcolor{lightgray}& \cellcolor{lightgray}43.15 & \cellcolor{lightgray}39.29 & \cellcolor{lightgray}40.87 &\cellcolor{lightgray}& \cellcolor{lightgray}\textbf{0.07} & \cellcolor{lightgray}\textbf{0.01} & \cellcolor{lightgray}\textbf{0.00} &\cellcolor{lightgray}& \cellcolor{lightgray}41.08 \\
\midrule
\multirow{8}{*}{Swin-T} & \multirow{8}{*}{K400} & No && \textbf{96.21} && 37.63 & 34.37 & 54.94 && 3.48 & 3.02 & 10.82 && 36.82 \\
~ & ~ & Mixup && 96.17 && 39.82 & 40.89 & 57.79 && 2.88 & 3.28 & 11.62 && 40.46 \\
~ & ~ & VideoMix && 96.00 && 28.59 & 37.36 & 58.26 && 7.81 & 11.40 & 20.60 && 29.37 \\
~ & ~ & SDN && 95.76 && 34.78 & 32.56 & 50.40 && 2.21 & 1.42 & 5.30 && 36.42 \\
~ & ~ & BE$\dagger$ && 96.06 && 39.76 & 36.16 & 56.01 && 3.55 & 2.93 & 10.15 && 38.62 \\
~ & ~ & ActorCutMix$\dagger$ && 95.87 && 51.02 & \textbf{55.28} & \textbf{69.53} && 8.00 & 8.43 & 19.32 && 46.87 \\
~ & ~ & FAME$\dagger$ && 95.81 && 40.62 & 44.56 & 37.54 && 5.74 & 6.50 & 6.84 && 35.14 \\
~ & ~ & \cellcolor{lightgray}\sysname{} (Ours) &\cellcolor{lightgray}& \cellcolor{lightgray}96.02 &\cellcolor{lightgray}& \cellcolor{lightgray}\textbf{55.22} & \cellcolor{lightgray}53.68 & \cellcolor{lightgray}65.75 &\cellcolor{lightgray}& \cellcolor{lightgray}\textbf{2.40} & \cellcolor{lightgray}\textbf{2.16} & \cellcolor{lightgray}\textbf{5.76} &\cellcolor{lightgray}& \cellcolor{lightgray}\textbf{54.90}  \\
\midrule\midrule
\multirow{8}{*}{TSM} & \multirow{8}{*}{ImageNet} & No && 84.84 && 13.89 & 8.73 & 9.58 && 7.89 & 4.76 & 6.21 && 6.13 \\
~ & ~ & Mixup && 86.72 && 27.52 & 25.96 & 24.47 && 9.83 & 8.22 & 10.60 && 17.88 \\
~ & ~ & VideoMix && 83.90 && 29.33 & 27.77 & 23.60 && 12.12 & 12.76 & 13.31 && 17.12 \\
~ & ~ & SDN && 80.41 && 10.11 & 6.74 & 6.82 && 3.44 & 2.37 & 2.26 && 5.44 \\
~ & ~ & BE$\dagger$ && 84.42 && 14.03 & 8.29 & 8.48 && 8.70 & 4.67 & 5.69 && 5.64 \\
~ & ~ & ActorCutMix$\dagger$ && 82.42 && \textbf{47.60} & \textbf{51.00} & \textbf{48.84} && 20.47 & 22.33 & 25.29 && \textbf{28.48} \\
~ & ~ & FAME$\dagger$ && 83.03 && 22.95 & 22.35 & 13.20 && 10.38 & 8.74 & 5.76 && 12.70 \\
~ & ~ & \cellcolor{lightgray}\sysname{} (Ours) &\cellcolor{lightgray}& \cellcolor{lightgray}\textbf{87.29} &\cellcolor{lightgray}& \cellcolor{lightgray}28.24 & \cellcolor{lightgray}20.98 & \cellcolor{lightgray}25.99 &\cellcolor{lightgray}& \cellcolor{lightgray}\textbf{0.42} & \cellcolor{lightgray}\textbf{0.30} & \cellcolor{lightgray}\textbf{1.21} &\cellcolor{lightgray}& \cellcolor{lightgray}24.60 \\
\midrule
\multirow{8}{*}{SlowFast} & \multirow{8}{*}{ImageNet} & No && 80.82 && 15.14 & 11.37 & 8.91 && 6.63 & 3.90 & 3.45 && 8.15 \\
~ & ~ & Mixup && 83.54 && 20.95 & 18.40 & 16.53 && 6.75 & 5.62 & 5.63 && 13.56 \\
~ & ~ & VideoMix && 81.26 && 20.09 & 19.90 & 19.88 && 7.93 & 8.21 & 9.17 && 14.01 \\
~ & ~ & SDN && 78.07 && 13.44 & 8.76 & 8.37 && 5.49 & 2.99 & 2.79 && 6.65 \\
~ & ~ & BE$\dagger$ && 81.51 && 16.36 & 11.99 & 8.45 && 6.72 & 4.11 & 3.07 && 8.55 \\
~ & ~ & ActorCutMix$\dagger$ && 81.54 && \textbf{30.71} & \textbf{28.50} & 21.63 && 8.38 & 6.61 & 6.18 && \textbf{20.48} \\
~ & ~ & FAME$\dagger$ && 80.82 && 22.37 & 23.09 & 15.83 && 7.11 & 6.68 & 4.24 && 15.54 \\
~ & ~ & \cellcolor{lightgray}\sysname{} (Ours) &\cellcolor{lightgray}& \cellcolor{lightgray}\textbf{84.96} &\cellcolor{lightgray}& \cellcolor{lightgray}20.42 & \cellcolor{lightgray}17.15 & \cellcolor{lightgray}\textbf{21.77} &\cellcolor{lightgray}& \cellcolor{lightgray}\textbf{0.01} & \cellcolor{lightgray}\textbf{0.01} & \cellcolor{lightgray}\textbf{0.07} &\cellcolor{lightgray}& \cellcolor{lightgray}19.76 \\
\midrule
\multirow{8}{*}{Swin-T} & \multirow{8}{*}{ImageNet} & No && 88.20 && 19.24 & 16.97 & 22.01 && 8.76 & 7.45 & 10.53 && 11.81 \\
~ & ~ & Mixup && 88.34 && 25.73 & 24.55 & 33.16 && 7.37 & 5.94 & 10.03 && 20.85 \\
~ & ~ & VideoMix && 88.45 && 33.28 & 42.35 & 44.98 && 17.31 & 23.16 & 24.02 && 21.63 \\
~ & ~ & SDN && 85.75 && 13.60 & 11.45 & 20.03 && 6.91 & 5.04 & 9.91 && 9.36 \\
~ & ~ & BE$\dagger$ && 87.80 && 19.30 & 16.31 & 20.64 && 9.06 & 7.09 & 9.09 && 11.38 \\
~ & ~ & ActorCutMix$\dagger$ && 88.73 && \textbf{55.59} & \textbf{59.83} & \textbf{59.88} && 23.30 & 29.87 & 29.48 && \textbf{32.77} \\
~ & ~ & FAME$\dagger$ && 86.00 && 27.41 & 30.62 & 21.05 && 7.13 & 8.50 & 4.95 && 20.11 \\
~ & ~ & \cellcolor{lightgray}\sysname{} (Ours) &\cellcolor{lightgray}& \cellcolor{lightgray}\textbf{88.92} &\cellcolor{lightgray}& \cellcolor{lightgray}32.16 & \cellcolor{lightgray}31.31 & \cellcolor{lightgray}36.91 &\cellcolor{lightgray}& \cellcolor{lightgray}\textbf{1.08} & \cellcolor{lightgray}\textbf{1.54} & \cellcolor{lightgray}\textbf{1.66} &\cellcolor{lightgray}& \cellcolor{lightgray}32.14  \\
\bottomrule
\end{tabular}
\vspace{5em}
\end{center}
\end{table*}

\begin{table*}[!ht]
\begin{center}
\caption{Action recognition accuracy (\%) of different methods on the Kinetics-400, Kinetics400-\datasetnameA{} and Kinetics400-\datasetnameB{} datasets.}
\label{ss_tab:res_k400}
\footnotesize
\setlength\tabcolsep{1pt}
\begin{tabular}{@{}ccccp{0.01cm}cp{0.01cm}cccp{0.01cm}cccp{0.01cm}c@{}}
\toprule
\multicolumn{2}{c}{\multirow{2}{*}{Method}} & \multirow{2}{*}{Pretraining} & \multirow{2}{*}{Debiasing} && \multirow{2}{*}{Kinetics400} && \multicolumn{3}{c}{Kinetics400-\datasetnameA{} ($\uparrow$)} && \multicolumn{3}{c}{Kinetics400-\datasetnameB{} ($\downarrow$)} && \multirow{2}{*}{\begin{tabular}[x]{@{}c@{}}Contra.\\Acc.\end{tabular}($\big\uparrow$)} \\
\cmidrule{8-10} \cmidrule{12-14}
~ & ~ & ~ & ~ && ~ && Place365 & VQGAN-CLIP & Sinusoid && Place365 & VQGAN-CLIP & Sinusoid \\
\midrule\midrule
\multirow{7}{*}{\begin{tabular}[x]{@{}c@{}} Supervised\\Action\\Recognition\\Models\end{tabular}} & \multirow{2}{*}{TSM} & \multirow{2}{*}{ImageNet} & - && 71.13 && 40.44 & 37.39 & 34.34 && 18.55 & 16.61 & 16.49 && 22.80 \\
~ & ~ & ~ & \cellcolor{lightgray} \sysname{} &\cellcolor{lightgray}& \cellcolor{lightgray}71.28 &\cellcolor{lightgray}& \cellcolor{lightgray}43.31 & \cellcolor{lightgray}40.97 & \cellcolor{lightgray}37.15 &\cellcolor{lightgray}& \cellcolor{lightgray}6.27 & \cellcolor{lightgray}5.17 & \cellcolor{lightgray}4.24 &\cellcolor{lightgray}& \cellcolor{lightgray}36.07 \\
\cmidrule{2-16}
~ & \multirow{2}{*}{SlowFast} & \multirow{2}{*}{ImageNet} & - && 65.63 && 36.96 & 35.54 & 34.88 && 21.74 & 20.19 & 20.51 && 18.98 \\
~ & ~ & ~ & \cellcolor{lightgray}\sysname{} &\cellcolor{lightgray}& \cellcolor{lightgray}65.65 &\cellcolor{lightgray}& \cellcolor{lightgray}37.63 & \cellcolor{lightgray}35.71 & \cellcolor{lightgray}35.46 &\cellcolor{lightgray}& \cellcolor{lightgray}14.78 & \cellcolor{lightgray}13.35 & \cellcolor{lightgray}12.70 &\cellcolor{lightgray}& \cellcolor{lightgray}25.01 \\
\cmidrule{2-16}
~ & \multirow{2}{*}{Swin-T} & \multirow{2}{*}{ImageNet} & - && 73.95 && 40.81 & 39.67 & 44.75 && 17.89 & 15.89 & 20.73 && 25.93 \\
~ & ~ & ~ & \cellcolor{lightgray}\sysname{} &\cellcolor{lightgray}& \cellcolor{lightgray}73.86 &\cellcolor{lightgray}& \cellcolor{lightgray}43.44 & \cellcolor{lightgray}42.81 & \cellcolor{lightgray}46.05 &\cellcolor{lightgray}& \cellcolor{lightgray}4.76 & \cellcolor{lightgray}4.37 & \cellcolor{lightgray}7.41 &\cellcolor{lightgray}& \cellcolor{lightgray}39.41 \\
\midrule
Debiasing & FAME & K400 & - && 70.95 && 37.10 & 36.34 & 38.20 && 18.15 & 16.10 & 17.01 && 23.14 \\
\midrule
Self-supervised & VideoMAE & K400 & - && 80.00 && 50.68 & 49.41 & 57.26 && 23.41 & 22.65 & 27.98 && 29.76 \\
\midrule
Multi-modal & X-CLIP & Web+K400 & - && 84.13 && 53.55 & 55.53 & 59.26 && 32.14 & 32.73 & 35.55 && 25.34 \\
\bottomrule
\end{tabular}
\end{center}
\end{table*}

\begin{table*}[!ht]
\begin{center}
\caption{Action recognition accuracy (\%) of different methods on the HMDB51, HMDB51-\datasetnameA{} and HMDB51-\datasetnameB{} datasets. K400 denotes Kinetics-400. Mini-K200 denotes Mini-Kinetics-200 \cite{xie2017rethinking}. $\dagger$ denotes zero-shot classification.}
\label{ss_tab:res_hmdb51}
\footnotesize
\setlength\tabcolsep{1pt}
\begin{tabular}{@{}ccccp{0.01cm}cp{0.01cm}cccp{0.01cm}cccp{0.01cm}c@{}}
\toprule
\multicolumn{2}{c}{\multirow{2}{*}{Method}} & \multirow{2}{*}{Pretrain} & \multirow{2}{*}{Debiasing} && \multirow{2}{*}{HMDB51} && \multicolumn{3}{c}{HMDB51-\datasetnameA{} ($\uparrow$)} && \multicolumn{3}{c}{HMDB51-\datasetnameB{} ($\downarrow$)} && \multirow{2}{*}{\begin{tabular}[x]{@{}c@{}}Contra.\\Acc.\end{tabular}($\big\uparrow$)} \\
\cmidrule{8-10} \cmidrule{12-14}
~ & ~ & ~ & ~ && ~ && Place365 & VQGAN-CLIP & Sinusoid && Place365 & VQGAN-CLIP & Sinusoid \\
\midrule\midrule
\multirow{14}{*}{\begin{tabular}[x]{@{}c@{}} Supervised\\Action\\Recognition\\Models\end{tabular}} & \multirow{4}{*}{TSM} & \multirow{2}{*}{ImageNet} & - && 47.56 && 19.39 & 16.99 & 8.49 && 11.78 & 11.12 & 6.41 && 6.50 \\
~ & ~ & ~ & \cellcolor{lightgray}\sysname{} &\cellcolor{lightgray}& \cellcolor{lightgray}54.66 &\cellcolor{lightgray}& \cellcolor{lightgray}39.52 & \cellcolor{lightgray}38.41 & \cellcolor{lightgray}33.00 &\cellcolor{lightgray}& \cellcolor{lightgray}6.98 & \cellcolor{lightgray}6.09 & \cellcolor{lightgray}3.58 &\cellcolor{lightgray}& \cellcolor{lightgray}33.14 \\
\cmidrule{3-16}
~ & ~ & \multirow{2}{*}{K400} & - && 70.39 && 45.09 & 42.16 & 26.84 && 23.26 & 20.03 & 14.40 && 22.02 \\
~ & ~ & ~ & \cellcolor{lightgray}\sysname{} &\cellcolor{lightgray}& \cellcolor{lightgray}71.52 &\cellcolor{lightgray}& \cellcolor{lightgray}53.91 & \cellcolor{lightgray}52.66 & \cellcolor{lightgray}38.13 &\cellcolor{lightgray}& \cellcolor{lightgray}11.63 & \cellcolor{lightgray}8.29 & \cellcolor{lightgray}5.38 &\cellcolor{lightgray}& \cellcolor{lightgray}42.05 \\
\cmidrule{2-16}
~ & \multirow{4}{*}{SlowFast} & \multirow{2}{*}{ImageNet} & - && 47.65 && 21.24 & 16.85 & 10.43 && 17.84 & 15.63 & 11.76 && 5.26 \\
~ & ~ & ~ & \cellcolor{lightgray}\sysname{} &\cellcolor{lightgray}& \cellcolor{lightgray}51.53 &\cellcolor{lightgray}& \cellcolor{lightgray}33.55 & \cellcolor{lightgray}32.14 & \cellcolor{lightgray}25.81 &\cellcolor{lightgray}& \cellcolor{lightgray}14.34 & \cellcolor{lightgray}13.68 & \cellcolor{lightgray}12.59 &\cellcolor{lightgray}& \cellcolor{lightgray}21.21 \\
\cmidrule{3-16}
~ & ~ & \multirow{2}{*}{K400} & - && 76.25 && 47.36 & 48.23 & 35.00 && 23.26 & 23.14 & 18.78 && 28.37 \\
~ & ~ & ~ & \cellcolor{lightgray}\sysname{} &\cellcolor{lightgray}& \cellcolor{lightgray}76.52 &\cellcolor{lightgray}& \cellcolor{lightgray}48.31 & \cellcolor{lightgray}47.75 & \cellcolor{lightgray}40.31 &\cellcolor{lightgray}& \cellcolor{lightgray}17.76 & \cellcolor{lightgray}17.51 & \cellcolor{lightgray}15.65 &\cellcolor{lightgray}& \cellcolor{lightgray}35.20 \\
\cmidrule{2-16}
~ & \multirow{4}{*}{Swin-T} & \multirow{2}{*}{ImageNet} & - && 53.62 && 23.45 & 19.39 & 18.24 && 18.25 & 14.80 & 15.94 && 6.56 \\
~ & ~ & ~ & \cellcolor{lightgray}\sysname{} &\cellcolor{lightgray}& \cellcolor{lightgray}55.36 &\cellcolor{lightgray}& \cellcolor{lightgray}34.73 & \cellcolor{lightgray}30.84 & \cellcolor{lightgray}30.83 &\cellcolor{lightgray}& \cellcolor{lightgray}16.13 & \cellcolor{lightgray}11.60 & \cellcolor{lightgray}13.20 &\cellcolor{lightgray}& \cellcolor{lightgray}21.40 \\
\cmidrule{3-16}
~ & ~ & \multirow{2}{*}{K400} & - && 73.92 && 47.61 & 42.77 & 41.41 && 20.68 & 17.90 & 22.80 && 27.84 \\
~ & ~ & ~ & \cellcolor{lightgray}\sysname{} &\cellcolor{lightgray}& \cellcolor{lightgray}74.82 &\cellcolor{lightgray}& \cellcolor{lightgray}53.27 & \cellcolor{lightgray}52.43 & \cellcolor{lightgray}49.73 &\cellcolor{lightgray}& \cellcolor{lightgray}13.39 & \cellcolor{lightgray}12.66 & \cellcolor{lightgray}14.13 &\cellcolor{lightgray}& \cellcolor{lightgray}40.28 \\
\midrule
\multirow{2}{*}{\begin{tabular}[x]{@{}c@{}} Debiasing\\Pretraining\end{tabular}} & SDN & Mini-K200 & - && 56.60 && 26.76 & 23.48 & 11.13 && 14.80 & 14.69 & 4.96 && 10.83 \\
\cmidrule{2-16}
~ & FAME & K400 & - && 61.10 && 31.45 & 28.67 & 25.12 && 13.28 & 13.67 & 12.93 && 17.21 \\
\midrule
Self-supervised & VideoMAE & HMDB51 & - && 62.60 && 23.24 & 27.19 & 18.55 && 10.58 & 10.43 & 10.94 && 15.01 \\
\midrule
Multi-modal & X-CLIP$\dagger$ & Web+K400 & - && 49.67 && 22.50 & 25.47 & 27.03 && 18.52 & 20.31 & 21.37 && 9.31 \\
\bottomrule
\end{tabular}
\end{center}
\end{table*}

\begin{table*}[!ht]
\begin{center}
\caption{Action recognition accuracy (\%) of different methods on the UCF101, UCF101-\datasetnameA{} and UCF101-\datasetnameB{} datasets. K400 denotes Kinetics-400. Mini-K200 denotes Mini-Kinetics-200 \cite{xie2017rethinking}. $\dagger$ means zero-shot classification.}
\label{ss_tab:res_ucf101}
\footnotesize
\setlength\tabcolsep{1pt}
\begin{tabular}{@{}ccccp{0.01cm}cp{0.01cm}cccp{0.01cm}cccp{0.01cm}c@{}}
\toprule
\multicolumn{2}{c}{\multirow{2}{*}{Method}} & \multirow{2}{*}{Pretraining} & \multirow{2}{*}{Debiasing} && \multirow{2}{*}{UCF101} && \multicolumn{3}{c}{UCF101-\datasetnameA{} ($\uparrow$)} && \multicolumn{3}{c}{UCF101-\datasetnameB{} ($\downarrow$)} && \multirow{2}{*}{\begin{tabular}[x]{@{}c@{}}Contra.\\Acc.\end{tabular}($\big\uparrow$)} \\
\cmidrule{8-10} \cmidrule{12-14}
~ & ~ & ~ & ~ && ~ && Place365 & VQGAN-CLIP & Sinusoid && Place365 & VQGAN-CLIP & Sinusoid \\
\midrule\midrule
\multirow{14}{*}{\begin{tabular}[x]{@{}c@{}} Supervised\\Action\\Recognition\\Models\end{tabular}} & \multirow{4}{*}{TSM} & \multirow{2}{*}{ImageNet} & - && 84.84 && 13.89 & 8.73 & 9.58 && 7.89 & 4.76 & 6.21 && 6.13 \\
~ & ~ & ~ & \cellcolor{lightgray} \sysname{} &\cellcolor{lightgray}& \cellcolor{lightgray}87.29 &\cellcolor{lightgray}& \cellcolor{lightgray}28.24 & \cellcolor{lightgray}20.98 & \cellcolor{lightgray}25.99 &\cellcolor{lightgray}& \cellcolor{lightgray}0.42 & \cellcolor{lightgray}0.30 & \cellcolor{lightgray}1.21 &\cellcolor{lightgray}& \cellcolor{lightgray}24.60 \\
\cmidrule{3-16}
~ & ~ & \multirow{2}{*}{K400} & - && 94.62 && 26.79 & 22.66 & 27.36 && 5.63 & 4.12 & 2.89 && 21.83 \\
~ & ~ & ~ & \cellcolor{lightgray} \sysname{} &\cellcolor{lightgray}& \cellcolor{lightgray}94.30 &\cellcolor{lightgray}& \cellcolor{lightgray}37.40 & \cellcolor{lightgray}33.85 & \cellcolor{lightgray}40.30 &\cellcolor{lightgray}& \cellcolor{lightgray}0.97 & \cellcolor{lightgray}0.81 & \cellcolor{lightgray}0.60 &\cellcolor{lightgray}& \cellcolor{lightgray}36.47 \\
\cmidrule{2-16}
~ & \multirow{4}{*}{SlowFast} & \multirow{2}{*}{ImageNet} & - && 80.82 && 15.14 & 11.37 & 8.91 && 6.63 & 3.90 & 3.45 && 8.15 \\
~ & ~ & ~ & \cellcolor{lightgray}\sysname{} &\cellcolor{lightgray}& \cellcolor{lightgray}84.96 &\cellcolor{lightgray}& \cellcolor{lightgray}20.42 & \cellcolor{lightgray}17.15 & \cellcolor{lightgray}21.77 &\cellcolor{lightgray}& \cellcolor{lightgray}0.02 & \cellcolor{lightgray}0.01 & \cellcolor{lightgray}0.07 &\cellcolor{lightgray}& \cellcolor{lightgray}19.76 \\
\cmidrule{3-16}
~ & ~ & \multirow{2}{*}{K400} & - && 95.96 && 34.34 & 31.00 & 30.19 && 2.44 & 1.51 & 1.14 && 30.25 \\
~ & ~ & ~ & \cellcolor{lightgray}\sysname{} &\cellcolor{lightgray}& \cellcolor{lightgray}95.85 &\cellcolor{lightgray}& \cellcolor{lightgray}43.15 & \cellcolor{lightgray}39.29 & \cellcolor{lightgray}40.87 &\cellcolor{lightgray}& \cellcolor{lightgray}0.07 & \cellcolor{lightgray}0.01 & \cellcolor{lightgray}0.00 &\cellcolor{lightgray}& \cellcolor{lightgray}41.08 \\
\cmidrule{2-16}
~ & \multirow{4}{*}{Swin-T} & \multirow{2}{*}{ImageNet} & - && 88.20 && 19.24 & 16.97 & 22.01 && 8.76 & 7.45 & 10.53 && 11.81 \\
~ & ~ & ~ & \cellcolor{lightgray}\sysname{} &\cellcolor{lightgray}& \cellcolor{lightgray}88.92 &\cellcolor{lightgray}& \cellcolor{lightgray}32.16 & \cellcolor{lightgray}31.31 & \cellcolor{lightgray}36.91 &\cellcolor{lightgray}& \cellcolor{lightgray}1.08 & \cellcolor{lightgray}1.54 & \cellcolor{lightgray}1.66 &\cellcolor{lightgray}& \cellcolor{lightgray}32.14 \\
\cmidrule{3-16}
~ & ~ & \multirow{2}{*}{K400} & - && 96.21 && 37.63 & 34.37 & 54.94 && 3.48 & 3.02 & 10.82 && 36.82 \\
~ & ~ & ~ & \cellcolor{lightgray}\sysname{} &\cellcolor{lightgray}& \cellcolor{lightgray}96.02 &\cellcolor{lightgray}& \cellcolor{lightgray}55.22 & \cellcolor{lightgray}53.68 & \cellcolor{lightgray}65.75 &\cellcolor{lightgray}& \cellcolor{lightgray}2.40 & \cellcolor{lightgray}2.16 & \cellcolor{lightgray}5.76 &\cellcolor{lightgray}& \cellcolor{lightgray}54.90 \\
\midrule
\multirow{2}{*}{\begin{tabular}[x]{@{}c@{}} Debiasing\\Pretraining\end{tabular}} & SDN & Mini-K200 & - && 84.17 && 10.31 & 7.82 & 8.59 && 1.85 & 1.47 & 1.69 && 7.74 \\
\cmidrule{2-16}
~ & FAME & K400 & - && 88.60 && 18.79 & 19.06 & 15.80 && 1.25 & 1.21 & 1.27 && 17.13 \\
\midrule
Self-supervised & VideoMAE & UCF101 & - && 91.30 && 19.10 & 18.77 & 19.38 && 0.59 & 0.66 & 1.03 && 18.50 \\
\midrule
Multi-modal & X-CLIP$\dagger$ & Web+K400 & - && 74.52 && 24.64 & 28.44 & 37.36 && 16.24 & 16.88 & 20.42 && 15.27 \\
\bottomrule
\end{tabular}
\end{center}
\end{table*}

\subsection{Full Results of Transfer Learning}
To evaluate the robustness of the learned action representations, we conduct tests of transferring action features across datasets. The rationale for this evaluation is that the static bias is likely idiosyncratic to the dataset and may not transfer well across datasets or class definitions. In comparison, the motion features should transfer well across datasets and class definitions. We adopt the linear probing protocol. After training on the source dataset, we fix the backbone network and train only a linear classifier on top of the backbone using the target dataset.

Table~\ref{ss_tab:transfer} shows the performance of different data augmentation and debiasing methods with different base models. From the results, we observe that the models trained with \sysname{} obtain the best performance in different transferring settings, especially in transferring across small datasets. For example, in transferring from HMDB51 to UCF101, \sysname{} outperforms other data augmentation methods by about 2\% of accuracy. These results illustrate that \sysname{} learns robust action representations that have better capability to transfer across action datasets.

\subsection{Ablation Study}
In this section, we provide more results of ablation studies.

\vspace{0.5em}
\noindent\textbf{Sampling biased frames improves debiasing.} In Table \ref{ss_tab:sampling}, we show more results of different frame sampling strategies on HMDB51 using different main networks and pretraining datasets. As in the main paper, we compare three frame sampling strategies: (1) \emph{No RefNet}; (2) \emph{RefNet}; (3) \emph{RefNet Inversed}. Comparing the results of \emph{RefNet Inversed} with the other two strategies, we observe that \emph{RefNet Inversed} obtains significantly lower OOD performance especially for ImageNet pretrained models (more than 3\%). Comparing the results of \emph{No RefNet} and \emph{RefNet}, we observe that they obtain similar IID performance but \emph{RefNet} performs better on OOD tests especially for ImageNet pretrained TSM and Swin-T (more than 2\%). The results show that sampling biased frames benefits bias mitigation. Even sampling frames with \emph{RefNet Inversed} or \emph{No RefNet}, \sysname{} still outperforms other methods (refer to the results in Table \ref{ss_tab:res_aug_hmdb51}), further indicating its effectiveness on bias mitigation.

\vspace{0.5em}
\noindent\textbf{Mixing action labels in \sysname{} decreases performance.} \sysname{} keeps the label unchanged after augmentation. Here, we investigate the effects of mixing action labels, \ie, $\tilde{y}_i=\lambda^{\prime} y_i+(1-\lambda^{\prime}) y^{\text{biased}}$ where $y^{\text{biased}}$ is the action label of the biased frame $\boldsymbol{z}^{\text{biased}}$. In Table \ref{ss_tab:mix_label}, we compare the performance of different values of $\lambda^{\prime}$ on UCF101 and HMDB51 using TSM as the main network. We observe that mixing action labels significantly decreases the OOD performance although it could slightly boost the IID performance for Kinetics400 pretrained TSM by around 0.5\% of accuracy. The results illustrate that mixing action labels in \sysname{} is detrimental to learning robust action representations, since it encourages models to learn biased static cues that are not robust in OOD scenarios.

\vspace{0.5em}
\noindent\textbf{Effects of Beta Distribution in \sysname{}.} With different Beta distribution parameters in \sysname{}, the mixing coefficient $\lambda$ (in Eq. (3) of the main paper) has different values of mean and variance. In Table \ref{ss_tab:beta}, we compare the performance of different Beta distribution parameters on UCF101 and HMDB51 using TSM as the main network. Comparing the results of different mean values of $\lambda$, we observe that both IID and OOD performance decrease when the mean value is large (\eg, 0.75). With large values of $\lambda$, the mixed videos approximate the original videos, so that the debiasing effects are weak. In contrast, small mean values of $\lambda$ (\eg, 0.5, 0.25) lead to good IID and OOD performance. The results indicate that sufficient mixing strength is necessary for \sysname{} to mitigate static bias. Comparing the results of different variances of $\lambda$, we observe that increasing the variances improves the performance. The reason may be that large variances could augment videos with various mixing strength, which creates diverse augmented samples that help training.

\vspace{0.5em}
\noindent\textbf{Effects of $P_{aug}$ in \sysname{}.} We apply StillMix to each video with a predefined probability $P_{aug}$ as in Eq. (3) of the main paper. In Table \ref{ss_tab:Paug}, we compare the performance of different $P_{aug}$ using TSM as the main network. On IID performance, we observe that small $P_{aug}$ obtains better results on UCF101 and Kinetics400 while large $P_{aug}$ obtains better results on HMDB51. We attribute this to the strong correlations between static cues and action class labels in UCF101 and Kinetics400. MMAction2~\cite{2020mmaction2} trained TSN~\cite{8454294} using only three frames per video, achieving 83.03\% classification accuracy on UCF101 and 70.60\% classification accuracy on Kinetics400. However, it only achieved 48.95\% with 8 frames on HMDB51\footnote{\url{https://github.com/open-mmlab/mmaction2/blob/02a06bb3180e951b00ccceb48dab055f95acd1a7/configs/recognition/tsn/README.md}}. These results shows static cues is useful to recognize actions in UCF101 and Kinetics400, on which large $P_{aug}$ discouraging the learning of static cues may result in degraded IID performance. While on HMDB51, the correlations between static cues and action class labels is relatively weak. Therefore, enhancing the learning of motion feature learning with large $P_{aug}$ would improve the performance. On OOD performance, we observe that small $P_{aug}$ obtains better results on UCF101 while large $P_{aug}$ obtains better results on HMDB51 and Kinetics400. We hypothesize this is because UCF101-\datasetnameA{} and UCF101-\datasetnameB{} is constructed from human bounding box annotations. As a result, the synthetic videos contain some background information that is useful for action recognition. Therefore, small $P_{aug}$ allowing learning background static cues may obtain better results. However, large $P_{aug}$ performs better on HMDB51-\datasetnameA{}, HMDB51-\datasetnameB{}, Kinetics400-\datasetnameA{} and Kinetics400-\datasetnameB{} that do not contain background information.

\subsection{Full Results of Debiasing Methods}
Table~\ref{ss_tab:res_aug_k400}, \ref{ss_tab:res_aug_hmdb51} and \ref{ss_tab:res_aug_ucf101} show the full results of different video data augmentation and debiasing methods on \datasetnameA{} and \datasetnameB{} videos of the Kinetics-400, HMDB51 and UCF101 datasets, respectively. The observations are similar to that in the main paper.

\subsection{Evaluation of Pretraining Methods}
In this section, we evaluate several pretraining methods on the synthetic OOD data to demonstrate how pretraining affects OOD generalization. We evaluate the following pretraining methods:

\vspace{0.2em}\noindent\textbf{Debiasing Pretraining Methods:} (1) SDN \cite{choi2019can}, a supervised debiasing pretraining method that minimizes scene information and maximizes human action information using adversarial classifiers. (2) FAME \cite{Ding_2022_CVPR}, a self-supervised debiasing pretraining method which carves out the foreground from the video and replace the background for training to mitigate background bias. We directly use the available pretrained checkpoints for evaluation.

\vspace{0.2em}\noindent\textbf{Self-supervised Pretraining Method:} VideoMAE \cite{tong2022videomae}, a strong self-supervised learner with masked autoencoder.

\vspace{0.2em}\noindent\textbf{Multi-modal Pretraining Model:} X-CLIP \cite{ni2022expanding}, an expanded language-image pretrained model with a video-specific prompting scheme.

\vspace{0.3em}
Table~\ref{ss_tab:res_k400}, \ref{ss_tab:res_hmdb51} and \ref{ss_tab:res_ucf101} compare the IID and OOD performance of different pretraining methods on Kinetics-400, HMDB51 and UCF101, respectively. From the results we make the following observations:

\vspace{0.2em}\noindent\textbf{Pretraining on large video datasets is by itself an effective method to debias action representations.} By comparing the performance of using ImageNet and Kinetics-400 as pretraining datasets in Table~\ref{ss_tab:res_hmdb51} and \ref{ss_tab:res_ucf101}, we observe that K400-pretrained models improve the performance on both IID test and \datasetnameA{} without too much performance sacrifice on \datasetnameB{}. The results demonstrate that pretraining on large video datasets is by itself an effective method to debias action representations. We hypothesize that the size of Kinetics-400 is so large that it contains reasonably balanced static cues. However, collecting, annotating, and training on large-scale datasets are still costly, while simple augmentations could mitigate static bias even for K400-pretrained models; the minimum improvement of \textit{Contra. Acc.} is 6.83\%.

\vspace{0.2em}\noindent\textbf{Debiasing pretraining does not mitigate static bias effectively.} SDN and FAME adopt debiasing pretraining on large datasets and finetuning on small datasets. In Table \ref{ss_tab:res_hmdb51} and \ref{ss_tab:res_ucf101}, SDN obtains comparable performance on IID and OOD test with ImageNet-pretrained models, though it is pretrained on Mini-Kinetics-200. FAME lags behind K400-pretrained models, though it is also pretrained on Kinetics-400. The results indicate that vanilla supervised pretraining is more effective than debiasing pretraining at mitigating static bias. Effective debiasing pretraining deserves further research attention.

\vspace{0.2em}\noindent\textbf{Self-supervised models and multi-modal pretraining models are still vulnerable to static bias.} As powerful video representation learners, VideoMAE and X-CLIP obtain good performance on IID tests and \datasetnameA{}, but the performance on \datasetnameB{} and the \textit{Contra. Acc.} is worse than ImageNet-pretrained Swin-T trained with \sysname{}. The results indicate that they could not effectively mitigate foreground static bias and learn robust action features. How to improve the robustness of action representations through self-supervised pretraining and prompting large language-vision pretrained models still deserves exploration.

\begin{figure*}[ht]
    \centering
    \includegraphics[width=\linewidth]{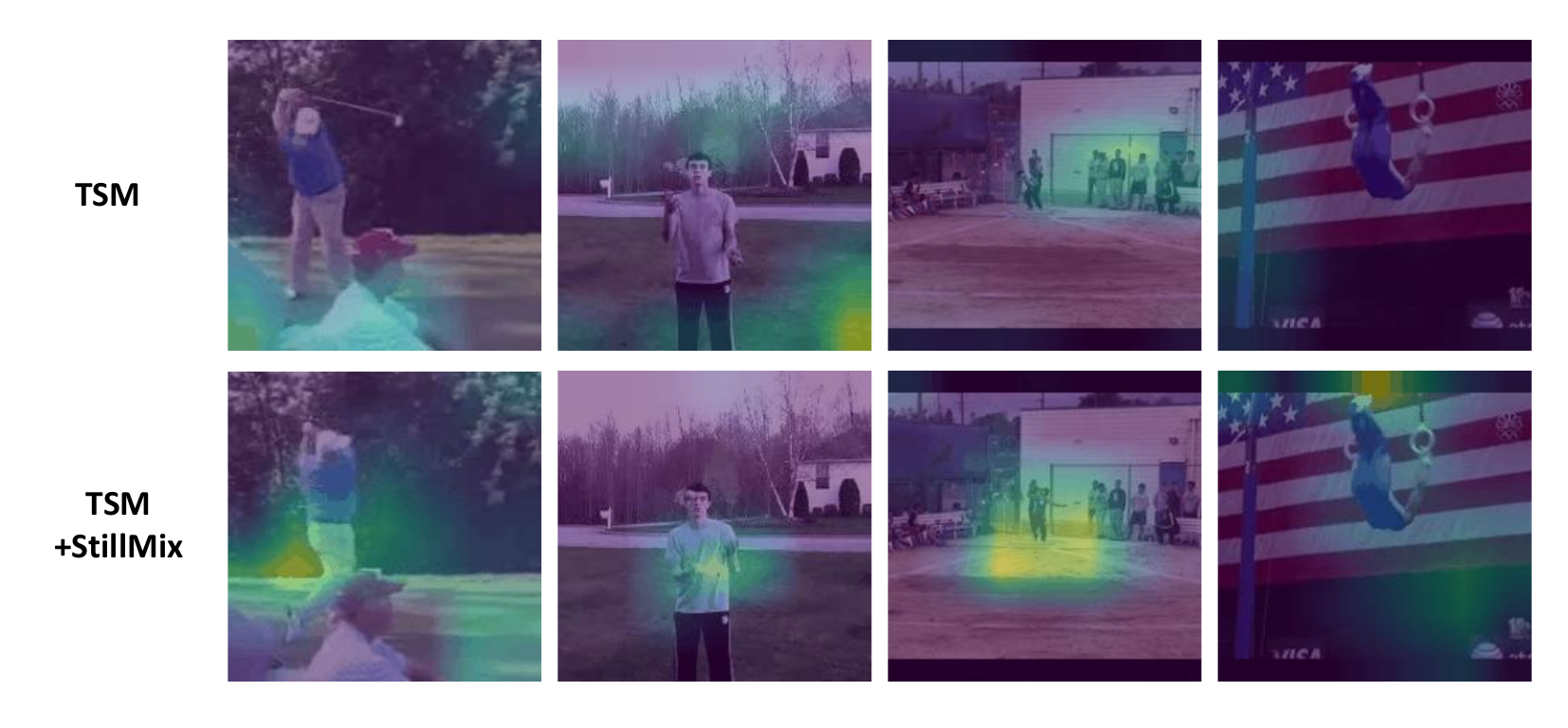}
    \caption{Grad-CAM visualization on videos from UCF101 dataset. The first row shows results of TSM trained without \sysname{}. The second row shows the results of TSM + \sysname{}. \sysname{} helps to focus on the motion regions.}
    \label{ss_fig:vis-gradcam}
\end{figure*}

\subsection{Grad-CAM Visualization}
In Figure \ref{ss_fig:vis-gradcam}, we visualize the Grad-CAM~\cite{selvaraju2017grad} on some videos from UCF101. Trained with \sysname{}, TSM focuses on the regions with motion. For example, in the second column, the action is ``Juggling Balls''. TSM trained without \sysname{} focuses on the background like the grass field and the trees. However, the model trained with \sysname{} learns to focus on the hand motion. The visualization results validate that \sysname{} helps to learn motion representations and mitigate reliance of background static cues.

\section{Construction Details of \datasetnameA{} and \datasetnameB{}}
\label{ss_sec:benchmark}

We synthesize \datasetnameA{} and \datasetnameB{} videos using videos in the test set of the first split of HMDB51~\cite{kuehne2011hmdb} and UCF101~\cite{soomro2012ucf101}, and the validation set of Kinetics-400~\cite{carreira2017quo}.

\subsection{Foreground Masks}
The details of collecting and producing foreground masks of the three datasets are described as follows.

\vspace{0.2em}\noindent\textbf{HMDB51.} We use human-annotated segmentation masks of people for 21 action classes from the JHMDB dataset~\cite{jhuang2013towards}. There are totally 256 videos in the test set of the first split having mask annotations.

\vspace{0.2em}\noindent\textbf{UCF101.} We use human-annotated bounding boxes of people for 24 action classes provided by the Thumos challenge~\cite{idrees2017thumos}. There are totally 910 videos in the test set of the first split having bounding box annotations.

\vspace{0.2em}\noindent\textbf{Kinetics-400.} We decode the validation videos into frames (we use 15 fps as the frame rate) to extract the foreground mask of each frame. Since there is no available human annotation of Kinetics-400, we use video semantic segmentation model VSS-CFFM~\cite{sun2022coarse} and video salient object segmentation model UFO~\cite{su2022unified} to extract foregrounds. In each frame, we extract the human mask from VSS-CFFM and the salient object mask from UFO, and each of them is smoothed using the masks in three adjacent frames, \ie, the union of the three masks is used as the smoothed mask. The foreground mask is the union of the smoothed human mask and the smoothed salient object mask. The videos in which more than 10\% of frames having small foreground masks (\ie, the area of the foreground mask is smaller than 10\% of the area of the whole frame) are discarded. Finally, we use the remaining 10,190 videos in the validation set to construct benchmarks.

\subsection{Background Images}
The details of generating background images by VQGAN-CLIP~\cite{crowson2022vqgan} and sinusoidal functions are described as follows.

\vspace{0.2em}\noindent\textbf{VQGAN-CLIP.} 2,000 background images of artistic style are generated by VQGAN-CLIP. Each image is generated from a sentence with the template: ``A painting / sketch / illustration / photograph of \emph{scene\_name} in the style of \emph{style\_name}''. In the template, the \emph{scene\_name} is the name of a random scene category in Place365~\cite{7968387}; the \emph{style\_name} is a random artistic style sampled from a list: \{``Art Nouveau'', ``Camille Pissarro'', ``Michelangelo Caravaggio'', ``Claude Monet'', ``Edgar Degas'', ``Edvard Munch'', ``Fauvism'', ``Futurism'', ``Impressionism'', ``Picasso'', ``Pop Art'', ``Modern art'', ``Surreal Art'', ``Sandro Botticelli'', ``Oil Paints'', ``Water Colours'', ``Weird Bananas'', ``Strange Colours''\}.

\vspace{0.2em}\noindent\textbf{Sinusoid.} Each stripe in the stripe images are defined by sinusoidal functions $y=A sin(\omega x+\phi)$. The ranges of each parameter are defined as follows:
\begin{itemize}
    \setlength{\itemindent}{0em}
    \setlength{\itemsep}{-0.25em}
    \item $0\leq\omega\leq 0.5\times\pi$
    \item $-100\times\omega\leq\phi\leq 100\times\omega$
    \item $10\leq A\leq 110$
\end{itemize}
The ranges of stripe widths $sw$ and space between two adjacent stripes $sp$ are defined as follows:
\begin{itemize}
    \setlength{\itemindent}{0em}
    \setlength{\itemsep}{-0.25em}
    \item $5\leq sw\leq 20$
    \item $3\times sw\leq sp\leq 5\times sw$
\end{itemize}
Each stripe image is generated by uniformly sampling these parameters from the corresponding ranges. The colors of the stripe areas and the background areas are also randomly chosen. After a stripe image is generated, we further rotate it by a random angle and crop the central area of $224\times 224$ as the final stripe image.

\subsection{Synthetic Videos}
\begin{figure}[t]
\centering
\begin{minipage}{1.0\linewidth}
    \includegraphics[width=0.24\linewidth]{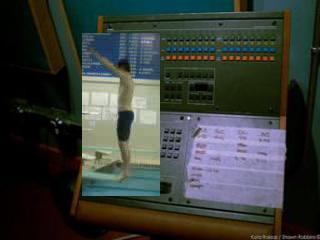}
    \includegraphics[width=0.24\linewidth]{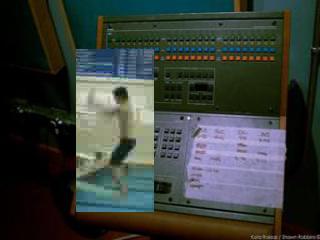}
    \includegraphics[width=0.24\linewidth]{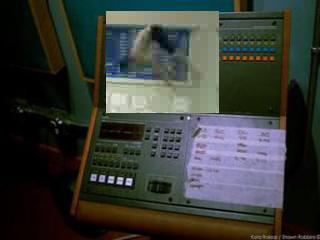}
    \includegraphics[width=0.24\linewidth]{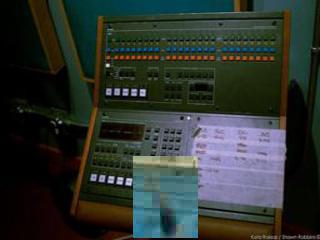}
    \centerline{\textit{Diving} with Place365 backgrounds.}
\end{minipage}
\begin{minipage}{1.0\linewidth}
    \includegraphics[width=0.24\linewidth]{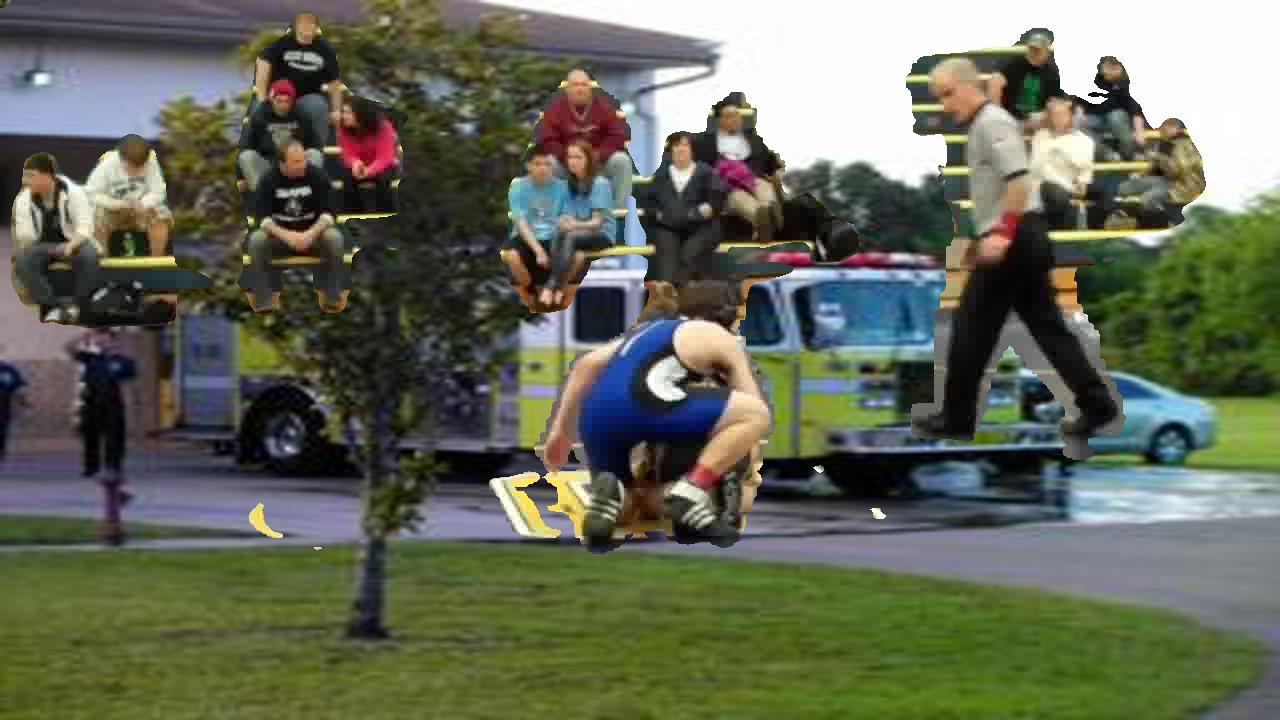}
    \includegraphics[width=0.24\linewidth]{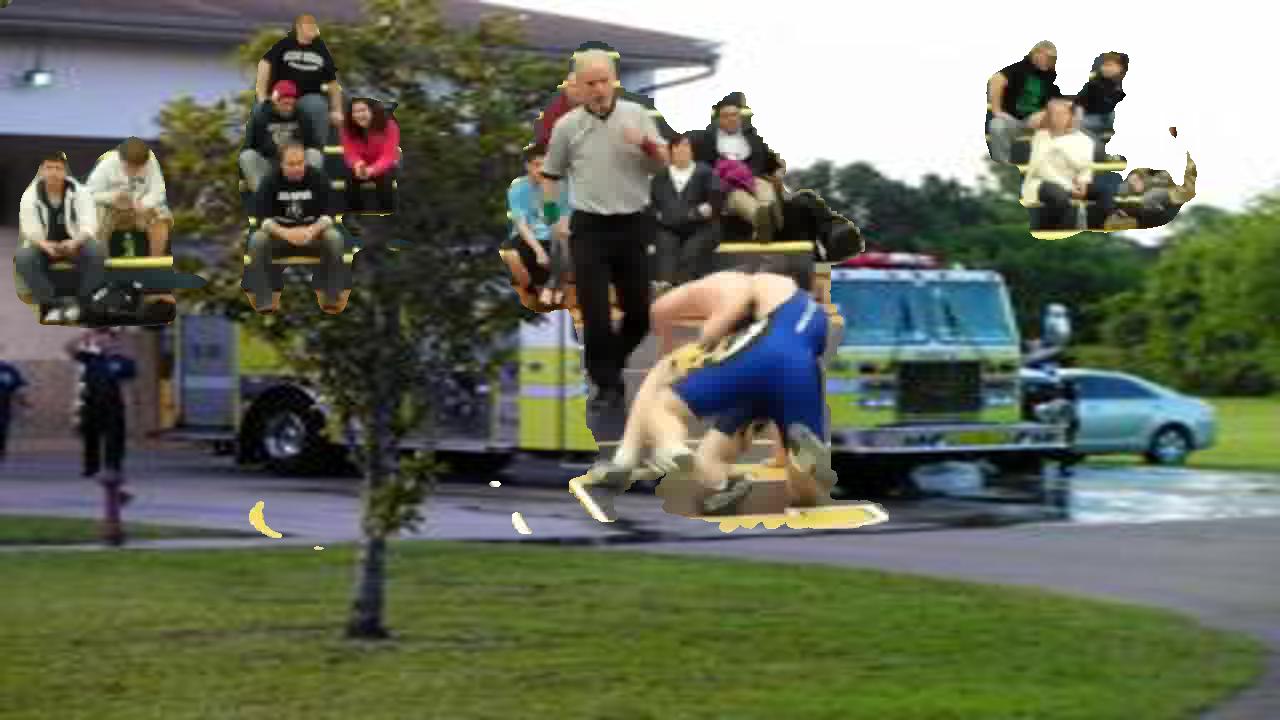}
    \includegraphics[width=0.24\linewidth]{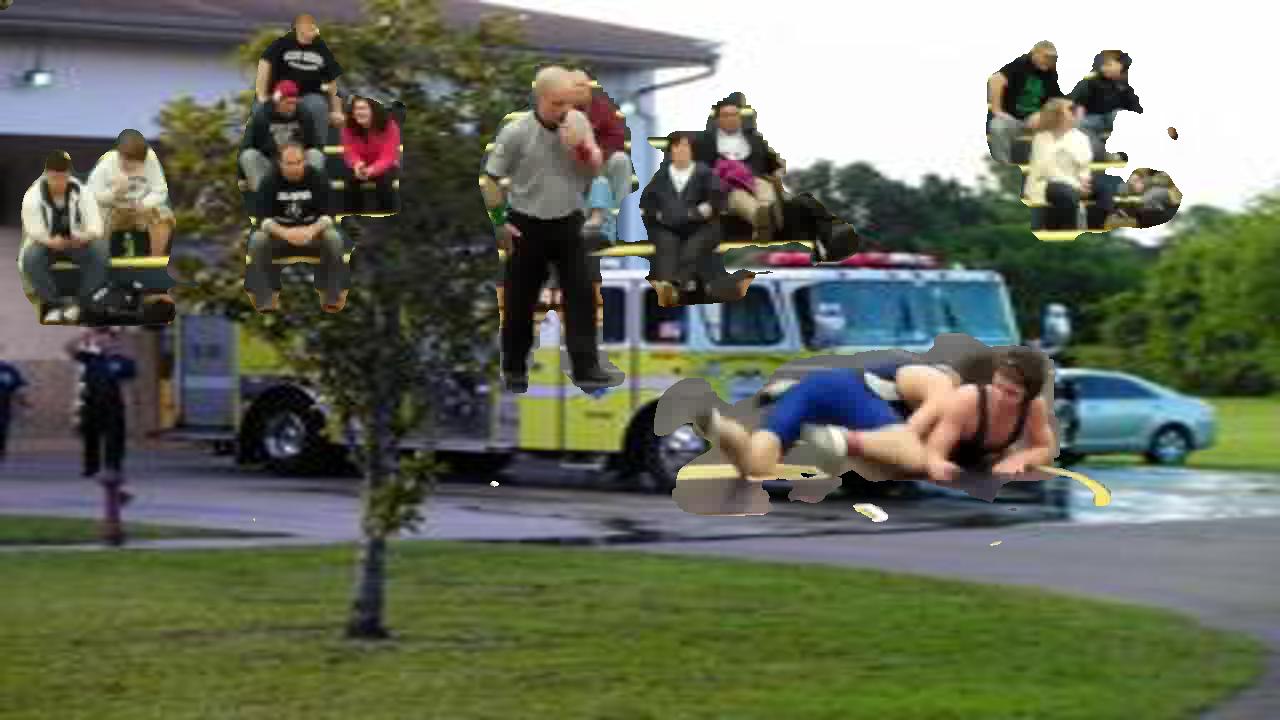}
    \includegraphics[width=0.24\linewidth]{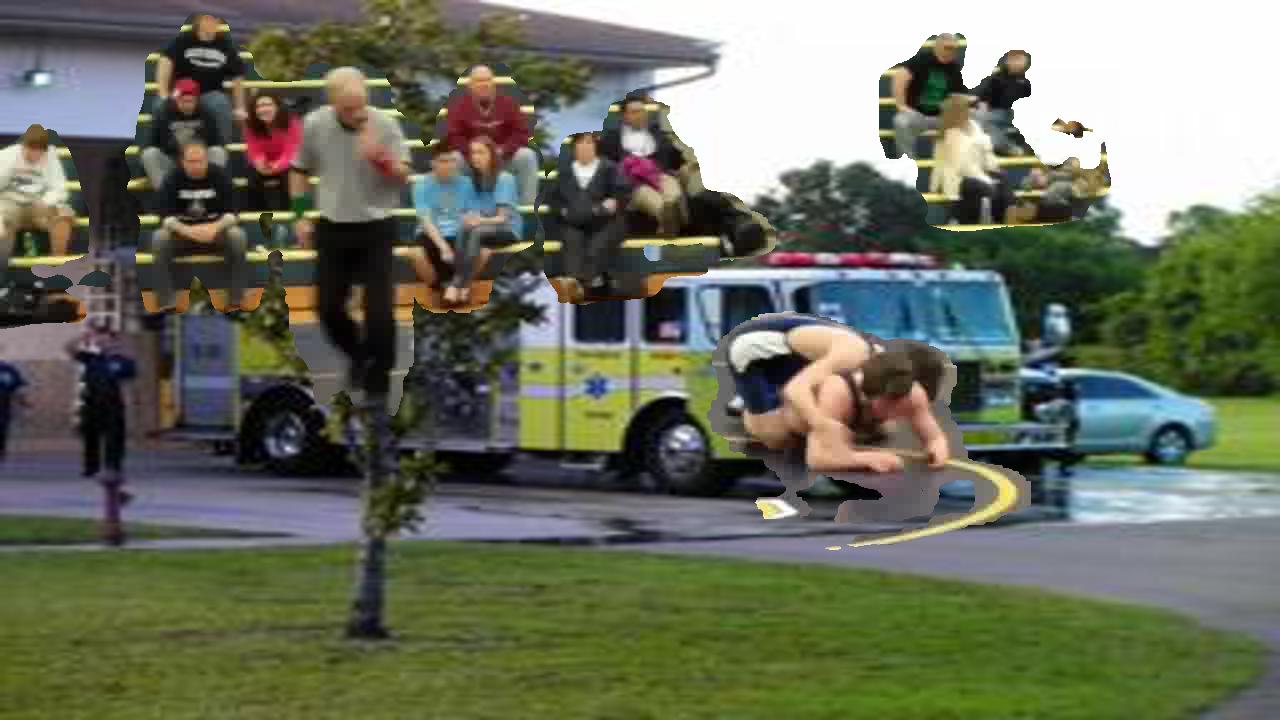}
    \centerline{\textit{Wrestling} with Place365 backgrounds.}
\end{minipage}
\begin{minipage}{1.0\linewidth}
    \includegraphics[width=0.24\linewidth]{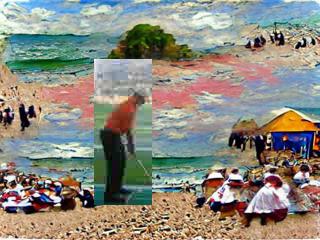}
    \includegraphics[width=0.24\linewidth]{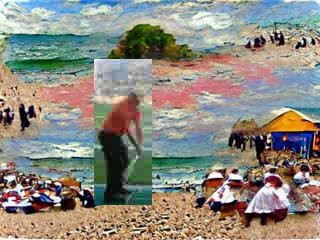}
    \includegraphics[width=0.24\linewidth]{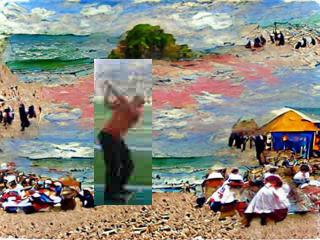}
    \includegraphics[width=0.24\linewidth]{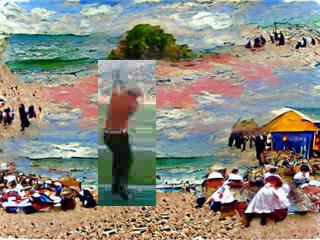}
    \centerline{\textit{GolfSwing} with VQGAN-CLIP backgrounds.}
\end{minipage}
\begin{minipage}{1.0\linewidth}
    \includegraphics[width=0.24\linewidth]{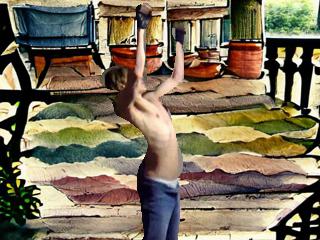}
    \includegraphics[width=0.24\linewidth]{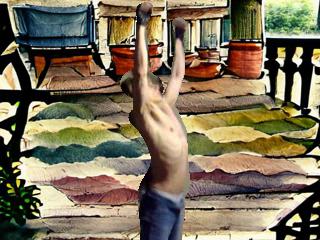}
    \includegraphics[width=0.24\linewidth]{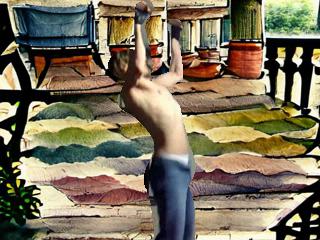}
    \includegraphics[width=0.24\linewidth]{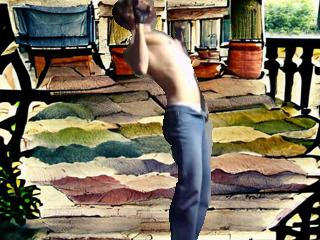}
    \centerline{\textit{Pullup} with VQGAN-CLIP backgrounds.}
\end{minipage}
\begin{minipage}{1.0\linewidth}
    \includegraphics[width=0.24\linewidth]{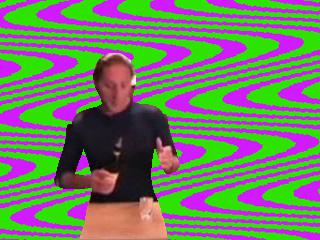}
    \includegraphics[width=0.24\linewidth]{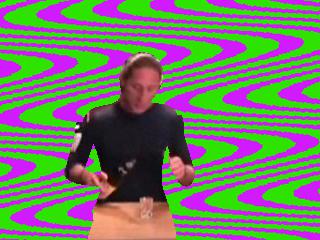}
    \includegraphics[width=0.24\linewidth]{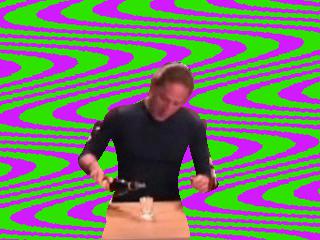}
    \includegraphics[width=0.24\linewidth]{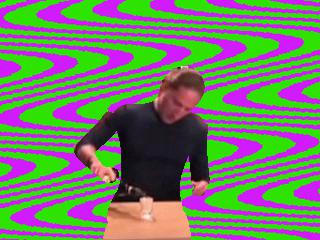}
    \centerline{\textit{Pour} with sinusoidal backgrounds.}
\end{minipage}
\begin{minipage}{1.0\linewidth}
    \includegraphics[width=0.24\linewidth]{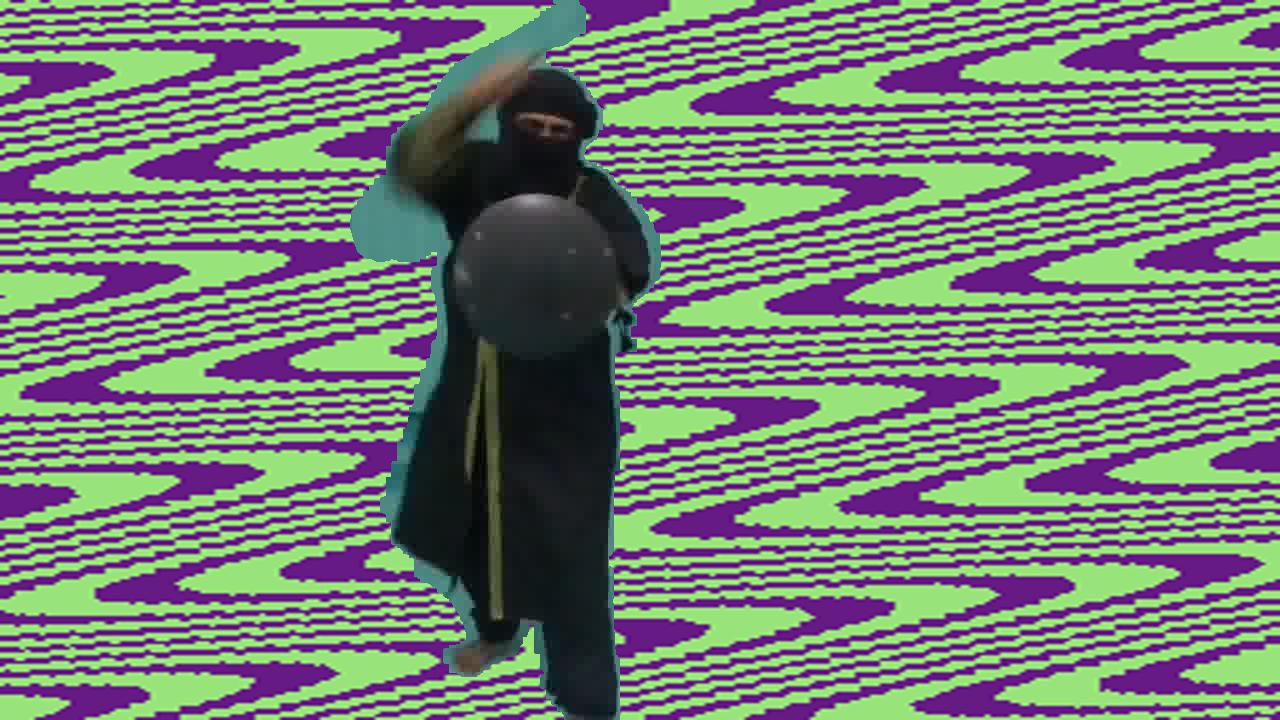}
    \includegraphics[width=0.24\linewidth]{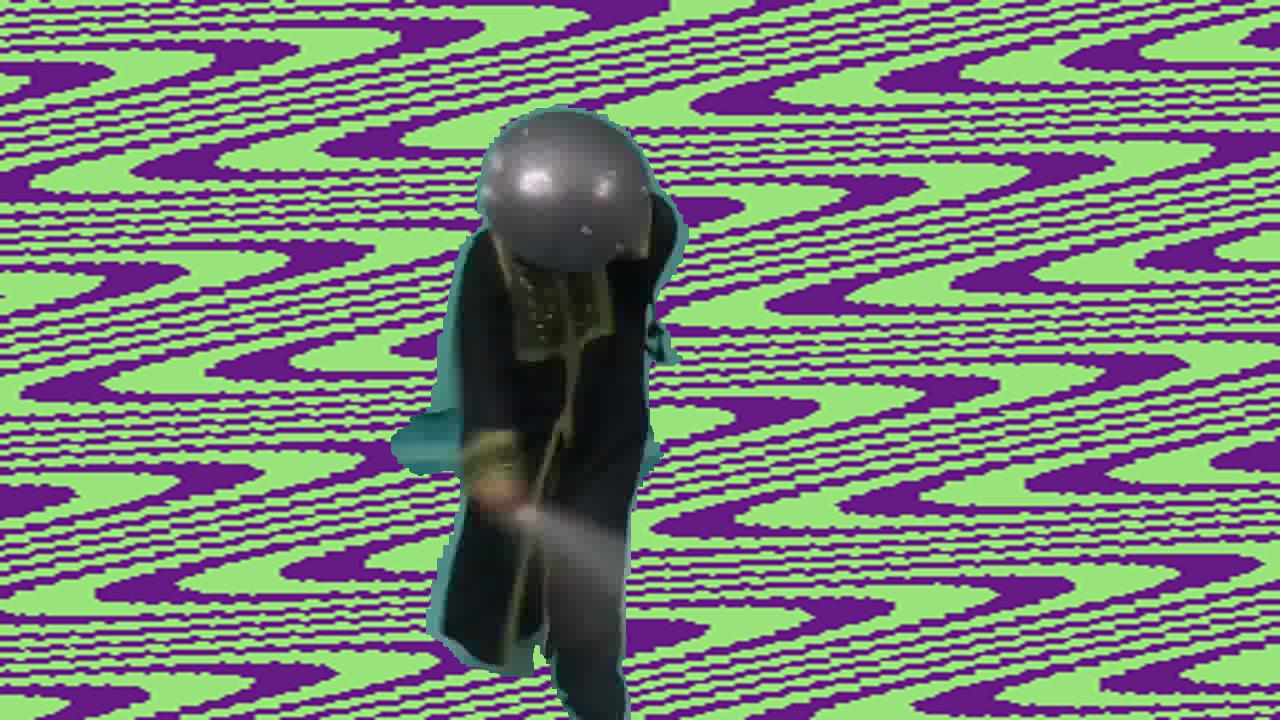}
    \includegraphics[width=0.24\linewidth]{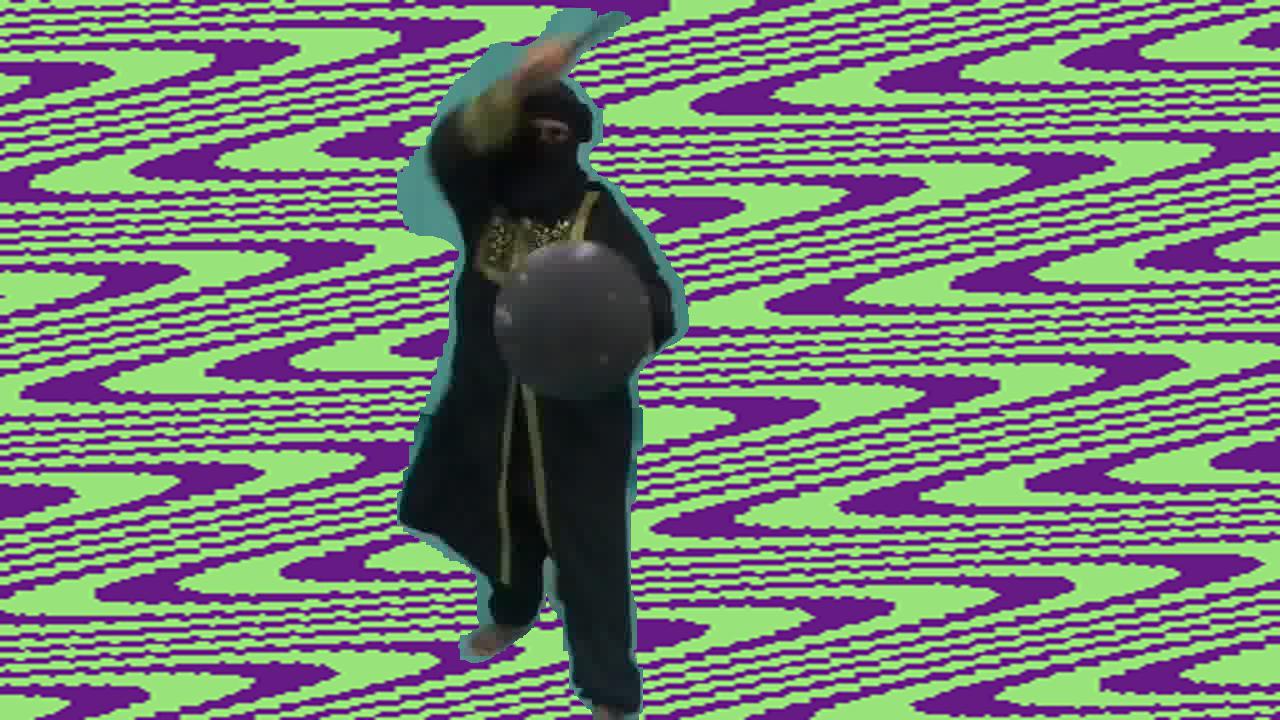}
    \includegraphics[width=0.24\linewidth]{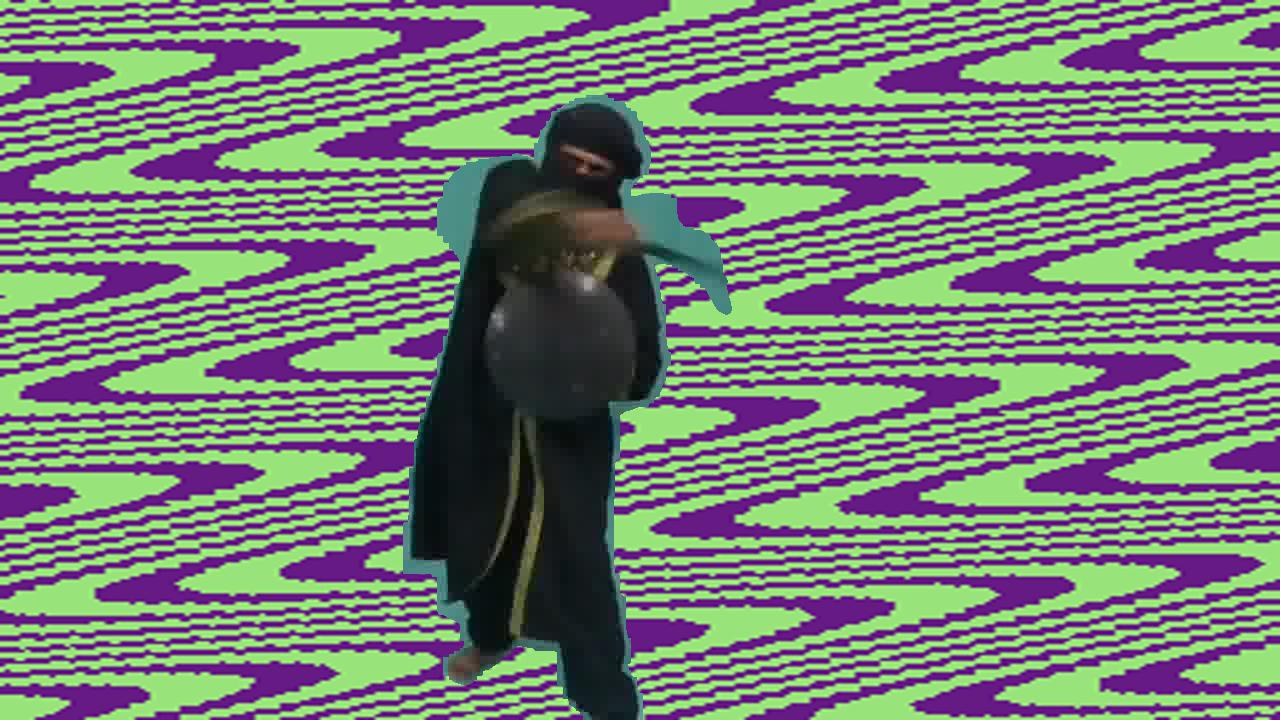}
    \centerline{\textit{Sword Fighting} with sinusoidal backgrounds.}
\end{minipage}
\caption{Examples of the synthetic videos.}
\label{ss_fig:sample_syn_video}
\end{figure}

In Figure \ref{ss_fig:sample_syn_video}, we show some examples of \datasetnameA{} videos. From the shown examples, we observe that the action information is reserved although the backgrounds of the videos are replaced, so that the actions in the synthetic videos are recognizable. From each background image source, we provide two \datasetnameA{} videos in the data appendix for readers' reference.

\subsection{Human Assessment}
We verify that the actions in \datasetnameA{} videos can be recognized by human on Amazon Mechanical Turk (AMT). From the same original video, we randomly sample one synthetic video for assessment. Totally, we have 256 synthetic videos in HMDB51-\datasetnameA{}, 910 synthetic videos in UCF101-\datasetnameA{} and 10,190 synthetic videos in Kinetics400-\datasetnameA{} to be assessed.

\begin{figure}[t]
\centering
\includegraphics[width=\linewidth]{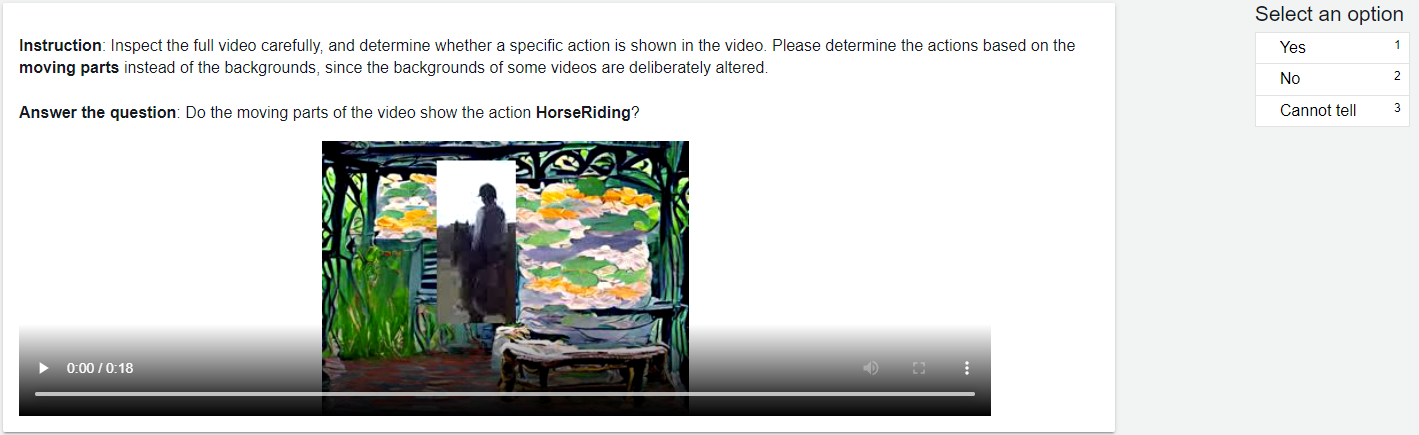}
\caption{Interface of AMT tasks.}
\label{ss_fig:amt_interface}
\end{figure}

The AMT workers are asked to determine whether the moving parts in the videos show the labeled action. Figure~\ref{ss_fig:amt_interface} shows the AMT interface of the assessment task. The interface shows the instruction of the task to workers: ``Inspect the full video carefully, and determine whether a specific action is shown in the video. Please determine the actions based on the moving parts instead of the backgrounds, since the backgrounds of some videos are deliberately altered.'' It also displays a video and a corresponding question: ``Do the moving parts of the video show the action \emph{action\_name}?'' The \emph{action\_name} is the name of the provided action label corresponding to the video. The workers are given three options to select: yes, no, and can’t tell.

\begin{table*}[tb]
\begin{center}
\caption{The hyper-parameters for training TSM, SlowFast, Swin-T without data augmentation or debiasing techniques. UCF, HMDB and K400 denote UCF101, HMDB51 and Kinetics-400, respectively.}
\label{ss_tab:hyperparameters-action-models}
\scriptsize
\setlength\tabcolsep{3pt}
\begin{tabular}{@{}lp{0.02cm}cccp{0.02cm}cccp{0.02cm}cccp{0.02cm}ccp{0.02cm}ccp{0.02cm}cc@{}}
\toprule
\multirow{4}{*}{Hyper-parameter} && \multicolumn{11}{c}{ImangeNet-pretrained} && \multicolumn{8}{c}{Kinetics-pretrained} \\
\cmidrule{3-13} \cmidrule{15-22}
~ && \multicolumn{3}{c}{TSM} && \multicolumn{3}{c}{SlowFast} && \multicolumn{3}{c}{Swin-T} && \multicolumn{2}{c}{TSM} && \multicolumn{2}{c}{SlowFast} && \multicolumn{2}{c}{Swin-T} \\
\cmidrule{3-5} \cmidrule{7-9} \cmidrule{11-13} \cmidrule{15-16} \cmidrule{18-19} \cmidrule{21-22}
~ && UCF & HMDB & K400 && UCF & HMDB & K400 && UCF & HMDB & K400 && UCF & HMDB && UCF & HMDB && UCF & HMDB \\
\midrule
frames per video && 8 & 8 & 8 && 64 & 64 & 64 && 32 & 32 & 32 && 8 & 8 && 64 & 64 && 32 & 32 \\
epoch && 100 & 100 & 50 && 100 & 100 & 50 && 30 & 30 & 30 && 25 & 25 && 25 & 25 && 30 & 30 \\
optimizer && \multicolumn{3}{c}{SGD} && \multicolumn{3}{c}{SGD} && \multicolumn{3}{c}{AdamW} && \multicolumn{2}{c}{SGD} && \multicolumn{2}{c}{SGD} && \multicolumn{2}{c}{AdamW} \\
linear warmup epochs && - & - & - && - & - & - && 2.5 & 2.5 & 2.5 && - & - && - & - && 2.5 & 2.5 \\
base learning rate && 0.0025 & 0.01 & 0.0025 && 0.0025 & 0.005 & 0.01 && 0.002 & 0.005 & 0.001 && 0.0025 & 0.005 && 0.005 & 0.005 && 0.0005 & 0.0005 \\
learning rate schedule && \multicolumn{7}{c}{$\times 0.1$ at 40\% and 80\% of the total epochs} && \multicolumn{3}{c}{cosine} && \multicolumn{5}{c}{$\times 0.1$ at 10$^{\text{th}}$ and 20$^{\text{th}}$ epoch}  && \multicolumn{2}{c}{cosine}\\
weight decay && 0.001 & 0.01 & 0.0001 && 0.001 & 0.01 & 0.0001 && 0.01 & 0.02 & 0.02 && 0.001 & 0.01 && 0.001 & 0.01 && 0.01 & 0.05 \\
\bottomrule
\end{tabular}
\end{center}
\end{table*}

\begin{table*}[tb]
\begin{center}
\caption{The hyper-parameters for training action recognition models with different video data augmentation and debiasing methods. UCF, HMDB and K400 denote UCF101, HMDB51 and Kinetics-400, respectively.}
\label{ss_tab:hyperparameters-data-augmentation}
\tiny
\setlength\tabcolsep{2pt}
\begin{tabular}{@{}llp{0.02cm}cccp{0.02cm}cccp{0.02cm}cccp{0.02cm}ccp{0.02cm}ccp{0.02cm}cc@{}}
\toprule
\multirow{4}{*}{\begin{tabular}[x]{@{}l@{}} Augmentation\\or Debiasing\end{tabular}} & \multirow{4}{*}{Hyper-parameter} && \multicolumn{11}{c}{ImangeNet-pretrained} && \multicolumn{8}{c}{Kinetics-pretrained} \\
\cmidrule{4-14} \cmidrule{16-23}
~ & ~ && \multicolumn{3}{c}{TSM} && \multicolumn{3}{c}{SlowFast} && \multicolumn{3}{c}{Swin-T} && \multicolumn{2}{c}{TSM} && \multicolumn{2}{c}{SlowFast} && \multicolumn{2}{c}{Swin-T} \\
\cmidrule{4-6} \cmidrule{8-10} \cmidrule{12-14} \cmidrule{16-17} \cmidrule{19-20} \cmidrule{22-23}
~ & ~ && UCF & HMDB & K400 && UCF & HMDB & K400 && UCF & HMDB & K400 && UCF & HMDB && UCF & HMDB && UCF & HMDB \\
\midrule
\multirow{4}{*}{Mixup} & base learning rate && 0.01 & 0.005 & 0.005 && 0.005 & 0.005 & 0.01 && 0.005 & 0.005 & 0.001 && 0.0025 & 0.005 && 0.005 & 0.005 && 0.0005 & 0.0005 \\
~ & weight decay && 0.001 & 0.01 & 0.0001 && 0.001 & 0.01 & 0.0001 && 0.01 & 0.02 & 0.02 && 0.001 & 0.01 && 0.001 & 0.01 && 0.01 & 0.05 \\
~ & $P_{\text{aug}}$ && 1.0 & 0.75 & 1.0 && 1.0 & 0.25 & 1.0 && 1.0 & 0.75 & 1.0 && 0.25 & 0.25 && 0.5 & 0.25 && 0.5 & 0.75 \\
~ & $Beta(\alpha, \beta)$ && \multicolumn{19}{c}{$(0.2, 0.2)$} \\
\midrule
\multirow{4}{*}{VideoMix} & base learning rate && 0.01 & 0.01 & 0.005 && 0.005 & 0.005 & 0.01 && 0.002 & 0.005 & 0.001 && 0.005 & 0.005 && 0.005 & 0.005 && 0.0002 & 0.0005 \\
~ & weight decay && 0.001 & 0.01 & 0.0001 && 0.001 & 0.01 & 0.0001 && 0.01 & 0.02 & 0.02 && 0.001 & 0.01 && 0.001 & 0.01 && 0.01 & 0.05 \\
~ & $P_{\text{aug}}$ && 1.0 & 0.25 & 0.75 && 0.5 & 0.5 & 1.0 && 0.75 & 0.75 & 0.75 && 0.25 & 0.5 && 0.25 & 0.25 && 0.5 & 0.25 \\
~ & $Beta(\alpha, \beta)$ && \multicolumn{19}{c}{$(1.0, 1.0)$} \\
\midrule
\multirow{2}{*}{SDN} & base learning rate && 0.02 & 0.02 & 0.01 && 0.02 & 0.02 & 0.01 && 0.002 & 0.002 & 0.001 && 0.002 & 0.01 && 0.02 & 0.01 && 0.0001 & 0.0002 \\
~ & weight decay && 0.001 & 0.01 & 0.0001 && 0.001 & 0.01 & 0.0001 && 0.01 & 0.02 & 0.02 && 0.001 & 0.01 && 0.001 & 0.01 && 0.01 & 0.05 \\
\midrule
\multirow{3}{*}{BE} & base learning rate && 0.0025 & 0.005 & 0.005 && 0.005 & 0.01 & 0.01 && 0.002 & 0.005 & 0.001 && 0.0025 & 0.005 && 0.005 & 0.005 && 0.0005 & 0.0005 \\
~ & weight decay && 0.001 & 0.01 & 0.0001 && 0.001 & 0.01 & 0.0001 && 0.01 & 0.02 & 0.02 && 0.001 & 0.01 && 0.001 & 0.01 && 0.01 & 0.05 \\
~ & $P_{\text{aug}}$ && 0.75 & 0.75 & 0.25 && 1.0 & 0.5 & 0.75 && 0.75 & 0.75 & 0.5 && 0.25 & 0.5 && 0.25 & 0.5 && 0.5 & 0.25 \\
\midrule
\multirow{3}{*}{ActorCutMix} & base learning rate && 0.005 & 0.01 & 0.01 && 0.01 & 0.005 & 0.01 && 0.002 & 0.005 & 0.001 && 0.0025 & 0.005 && 0.005 & 0.005 && 0.0001 & 0.0005 \\
~ & weight decay && 0.001 & 0.01 & 0.0001 && 0.001 & 0.01 & 0.0001 && 0.01 & 0.02 & 0.02 && 0.001 & 0.01 && 0.001 & 0.01 && 0.01 & 0.05 \\
~ & $P_{\text{aug}}$ && 0.75 & 0.5 & 0.25 && 0.25 & 0.5 & 0.25 && 0.5 & 0.75 & 0.25 && 0.25 & 0.25 && 0.5 & 0.5 && 0.5 & 0.25 \\
\midrule
\multirow{3}{*}{FAME} & base learning rate && 0.0025 & 0.01 & 0.005 && 0.0025 & 0.005 & 0.005 && 0.005 & 0.005 & 0.001 && 0.0025 & 0.005 && 0.005 & 0.005 && 0.0005 & 0.0005 \\
~ & weight decay && 0.001 & 0.01 & 0.0001 && 0.001 & 0.01 & 0.0001 && 0.01 & 0.02 & 0.02 && 0.001 & 0.01 && 0.001 & 0.01 && 0.01 & 0.05 \\
~ & $P_{\text{aug}}$ && 0.25 & 0.75 & 0.25 && 0.25 & 0.5 & 0.25 && 0.25 & 0.5 & 0.25 && 0.25 & 0.25 && 0.25 & 0.25 && 0.25 & 0.5 \\
\midrule
\multirow{6}{*}{\sysname{}} & base learning rate && 0.005 & 0.005 & 0.005 && 0.0025 & 0.005 & 0.005 && 0.002 & 0.005 & 0.001 && 0.0025 & 0.005 && 0.005 & 0.005 && 0.0005 & 0.0005 \\
~ & weight decay && 0.001 & 0.01 & 0.0001 && 0.001 & 0.01 & 0.0001 && 0.01 & 0.02 & 0.02 && 0.001 & 0.01 && 0.001 & 0.01 && 0.01 & 0.05 \\
~ & $P_{\text{aug}}$ && 0.25 & 0.5 & 0.25 && 0.75 & 0.25 & 0.125 && 0.75 & 0.5 & 0.25 && 0.25 & 0.25 && 0.5 & 0.25 && 0.25 & 0.75 \\
~ & $\tau$ && 25 & 15 & 25 && 15 & 25 & 25 && 50 & 50 & 50 && 10 & 50 && 10 & 75 && 50 & 10 \\
~ & frame bank size && \multicolumn{19}{c}{4096} \\
~ & $Beta(\alpha, \beta)$ && $(200, 200)$ & $(200, 200)$ & $(20, 20)$ && $(200, 200)$ & $(200, 200)$ & $(2, 2)$ && $(200, 200)$ & $(200, 200)$ & $(20, 60)$ && \multicolumn{2}{c}{$(200, 200)$} && \multicolumn{2}{c}{$(200, 200)$} && \multicolumn{2}{c}{$(200, 200)$} \\
\bottomrule
\end{tabular}
\end{center}
\end{table*}

We divide the videos into a number of groups for assessment. In each group, we create the following questions:
\begin{itemize}
    \setlength{\itemindent}{0em}
    \setlength{\itemsep}{-0.25em}
    \item Experimental group (47.5\% of the total questions): contains synthetic videos with correct labels.
    \item Control group (47.5\% of the total questions): contains synthetic videos with random incorrect labels. The control group is constructed to prevent the workers from always answering yes to synthetic videos.
    \item Control questions (5\% of the total questions): contain original videos, half of which are assigned correct labels and the other are assigned incorrect labels. The control questions are used to detect random clicking.
\end{itemize}

Each question is assigned to three different workers to answer. We accept the answers of a worker only if he or she satisfies the following criteria:
\begin{itemize}
    \setlength{\itemindent}{0em}
    \setlength{\itemsep}{-0.25em}
    \item Answered more than one control questions and reached at least 75\% of accuracy on the answered control questions.
    \item Reached at least 90\% of accuracy on the answered questions in the control group, in which the synthetic videos are assigned incorrect labels. If a worker does not reach high accuracy on these questions, he or she may tend to answer yes to synthetic videos, which affects the assessment results.
\end{itemize}
The final answer for each question is obtained by majority voting.

According to the collected answers, the AMT workers were able to recognize the correct action in 876 videos out of 910 UCF101-\datasetnameA{} videos (96.15\%), 222 videos out of 256 HMDB51-\datasetnameA{} videos (86.33\%) and 8681 videos out of 10190 Kinetics400-\datasetnameA{} videos (85.19\%).

\section{Implementation Details}
\label{ss_sec:imple}

\subsection{Datasets}
\noindent\textbf{UCF101}~\cite{soomro2012ucf101} has 13,320 web videos recorded in unconstrained environments, belonging to 101 classes. We use the first official train-test split in our experiments and report the performance on the test set. 

\noindent\textbf{HMDB51}~\cite{kuehne2011hmdb} consists of 51 classes and 6,766 videos extracted from a variety of sources ranging from digitized movies to YouTube videos. We use the first official train-test split and report the performance on the test set.

\noindent\textbf{Kinetics-400}~\cite{carreira2017quo} contains more than 250k videos in 400 classes. We train the models on the training set (around 240k videos) and reported performance on the validation set (around 20k videos) as in prior works~\cite{lin2019tsm,feichtenhofer2019slowfast,liu2021video,wang2021tdn}.

\subsection{Action Recognition Models}
For TSM~\cite{lin2019tsm}, we use ResNet-50 as the backbone. For SlowFast~\cite{feichtenhofer2019slowfast}, we use 3D ResNet-50 with filters inflated from 2D to 3D \cite{carreira2017quo} as the backbone. And we use the version of $4\times 16$ ($T\times\tau$) in our experiments. For Video Swin Transformer~\cite{liu2021video}, we use the tiny version (denoted as Swin-T) in our experiments.

\subsection{Computational Resources}
Our experiments are conducted on GPU clusters (containing Tesla V100, Tesla P100, GeForce RTX 3090, RTX A6000) with the PyTorch codebase MMAction2 \cite{2020mmaction2}.

\subsection{Training the Reference Network of \sysname{}}
We train the Reference Network $\mathcal{R}$ of \sysname{} with the following settings:
\begin{itemize}
\setlength{\itemindent}{0em}
\setlength{\itemsep}{-0.25em}
\item Network: ResNet-50, SlowFast-2D (ResNet-50 as backbone), tiny Swin Transformer
\item Pretrained: ImageNet
\item Optimizer: SGD
\item Base learning rate: 0.01. The base learning rate corresponds to the batch size of 64. We apply the Linear Scaling Rule \cite{goyal2017accurate} to set the learning rate according to the real batch sizes.
\item Epochs: 50
\item Learning rate schedule: learning rate is divided by 10 at the 20$^{\text{th}}$ and 40$^{\text{th}}$ epoch
\item Weight decay: 0.00001
\end{itemize}

\subsection{Training the Main Network}
\noindent\textbf{Random Seeds.} On UCF101 and HMDB51, for each model and each data augmentation or debiasing method, we fix the random seeds as 1, 2, and 3 to conduct three times of training. The reported accuracies are the mean accuracies of the three runs. On Kinetics-400, we fix the random seeds as 1 and conduct only one time of training.

\vspace{0.2em}\noindent\textbf{Other Data Augmentations.} Except the data augmentation methods discussed in the main paper, we also use two commonly used data augmentations for each model during training: \emph{(1)} The shorter ends of video frames are resized to 256 and an area of $224\times 224$ is randomly cropped. \emph{(2)} Each video is flipped horizontally with a probability of 0.5.

\vspace{0.2em}\noindent\textbf{Hyper-parameters.}
On UCF101 and HMDB51, we randomly sample 20\% of the training samples to form a validation set for hyper-parameter tuning. On Kinetics-400, we randomly sample 50\% of the training samples to form a training-validation split to tune hyper-parameters. In the training-validation split, the proportion of training and validation samples is 19:1. After the best hyper-parameters are selected, we train the models on the full training set with the best hyper-parameters. 

For action recognition models without data augmentation or debiasing methods applied, we tune the base learning rate and the weight decay on the validation set and fix other hyper-parameters as pre-defined values. The base learning rate corresponds to the batch size of 64, and we apply the Linear Scaling Rule \cite{goyal2017accurate} to set the learning rate according to the real batch sizes. For data augmentation methods, we additionally tune the augmentation probability $P_{\text{aug}}$ on the validation set. Table \ref{ss_tab:hyperparameters-action-models} and \ref{ss_tab:hyperparameters-data-augmentation} show the hyper-parameters we used in different action recognition models as well as the data augmentation and debiasing methods.

\subsection{Evaluation}
The checkpoint at the last epoch is used for evaluation. Given a video, we resize the shorter ends of frames to $256$ and use a center crop of $224\times 224$ from a single clip for evaluation.



\clearpage
{\small
\bibliographystyle{ieee_fullname}
\bibliography{egbib}
}

\end{document}